\documentclass[a4paper,11pt]{article}

\usepackage{jmlr2e}

\usepackage{macrosArticle}
\usepackage{macrosText}

\jmlrshortheading{2018}{Lilian Besson and Emilie Kaufmann}

\begin{document}

% -----------------------------------------------------------------
\editor{Kevin Murphy and Bernhard Sch{\"o}lkopf}
% FIXME uncomment the \editor part in the jmlr2e.sty

\title{Multi-Player Bandits Revisited}

% -----------------------------------------------------------------
% % the name(s) of the author(s) follow(s) next
% \author{
%     Lilian Besson \\
%     CentraleSup\'elec, IETR, SCEE Team, \\
%     Avenue de la Boulaie -- CS 47601, \\
%     $35576$ Cesson-S\'evign\'e, France
%   \and
%     Emilie Kaufmann \\
%     Univ. Lille 1, CNRS, Inria, SequeL Team \\
%     UMR 9189 -- CRIStAL, \\
%     $59000$ Lille, France
% }

\author{\name Lilian Besson\textsuperscript{$\dagger$} \email {Lilian}{.}{Besson}{@}{CentraleSupelec}{.}{fr} \\
        \addr CentraleSup\'elec (campus of Rennes), IETR, SCEE Team,\\
        Avenue de la Boulaie -- CS $47601$, F-$35576$ Cesson-S\'evign\'e, France
        \AND
        \name Emilie Kaufmann \email {emilie}{.}{kaufmann}{@}{univ}{-}{lille1}{.}{fr} \\
        \addr CNRS \& Universit\'e de Lille, Inria SequeL team\\
        UMR 9189 -- CRIStAL,  F-$59000$ Lille, France
        % \AND
        % \name Christophe Moy \email {Christophe}{.}{Moy}{@}{Univ}-{Rennes1}{.}{fr} \\
        % \addr Universit\'e de Rennes 1, IETR, SCEE Team,\\
        %   263 ave. du Général Leclerc -- CS $74205$, F-$35042$ Rennes, France
}

\date{}  % remove date

\hypersetup{pageanchor=false}
\maketitle

% -----------------------------------------------------------------
\begin{abstract}%   <- trailing '%' for backward compatibility of .sty file

  Multi-player Multi-Armed Bandits (MAB) have been extensively studied in the literature, motivated by applications to Cognitive Radio systems.
  Driven by such applications as well, we motivate the introduction of several levels of feedback for multi-player MAB algorithms. Most existing work assume
  that \emph{sensing information} is available to the algorithm. Under this assumption, we improve the state-of-the-art lower bound for the regret of any decentralized algorithms
  and introduce two algorithms, \RandTopM{} and \MCTopM, that are shown to empirically outperform existing algorithms. Moreover, we provide strong theoretical guarantees for these algorithms, including
  a notion of asymptotic optimality in terms of the number of selections of bad arms.
  We then introduce a promising heuristic, called \Selfish, that can operate
  without sensing information, which is crucial for emerging applications to Internet of Things networks. We investigate the empirical performance of this algorithm
  and provide some first theoretical elements for the understanding of its behavior.

  % % -----------------------------------------------------------------
  % \textbf{Keywords:}
  %   Multi-Armed Bandits; Decentralized algorithms; Reinforcement learning;
  %   Cognitive Radio; Opportunistic Spectrum Access.
  %   % XXX check if this kind of article on JMLR has other keywords? @Emilie?
\end{abstract}

% % -----------------------------------------------------------------
% \iffalse  % XXX uncomment at the end !
% \tableofcontents  % https://en.wikibooks.org/wiki/LaTeX/Document_Structure#Table_of_contents
% \listoffigures  % https://en.wikibooks.org/wiki/LaTeX/Document_Structure#Table_of_contents
% \listofalgorithms
% % \listoftables  % https://en.wikibooks.org/wiki/LaTeX/Document_Structure#Table_of_contents
% \fi  % XXX uncomment at the end !

\begin{keywords}
    Multi-Armed Bandits; Decentralized algorithms; Reinforcement learning;
    Cognitive Radio; Opportunistic Spectrum Access.
    % XXX check if this kind of article on JMLR has other keywords? @Emilie?
\end{keywords}

\hypersetup{pageanchor=true}
\pagenumbering{arabic}

% -----------------------------------------------------------------
% -----------------------------------------------------------------
\section{Introduction}
\label{sec:introduction}
% the paragraph below may be useful to state in the intro or early in the paper

% general intro, UCB

Several sequential decision making problems under the constraint of partial information
have been studied since the 1950s under the name of Multi-Armed Bandit (MAB) problems \citep{Robbins52,LaiRobbins85}.
In a stochastic MAB model, an agent is facing $K$ unknown probability distributions, called arms in reference to the arms of a one-armed bandit (or slot machine) in a casino.
Each time she selects (or draws) an arm, she receives a reward drawn from the associated distribution.
Her goal is to build a sequential selection strategy that maximizes the total reward received.
A class of algorithms to solve this problem is based on Upper Confidence Bounds (UCB), first proposed by \cite{LaiRobbins85,Agrawal95} and further popularized by \cite{Auer02}.
The field has been very active since then, with several algorithms proposed and analyzed, both theoretically and empirically, even beyond the stochastic assumption on arms, as explained in the survey by \cite{Bubeck12}.

% appli, from clinical trials to cognitive radios

The initial motivation to study MAB problems arose from clinical trials (the first MAB model can be traced back to $1933$, by \citeauthor{Thompson33}), in which a doctor sequentially allocates treatments (arms) to patients and observes their efficacy (reward).
More recently, applications of MAB have shifted towards sequential content recommendation, \eg{} sequential display of advertising to customers or A/B testing \citep{Li10contextual,Chapelleetal14Ad}.
In the mean time, MAB were found to be relevant to the field of Cognitive Radio (CR, \cite{Mitola99}),
and \cite{Jouini09,Jouini10} first proposed to use \UCB{} for the Opportunistic Spectrum Access (OSA) problem,
and successfully conducted experiments on real radio networks demonstrating its usefulness.
For CR applications, each arm models the quality or availability of a radio channel (a frequency band) in which there is some background traffic (\eg, primary users paying to have a guaranteed access to the channel in the case of OSA).
A smart radio device needs to insert itself in the background traffic, by sequentially choosing a channel to access and try to communicate on, seeking to optimize the quality of its global transmissions.

% the need for multiple players in cognitive radio

For the development of CR, a crucial step is to insert \emph{multiple} $M \geq 2$ smart devices in the \emph{same} background traffic.
With the presence of a central controller that can assign the devices to separate channels, this amounts to choosing at each time step \emph{several} arms of a MAB in order to maximize the global rewards, and can thus be viewed as an application of the multiple-play bandit, introduced by \cite{Anantharam87a} and recently studied by \cite{Komiyama15}.
%who essentially proved that existing algorithms can be easily extended to the multiple-play case, with provable guarantees on their regret.
% In the context of CR, this first extension can model a centralized agent taking decisions for end-devices, for instance a base station affecting devices to channels.
% Multiple-Play MAB algorithms successfully used for other Cognitive Radio tasks,
% for instance by \cite{Maghsudi16} for 5G small-cells or by \cite{Bonnefoi16} for Internet-of-Things (IoT) networks.
%
% Another generalization targets the opposite goal, where several devices
% are learning and conjointly accessing the same network,
% trying to find a consensus in a decentralized learning way.
% It is harder as there is a game theoretic-like situation, for instance where devices try to avoid collisions.
%
Due to the communication cost implied by a central controller, a more relevant model is the
\emph{decentralized multi-player} multi-armed bandit model, introduced by \cite{Zhao10} and \cite{Anandkumar10,Anandkumar11}, in which players select arms individually and collisions may occur, that yield a loss of reward.
Further algorithms were proposed in similar models by \cite{Tekin12IEEE} and \cite{Kalathil12} (under the assumption that each arm is a Markov chain)
and by \cite{Avner15,Avner16} and \cite{Rosenski16} (for \iid{} or piece-wise \iid{} arms).
The goal for every player is to select most of the time one of the $M$ best arms, without colliding too often with other players.
A first difficulty relies in the well-known trade-off between \emph{exploration} and \emph{exploitation}: players need to explore all arms to estimate their means while trying to focus on the best arms to gain as much rewards as possible.
The decentralized setting considers no exchange of information between players, that only know $K$ and $M$, and to avoid collisions, players should furthermore find orthogonal configurations (\ie, the $M$ players use the $M$ best arms without any collision), without communicating.
Hence, in that case the trade-off is to be found between exploration, exploitation \emph{and} low collisions.

All these above-mentioned works are motivated by the OSA problem, in which it is assumed that \emph{sensing} occurs, that is each smart device observes the availability of a channel (sample from the arm) \emph{before} trying to transmit and possibly experiment a collision with other smart devices.
However some real radio networks do not use sensing at all, \eg, emerging standards developed for \emph{Internet of Things} (IoT) networks such as LoRaWAN.
Thus, to take into account these new applications, algorithms with additional constraints on the available feedback have to be proposed within the multiple-player MAB model.
Especially, the typical approach that combines a (single-player) bandit algorithm based on the sensing information --to learn the quality of the channels while targeting the best ones-- with a low-complexity decentralized collision avoidance protocol, is no longer possible.

In this paper, we take a step back and present the different feedback levels possible for multi-player MAB algorithms. For each of them, we propose algorithmic solutions supported by both experimental and theoretical guarantees. In the presence of sensing information, our contributions are a new problem-dependent regret lower bound, tighter than previous work, and the introduction of two algorithms, \RandTopM{} and \MCTopM. Both are shown to achieve an asymptotically optimal number of selections of the sub-optimal arms, and for \MCTopM{} we furthermore establish a logarithmic upper bound on the regret, that follows from a careful control of the number of collisions. In the absence of sensing information, we propose the \Selfish{} heuristic and investigate its performance. Our study of  this algorithm is supported by (promising) empirical performance and some first (disappointing) theoretical elements.

The rest of the article is organized as follows.
We introduce the multi-player bandit model with three feedback levels in Section~\ref{sec:model}, and give a new regret lower bound in Section~\ref{sec:lowerbound}.
The \RandTopM, \MCTopM{} and \Selfish{} algorithms are introduced in Section~\ref{sec:algorithms}, with the result of our experimental study reported in
Section~\ref{sec:experiments}. Theoretical elements are then presented in Section~\ref{sec:upperbounds}.
% and finally Section~\ref{sec:conclusion} concludes.

% Short headings should be running head and authors last names
\ShortHeadings{Multi-Player Bandits Models Revisited}{Besson and Kaufmann}

% -----------------------------------------------------------------
% -----------------------------------------------------------------
\section{Multi-Player Bandit Model with Different Feedback Levels}
\label{sec:model}

% describe our stochastic assumptions
We consider a $K$-armed Bernoulli bandit model, % (for $K\geq2$),
in which arm $k$ is a Bernoulli distribution with mean $\mu_k\in[0,1]$.
We denote $(Y_{k,t})_{t\in\N}$ the \iid{} (binary) \emph{reward stream} for arm $k$, that satisfies $\Pr(Y_{k,t}=1) = \mu_k$ and that is independent from the other rewards streams.
However we mention that our lower bound and all our algorithms (and their analysis) can be easily extended to one-dimensional exponential families (just like for the \klUCB{} algorithm of \cite{KLUCBJournal}). For simplicity, we focus on the Bernoulli case, that is also the most relevant for Cognitive Radio, as it can model channel availabilities.

% stochastic bandit model in which each arm distribution is assumed to belong to a one-dimensional canonical exponential family. That is, arm $k$ has a density with respect to a reference measure that takes the form
% \[f_{\theta_k}(x) = \exp(\theta_k x - b(\theta_k)),\]
% there $\theta_k \in \R$ is some parameter and $b$ is a twice-differentiable function. It is well known that such distributions can be alternatively parameterized by their means (see, \eg, KLUCB), and we denote by $\mu_k$ the mean of arm $k$. Examples of exponential families include Gaussian distribution with known variance, exponential, Poisson distributions and Bernoulli distributions. For cognitive radio applications, it is often assumed that each arm is a Bernoulli distribution, giving 1 if the associated channel is available, 0 other. We let $(Y_{k,t})_{t\in\N}$ denote the \iid{} reward stream for arm $k$, whose distribution has mean $\mu_k$ and that is independent from other arms.

% explain the interaction protocol and the GOAL
In the multi-player MAB setting, there are $M \in \{1,\dots,K\}$ players (or agents),
that have to make decisions at some pre-specified time instants.
At time step $t \in\mathbb{N},t\geq1$, player $j$ selects an arm $A^j(t)$, independently from the other players' selections.
A \emph{collision} occurs at time $t$ if at least two players choose the same arm.
We introduce the two events, for $j\in\{1,\dots,M\}$ and $k\in\{1,\dots,K\}$,
\begin{equation}
  C^j(t) :=  \{ \exists j' \neq j : A^{j'}(t) = A^j(t) \}
  \ \ \ \text{and} \ \ \ C_k(t) :=  \left\{ \# \{ j : A^j(t) = k\} > 1 \right\},
\end{equation}
that respectively indicate that a collision occurs at time $t$ for player $j$ and that a collision occurs at time $t$ on arm $k$.
Each player $j$ then receives (and observes) the \emph{binary rewards}
$r^j(t) \in \{0,1\}$,
\begin{equation}
  r^j(t) := Y_{A^j(t),t} \; \indic(\overline{C^j(t)}).
\end{equation}
In words, she receives the reward of the selected arm if she is the only one to select this arm, and a reward zero otherwise\footnote{This provides another reason to focus on the Bernoulli model. It is the hardest model, in the sense that receiving a reward zero is not enough to detect collisions. For other models, the data streams $(Y_{k,s})_s$ are usually continuously distributed, with no mass at zero. Hence receiving $r^j(t) = 0$ directly gives $\indic(C^j(t)) = 1$.}.
Other models for rewards loss have been proposed in the literature (\eg, the reward is randomly allocated to one of the players selecting it), but we focus on full reward occlusion in this article.

A multi-player MAB strategy is a tuple $\rho = (\rho^1,\dots,\rho^M)$ of arm selection strategies for each player, and the goal is to propose a strategy that maximizes the total reward of the system, under some constraints.
First, each player $j$ should adopt a \emph{sequential} strategy $\rho^j$, that decides which arm to select at time $t$ based on \emph{previous observations}.
Previous observations for player $j$ at time $t$ always include the previously chosen arms $A^j(s)$ and received rewards $r^j(s)$ for $s<t$, but may also include the \emph{sensing information} $Y_{A^j(t),t}$ or the \emph{collision information} $C^j(t)$.
More precisely, depending on the application, one may consider the following three observation models, \modelun, \modeldeux{} and \modeltrois.
%
% If the arms have continuous distributions (or are such that  $\Pr(Y_{k,s}=0)=0$), the \emph{sensing information} $Y_{A^j(t),t}$ and \emph{collision information} $C^j(t)$ can always be extracted from the reward information. But for Bernoulli distribution, one may consider the following three observation models, that are not equivalent:
\begin{itemize}
  \item[\modelun]
    \textbf{Simultaneous sensing and collision}: player $j$ observes  $Y_{A^j(t),t}$ \emph{and} $C^j(t)$ (not previously studied).
  \item[\modeldeux]
    \textbf{Sensing, then collision}: player $j$ observes $Y_{A^j(t),t}$, \emph{then} observes the reward, and thus also $C^j(t)$ only if $Y_{A^j(t),t} = 1$.
    This common setup, studied for example by \cite{Anandkumar11,Avner15,Rosenski16}, is relevant to model the OSA problem: the device first checks for the presence of primary users in the chosen channel.
    If this channel is free ($Y_{A^j(t),t}=1$), the transmission is successful ($r^j(t)=1$) if no collision occurs with other smart devices ($\overline{C^j(t)}$).
  \item[\modeltrois]
    \textbf{No sensing}: player $j$ only observes the reward $r^j(t)$.
    For IoT networks, this reward can be interpreted as an acknowledgement from a Base Station,
    received when a communication was successful.
    A lack of acknowledgment may be due to a collision
    with a device from the background traffic $(Y_{A^j(t),t}=0)$,
    or to a collision with one of the others players ($C^j(t)$).
    However, the sensing and collision information are censored.
    Recently, \cite{Bonnefoi17} presented the first (bandit-based) algorithmic solutions under this (harder) feedback model, in a slightly different setup, more suited to large scale IoT applications. %\footnote{\cite{Bonnefoi17} considers a different setting, more suited for large-scale IoT networks, and a future work will be to extend our work to this setting with more than $K$ players but small probability of activations.}.
\end{itemize}
\noindent Under each of these three models, we define $\cF^j_t$ to be the filtration generated by the observations gathered by player $j$ up to time $t$ (which contains different information under models \modelun, \modeldeux{} and \modeltrois).
While a \emph{centralized} algorithm may select the vector of actions for all players $(A^1(t),\dots,A^M(t))$ based on all the observations from $\bigcup_j \cF^j_{t-1}$, under a \emph{decentralized} algorithm the arm selected at time $t$ by player $j$ only depends on the past observation of this player.
More formally, $A^j(t)$ is assumed to be $\cF^j_{t-1}$-measurable.

% [Index sets \Mworst, \Mbest]
\begin{definition}\label{def:MbestMworst}
  We denote by $\mu_1^*$ the best mean, $\mu_2^*$ the second best etc, and
  by \Mbest{} the (non-sorted) set of the indices of the $M$ arms with largest mean (\emph{best arms}): if $\mu_1^* = \mu_{k_1}, \dots, \mu_M^* = \mu_{k_M}$
  then $\Mbest = \{k_1, \dots, k_M\}$.
  Similarly, \Mworst{} denotes the set of indices of the $K-M$ arms with smallest means (\emph{worst arms}),
  $\{1, \dots, K\} \setminus \Mbest$.
  Note that they are both uniquely defined if $\mu_M^* > \mu_{M+1}^*$.
\end{definition}

Following a natural approach in the bandit literature, we evaluate the performance of a multi-player strategy using the \emph{expected regret} (later simply referred to as regret), that measures the performance gap with respect to the best possible strategy.
The regret of the strategy $\rho$ at horizon $T$ is the difference between the cumulated reward of an oracle strategy, assigning in this case the $M$ players to \Mbest,
and the cumulated reward of strategy $\rho$:
\begin{equation}\label{eq:regret}
  R_T(\boldsymbol{\mu}, M, \rho) := \left(\sum_{k=1}^{M}\mu_k^*\right)T - \E_{\mu}\left[\sum_{t=1}^T\sum_{j=1}^M r^j(t)\right].
\end{equation}
Maximizing the expected sum of global reward of the system is indeed equivalent to minimizing the regret, and we now investigate the best possible \emph{regret rate} of a decentralized multi-player algorithm.

% -----------------------------------------------------------------
% -----------------------------------------------------------------
\section{An Asymptotic Regret Lower Bound}
\label{sec:lowerbound}

In this section, we provide a useful decomposition of the regret (Lemma~\ref{lem:DecompositionRegret}) that permits to establish a new problem-dependent lower bound on the regret (Theorem~\ref{thm:BetterLowerBound}), and also provides key insights on the derivation of regret upper bounds (Lemma~\ref{lem:1stUpperBound}).

% -----------------------------------------------------------------
\subsection{A Useful Regret Decomposition}
\label{sub:defregret}

We introduce additional notations in the following definition.

\begin{definition}
  % [$T_k(T)$ and $\cC_k(T)$]
  \label{def:nbSelections_nbCollisions}
  % \label{def:nbSelections}
  Let $T^j_k(T) := \sum_{t=1}^T \indic(A^j(t) = k)$,
  and denote $T_k(T) := \sum_{j=1}^M T^j_k(T)$ the \emph{number of selections} of arm $k\in\{1,\dots,K\}$ by any player $j\in\{1,\dots,M\}$, up to time $T$.
  % and $\cC_k(T)$ be the number of collisions on arm $k$ up-to time $T$.
% \end{definition}

% \begin{definition}[]\label{def:nbCollisions}
  Let $\cC_k(T)$ be the \emph{number of colliding players}\footnote{When $n$ players choose arm $k$ at time $t$, this counts as $n$ collisions, not just one. So $\cC_k(T)$ counts the total \emph{number of colliding players} rather than the number of collision events. Hence there is small abuse of notation when calling it a number of collisions.}
  on arm $k\in\{1,\dots,K\}$ up to horizon $T$:
  \vspace*{-5pt}  % XXX
  \begin{equation}
    \cC_k(T) :=
    \sum_{t=1}^{T} \sum_{j=1}^{M} \indic(C^j(t)) \indic(A^j(t) = k).
  \end{equation}
\end{definition}

Letting $\cP_M = \left\{ \boldsymbol{\mu} \in [0,1]^K : \mu_M^* > \mu_{M+1}^*\right\}$
be the set of bandit instances such that there is a strict gap between the $M$ best arms and the other arms, we now provide a regret decomposition for any $\boldsymbol{\mu} \in \cP_M$.

\begin{lemma}\label{lem:DecompositionRegret}
  For any bandit instance $\boldsymbol{\mu}\in\cP_M$ such that $\mu_M^* > \mu_{M+1}^*$, it holds that
  \hfill{}
  (Proved in App.\ref{proof:DecompositionRegret})
  \vspace*{-5pt}  % XXX
  \begin{equation*}\label{eq:termeundeuxtermetrois}
    R_T(\boldsymbol{\mu}, M, \rho) =
    \underbrace{\sum_{k \in \Mworst} (\mu_M^* -  \mu_k) \E_{\mu}[T_k(T)]}_{\emph{\mytag{(a)}{term1}}}
    + \underbrace{\sum_{k \in \Mbest} (\mu_k -  \mu_M^*) (T - \E_{\mu}[T_k(T)])}_{\emph{\mytag{(b)}{term2}}}
    + \underbrace{\sum_{k=1}^{K} \mu_k \E_{\mu}[\cC_k(T)]}_{\emph{\mytag{(c)}{term3}}}.
  \end{equation*}
\end{lemma}

In this decomposition, term \ref{term1} counts the lost rewards due to \emph{sub-optimal arms} selections ($k \in \Mworst$), term \ref{term2} counts the number of times the \emph{best arms} were \emph{not} selected ($k \in \Mbest$), and term \ref{term3} counts the weighted number of collisions, on \emph{all arms}. It is valid for both centralized and decentralized algorithms. For centralized algorithms, due to the absence of collisions, \termetrois{} is obviously zero, and \termedeux{} is non-negative, as $T_k(T) \leq T$. For decentralized algorithms, \termetrois{} may be significantly large, and term \termedeux{} may be negative, as many collisions on arm $k$ may lead to $T_k(T) > T$. However, a careful manipulation of this decomposition (see Appendix~\ref{proof:1stLowerBound}) shows that the regret is always lower bounded by term \termeun.

\begin{lemma}\label{lem:1stLowerBound}
    For any strategy $\rho$ and $\boldsymbol{\mu}\in\cP_M$, it holds that
    $ R_T(\boldsymbol{\mu}, M, \rho)    \geq \sum\limits_{k \in \Mworst} (\mu_M^*- \mu_k) \E_{\mu}[T_k(T)]$.
\end{lemma}

% -----------------------------------------------------------------
\subsection{An Improved Asymptotic Lower Bound on the Regret}
\label{sub:betterLowerBound}

To express our lower bound, we need to introduce $\kl(x,y)$ as the Kullback-Leibler divergence between the Bernoulli distribution of mean $x \neq 0,1$ and that of mean $y \neq 0,1$, so that $\kl(x,y) := x\log(x/y) + (1-x)\log((1-x)/(1-y))$.
We first introduce the assumption under which we derive a regret lower bound, that generalizes a classical assumption made by \cite{LaiRobbins85} in single-player bandit models.

\begin{definition}\label{def:DecentralizedUniformEfficiency}
  A strategy $\rho$  is \emph{\textbf{strongly uniformly efficient}} if for all $\boldsymbol{\mu} \in \cP_M$ and for all $\alpha \in (0,1)$,
  \vspace*{-5pt}  % XXX
  \begin{equation}
    R_T(\boldsymbol{\mu},M,\rho) \mathop{=}\limits_{T \to +\infty} o(T^\alpha) \ \ \ \text{and } \ \
    \forall j \in \{1,\dots,M\}, k \in \Mbest, \;\;
    % a_{j,k}
    \frac{T}{M}
    - \E_{\mu}[T^j_k(T)] \mathop{=}\limits_{T \to +\infty} o(T^{\alpha}).
    \label{eq:SUE}
  \end{equation}
\end{definition}

Having a small regret on every problem instance,
\ie, uniform efficiency,
is a natural assumption for algorithms,
that rules out algorithms tuned to perform well on specific instances only.
From this assumption $\left(R_T(\boldsymbol{\mu},M,\rho)=o(T^\alpha)\right)$ and the decomposition of Lemma~\ref{lem:DecompositionRegret} one can see\footnote{With some arguments used in the proof of Lemma~\ref{lem:1stLowerBound} to circumvent the fact that \termedeux{} may be negative.} that for every $k \in \Mbest$,
${T}- \E_{\mu}[T_k(T)] {=} o(T^{\alpha})$, and so
\begin{equation}\label{eq:intermediate}
  \vspace*{-5pt}  % XXX
  \sum_{j =1}^M\left(\frac{T}{M} - \E_{\mu}[T^j_k(T)]\right) = o(T^{\alpha}).
  % \vspace*{-5pt}  % XXX
\end{equation}
The additional assumption in \eqref{eq:SUE} further implies some notion of \emph{fairness}, as it suggests that each of the $M$ players spends on average the same amount of time on each of the $M$ best arms. Note that this assumption is satisfied by any strategy that is invariant under every permutation of the players, \ie, for which the distribution of the observations under $\rho^{\gamma} = (\rho^{\gamma(1)},\dots,\rho^{\gamma(M)})$ is independent from the choice of permutation $\gamma \in \Sigma_M$. In that case, it holds that  $\E_{\mu}[T_k^j(T)]=\E_{\mu}[T_k^{j'}(T)]$ for every arm $k$ and $(j,j') \in \{1,\dots,M\}$, hence \eqref{eq:SUE} and \eqref{eq:intermediate} are equivalent, and strong uniform efficiency is equivalent to standard uniform efficiency. Note that all our proposed algorithms are permutation invariant and \MCTopM{} is thus an example of strongly uniformly efficient algorithm, as we prove in Section~\ref{sec:upperbounds} that its regret is logarithmic on every instance $\mu \in \cP_M$.

We now state a problem-dependent asymptotic lower bound on the number of sub-optimal arms selections under a \emph{decentralized strategy} that has access to the sensing information.
This result, proved in Appendix~\ref{proof:BetterLowerBound}, yields an asymptotic logarithmic lower bound on the regret, also given in Theorem~\ref{thm:BetterLowerBound}.

\begin{theorem}\label{thm:BetterLowerBound}
  Under observation models $\modelun$ and $\modeldeux$, for any strongly uniformly efficient \emph{decentralized} policy $\rho$ and $\boldsymbol{\mu}\in\cP_M$,
  \vspace*{-10pt}  % XXX
  \begin{equation}\label{eq:LBDraws}
    \forall j \in \{1,\dots,M\}, \ \forall k \in \Mworst, \ \ \ \liminf_{T\to \infty} \frac{\E_{\mu}[T_k^j(T)]}{\log(T)} \geq \frac{1}{\kl(\mu_k, \mu_M^*)}.
  \end{equation}

  % \vspace{-0.3cm}  % XXX
  \noindent From Lemma~\ref{lem:1stLowerBound}, it follows that
  \vspace*{-5pt}  % XXX
  \begin{equation}\label{eq:ourLowerBound}
    \mathop{\lim\inf}\limits_{T \to +\infty} \frac{R_T(\boldsymbol{\mu}, M, \rho)}{\log(T)}
    \geq M \times \left( \sum_{k \in \Mworst} \frac{(\mu_M^* -  \mu_k)}{\kl(\mu_k, \mu_M^*)} \right) .
  \end{equation}
\end{theorem}

Observe that the regret lower bound \eqref{eq:ourLowerBound} is tighter than the state-of-the-art lower bound in this setup
given by \cite{Zhao10}, that states that
\begin{equation}\label{eq:Zhao10LowerBound}
  \mathop{\lim\inf}\limits_{T \to +\infty} \frac{R_T(\boldsymbol{\mu}, M, \rho)}{\log(T)}
  \geq \sum_{k \in \Mworst} \left( \sum_{j=1}^{M} \frac{(\mu_M^* -  \mu_k)}{\kl(\mu_k, \mu_{j}^*)} \right),
\end{equation}
as for every $k \in \Mworst$ and $j \in \{1,\dots,M\}$, $\kl(\mu_k, \mu_j^*) \geq \kl(\mu_k, \mu_M^*)$
(see Figure~\ref{fig:CompLowerBounds} in Appendix~\ref{app:illustrationLowerBound}).
It is worth mentioning that \cite{Zhao10} proved a lower bound under the more general assumption for $\rho$ that there exists some numbers $(a_{k}^j)$ such that $a_{k}^j T - \E_{\mu}[T_k^j(T)] = o(T^\alpha)$ whereas in Definition~\ref{def:DecentralizedUniformEfficiency} we make the choice $a_{k}^j = 1/M$.
Our result could be extended to this case but we chose to keep the notation simple and focus on \emph{fair allocation} of the optimal arms between players.

Interestingly, our lower bound is exactly a multiplicative constant factor $M$ away from the lower bound given by \cite{Anantharam87a} for centralized algorithms (which is clearly a simpler setting). This intuitively suggests the number of players $M$ as the (multiplicative) \emph{``price of decentralized learning''}. However, to establish our regret bound, we lower bounded the number of collisions by zero, which may be too optimistic.
Indeed, for an algorithm to attain the lower bound \eqref{eq:ourLowerBound}, the number of selections of each sub-optimal arm should match the lower bound \eqref{eq:LBDraws} \emph{and} term \termedeux{} and term \termetrois{} in the regret decomposition of Lemma~\ref{lem:DecompositionRegret} should be negligible compared to  $\log(T)$.
To the best of our knowledge, no algorithm has been shown to experience only $o(\log(T))$ collisions so far,
for every $M \in \{2,\dots,K\}$ and $\boldsymbol{\mu} \in \cP_M$.

A lower bound on the minimal number of collisions experienced by any strongly uniformly efficient decentralized algorithm would thus be a nice complement to our Theorem~\ref{thm:BetterLowerBound}, and it is left as future work.

% -----------------------------------------------------------------
\subsection{Towards Regret Upper Bounds}

A natural approach to obtain an upper bound on the regret of an algorithm is to upper bound separately each of the three terms defined in Lemma~\ref{lem:DecompositionRegret}.
The following result shows that term $(b)$ can be related to the number of sub-optimal selections and the number of collisions that occurs on the $M$ best arms.

% \vspace*{-15pt}  % XXX remove if problem
\begin{lemma}\label{lem:1stUpperBound}
  The term $(b)$ in Lemma~\ref{lem:DecompositionRegret} is upper bounded as
  \hfill{}
  (Proved in Appendix~\ref{proof:1stUpperBound})
  \begin{equation}\label{eq:1stUpperBound}
    (b) \leq (\mu_1^* - \mu_M^*) \Bigl(    \sum_{k \in \Mworst} \E_{\mu}[T_k(T)]
    + \sum_{k \in \Mbest} \E_{\mu}[C_{k}(T)]
    \Bigr).
  \end{equation}
  \vspace*{-5pt}  % XXX
\end{lemma}

This result can also be used to recover Proposition 1 from \cite{Anandkumar11}, giving an upper bound on the regret that only depends on
the \emph{expected number of sub-optimal selections} -- $\E_{\mu}[T_k(T)]$ for $k \in \Mworst$ --
and the \emph{expected number of colliding players on the optimal arms} -- $\E_{\mu}[\cC_k(T)]$ for $k \in \Mbest$. Note that, in term (c) the number of colliding players on the sub-optimal arm $k$ may be upper bounded as $\E_{\mu}[\cC_k(T)] \leq M \E_{\mu}[T_k(T)]$.

In the next Section, we present an algorithm that has a logarithmic regret,
while ensuring that the number of sub-optimal selections is matching the lower bound of Theorem~\ref{thm:BetterLowerBound}.

% -----------------------------------------------------------------
% -----------------------------------------------------------------
\section{New Algorithms for Multi-Player Bandits}
\label{sec:algorithms}

When sensing is possible, that is under observation models \modelun{} and \modeldeux, most existing strategies build on a \emph{single-player bandit algorithm} (usually an \emph{index policy}) that relies on the sensing information, together with an \emph{orthogonalization strategy} to deal with collisions. We present this approach in more details in Section~\ref{sub:RandTopM_and_MCTopM} and introduce two new algorithms of this kind, \RandTopM{} and \MCTopM.
Then, we suggest in Section~\ref{sub:Selfish} a completely different approach, called \Selfish, that no longer requires an orthogonalization strategy as the collisions are directly accounted for in the indices that are used.
\Selfish{} can also be used under observation model \modeltrois{} --\emph{without sensing}--, and without the knowledge of $M$.

% -----------------------------------------------------------------

\subsection{Two New Strategies Based on Indices and Orthogonalization: \RandTopM{} and \MCTopM}
\label{sub:RandTopM_and_MCTopM}

% first, what is an index policy

In a single-player setting, \emph{index policies} are popular bandit algorithms: at each round one index is computed for each arm, that only depends on the history of plays of this arm and (possibly) some exogenous randomness. Then, the arm with highest index is selected. This class of algorithms includes the UCB family, in which the index of each arm is an Upper Confidence Bound for its mean, but also some Bayesian algorithms like Bayes-UCB \citep{Kaufmann12BUCB} or the randomized Thompson Sampling algorithm \citep{Thompson33,AgrawalGoyal11,Kaufmann12Thompson}.

% concrete examples and how their are used within MPB

The approaches we now describe for multi-player bandits can be used in combination with any index policy, but we restrict our presentation to UCB algorithms, for which strong theoretical guarantees can be obtained. In particular, we focus on two types of indices:
\UCB{} indices \citep{Auer02}
and \klUCB{} indices \citep{KLUCBJournal}, that can be defined for each player $j$ in the following way.
Letting $S_k^j(t) := \sum_{s=1}^t Y_{k,s} \indic(A^j(t) = k)$ the current sum of sensing information obtained by player $j$ for arm $k$, $\widehat{\mu}_k^j(t) = S_k^j(t)/T_k^j(t)$ (if $T_k^j(t)\neq 0$) is the empirical mean of arm $k$ for player $j$ and one can define the index
\begin{equation}\label{eq:indexFor_UCB_klUCB}
  g_k^j(t) := \begin{cases}
      \widehat{\mu}_k^j(t)  + \sqrt{  f(t) / (2T_k^j(t))}
      &\text{for } \UCB, \\
      \sup\left\{ q \in [0, 1]: T_k^j(t)\times \kl(\widehat{\mu}_k^j(t), q) \leq f(t) \right\}
      &\text{for } \klUCB,
  \end{cases}
\end{equation}
where $f(t)$ is some \emph{exploration function}. $f(t)$ is usually taken to be $\log(t)$ in practice, and slightly larger in theory, which ensures that  $\Pr(g_k^j(t) \geq \mu_k) \gtrsim 1 - 1/t$ (see \cite{KLUCBJournal}).
A classical (single-player) UCB algorithm aims at the arm with largest index. However, if each of the $M$ players selects the arm with largest UCB, all the players will end up colliding most of the time on the best arm.
To circumvent this problem, several coordination mechanisms have emerged, that rely on \emph{ordering} the indices and targeting \emph{one of} the $M$-best indices.

% RhoRand

While the \TDFS{} algorithm \citep{Zhao10} relies on the player agreeing in advance on the time steps at which they will target each of the $M$ best indices  (even though some alternative without pre-agreement are proposed),
the \rhoRand{} algorithm \citep{Anandkumar11} relies on randomly selected \emph{ranks}. %\footnote{\rhoRand{} only works if $M$ is known, and \cite{Anandkumar11} extended it to the case of an unknown $U$, with \rhoRandEst. Extending our algorithms for the case of unknown number of players $M$ is an interesting future work.}.
More formally, letting $\pi(k,\mathbf{g})$ be the index of the $k$-th largest entry in a vector $\mathbf{g}$,
% (for $k\in\{1,\dots,K\}$),
in \rhoRand{} each player maintains at time $t$ an internal rank $R^j(t)\in\{1,\dots,M\}$
and selects at time $t$,
\begin{equation}
  A^j(t) := \pi\left(R^j(t), [g^j_\ell(t)]_{\ell=1,\dots,K}\right).
\end{equation}
If a collision occurs, a new rank is drawn uniformly at random: $R^j(t+1) \sim \cU(\{1,\dots,M\})$.

% Now, our algorithms !!

We now propose two alternatives to this strategy, that do not rely on ranks and rather randomly fix themselves on one \emph{arm} in $\TopM(t)$, that is defined as the set of arms that have the $M$ largest indices:
\begin{equation}
  \TopM(t) := \left\{ \pi\left(k, \{g^j_\ell(t)\}_{\ell=1,\dots,K}\right), k=1,\dots,M\right\}.
\end{equation}

Our proposal \MCTopM{} is stated below as Algorithm~\ref{algo:MCTopM},
while a simpler variant, called \RandTopM,
is stated as Algorithm~\ref{algo:RandTopM} in Appendix~\ref{app:RandTopM}.
We focus on \MCTopM{} as it is easier to analyze and performs better.
Both algorithms ensure that player $j$ always
selects at time $t+1$ an arm from $\TopM(t)$. When a collision occurs \RandTopM{} randomly switches arm within $\TopM$, while \MCTopM{} uses a more sophisticated mechanism, that is reminiscent of ``Musical Chair'' (MC) and inspired by the work of \cite{Rosenski16}: players tend to fix themselves on arms (``chairs'') and ignore future collision when this happens.

% \vspace*{-5pt}  % XXX remove if problem
\begin{small}  % XXX remove if problem
  \begin{figure}[!ht]
      % \begin{framed}  % XXX remove if problem
      \begin{small}  % XXX remove if problem
      \centering
      % Documentation at http://mirror.ctan.org/tex-archive/macros/latex/contrib/algorithm2e/doc/algorithm2e.pdf if needed
      % Or https://en.wikibooks.org/wiki/LaTeX/Algorithms#Typesetting_using_the_algorithm2e_package
      % \removelatexerror% Nullify \@latex@error % Cf. http://tex.stackexchange.com/a/82272/
      \begin{algorithm}[H]
          % XXX Options
          % \LinesNumbered  % XXX Option to number the line
          % \RestyleAlgo{boxed}
          % XXX Input, data and output
          % \KwIn{$K$ and policy $P^j$ for arms set $\{1,\dots,K\}$\;}
          % \KwData{Data}
          % \KwResult{Result}
          % XXX Algorithm
              Let $A^j(1) \sim \cU(\{1,\dots,K\})$ and $C^j(1)=\mathrm{False}$ and $s^j(1)=\mathrm{False}$ \\
              \For{$t = 0, \dots, T-1$}{
                   \uIf(\tcp*[f]{transition $(3)$ or $(5)$}){
                      $A^j(t) \notin \TopM(t)$}
                    {
                      $A^j(t+1) \sim \cU \left(\TopM(t) \cap \left\{k : g_k^j(t-1) \leq g^j_{A^j(t)}(t-1)\right\}\right)$
                      \tcp*[f]{not empty} \\
                      % \tcp*[f]{randomly switch on an arm that had smaller UCB at $t-1$}
                      $s^j(t+1) = \mathrm{False}$
                      \tcp*[f]{aim at an arm with a smaller UCB at $t-1$}
                    }
                    \uElseIf(\tcp*[f]{collision and not fixed}){
                        $C^j(t)$ \emph{and} $\overline{s^j(t)}$}
                      {
                        $A^j(t+1) \sim \cU \left(\TopM(t)\right)$
                        \tcp*[f]{transition $(2)$} \\
                        $s^j(t+1) = \mathrm{False}$
                    }
                    \Else(\tcp*[f]{transition $(1)$ or $(4)$}){
                      $A^j(t+1) = A^j(t)$ \tcp*[f]{stay on the previous arm} \\
                      $s^j(t+1) = \mathrm{True}$ \tcp*[f]{become or stay fixed on a ``chair''}
                    }
                  Play arm $A^j(t+1)$, get new observations (sensing and collision), \\
                  Compute the indices $g^j_k(t+1)$ and set $\TopM(t+1)$ for next step.
              }
              \caption{The \MCTopM{} decentralized learning policy (for a fixed underlying index policy $g^j$).}
          \label{algo:MCTopM}
      \end{algorithm}
      \end{small}  % XXX remove if problem
      % \end{framed}  % XXX remove if problem
  \end{figure}
\end{small}  % XXX remove if problem
\vspace*{-5pt}  % XXX remove if problem

More precisely, under \MCTopM,
% \footnote{This choice is similar to the elementary step used in the \MusicalChair{} algorithm introduced by \cite{Rosenski16}, that gave its name to \MCTopM, even if they restrict to using the empirical averages as indices (\ie, the $0$-greedy algorithm).}
if player $j$ did not encounter a collision when using arm $k$ at time $t$,
then she marks her current arm as a ``chair'' ($s^j(t+1)=\mathrm{True}$),
and will keep using it even if collisions happen in the future (Lines~$9$-$11$).
As soon as this ``chair'' $k$ is no longer in $\widehat{M_j}(t)$,
a new arm is sampled uniformly from a subset of $\TopM(t)$,
defined with the previous indices $g^j(t-1)$ (Lines~$3$-$5$).
The subset enforces a certain inequality on indices,
$g_{k'}^j(t-1) \leq g^j_{k}(t-1)$ and $g_{k'}^j(t) \geq g^j_{k}(t)$,
when switching from $k=A^j(t)$ to $k'=A^j(t+1)$.
This helps to control the number of such changes of arm,
as shown in Lemma~\ref{lem:elementaryLemma_RandTopM_MCTopM}.
% the Appendix~\ref{proof:collisionsMCTopM}
The considered subset is never empty as it contains
at least the arm replacing the $k\in\TopM(t-1)$ in $\TopM(t)$.
Collisions are dealt with only for non-fixed player $j$,
and when the previous arm is still in $\TopM(t)$.
In this case, a new arm is sampled uniformly from $\TopM(t)$ (Lines~$6$-$8$).
This stationary aspect helps to minimize the number of collisions,
as well as the number of switches of arm.
The five different transitions $(1)$, $(2)$, $(3)$, $(4)$, $(5)$ refer to the notations used in the analysis of \MCTopM{} (see Figure~\ref{fig:StateMachineAlgorithm_MCTopM} in Appendix~\ref{proof:collisionsMCTopM}).

% -----------------------------------------------------------------
\subsection{The \Selfish{} Approach}
\label{sub:Selfish}

Under observation model \modeltrois{} no sensing information is available and the previous algorithms cannot be used, as the sum of sensing information $S_k^j(t)$ and thus the empirical mean $\widehat{\mu}_k^j(t)$ cannot be computed, hence neither the indices $g_k^j(t)$. However, one can still define a notion of \emph{empirical reward} received from arm $k$ by player $j$, by introducing
\vspace*{-5pt}
\begin{equation}
  \widetilde{S_k}^j(t) = \sum_{t=1}^T r^j(t) \indic(A^j(t) = k)
  \ \ \ \text{and letting} \ \ \ \widetilde{\mu_k}^j(t) := \widetilde{S_k}^j(t) \;/\; T_k^j(t).
\end{equation}

Note that $\widetilde{\mu_k}^j(t)$ is no longer meant to be an unbiased estimate of $\mu_k$ as it also takes into account the collision information, that is present in the reward. Based on this empirical reward, one can similarly defined modified indices as
\begin{equation}\label{eq:indexTilde}
  \widetilde{g_k}^j(t) = \begin{cases}
      \widetilde{\mu_k}^j(t)  + \sqrt{  f(t) / (2T_k^j(t))}
      &\text{for } \UCB, \\
      \sup\left\{ q \in [0, 1]: T_k^j(t)\times \kl(\widetilde{\mu_k}^j(t), q) \leq f(t) \right\}
      &\text{for } \klUCB.
  \end{cases}
\end{equation}

Given any of these two index policies (\UCB{} or \klUCB), the \Selfish{} algorithm is defined by,
\begin{equation}
  A^j(t) = \argmax{ k \in \{1,\dots,K\}} \ \widetilde{g_k}^j(t-1).\label{algo:Selfish}
\end{equation}
The name comes from the fact that each player is targeting, in a ``selfish'' way, the arm that has the highest index, instead of accepting to target only one of the $M$ best.
The reason that this may work precisely comes from the fact that $\widetilde{g_k}^j(t)$ is no longer an upper-confidence on $\mu_k$,
but some hybrid index that simultaneously increases when a transmission occurs and decreases when a collision occurs.

This behavior is easier to be understood for the case of \Selfish-\UCB{} in which, letting $N_k^{j,C}(t) = \sum_{s=1}^t \indic(C^j(t))$ be the number of collisions on arm $k$, one can show that the hybrid \Selfish{} index induces a penalty proportional to the fraction of collision on this arm and the quality of the arm itself:
\begin{equation}
  \widetilde{g_k}^j(t) = g_k^j(t) -
  \underbrace{\left(\frac{N_k^{j,C}(t)}{N_k^j(t)}\right)}_{\text{fraction of collisions}}
  \underbrace{\left(\frac{1}{N_k^{j,C}(t)}\sum_{t=1}^{T}Y_{A^j(t),t} \indic(C^j(t)) \indic(A^j(t) = k)\right)}_{\text{estimate of } \mu_k }.
\end{equation}

From a bandit perspective, it looks like each player is using a stochastic bandit algorithm (\UCB{} or \klUCB) when interacting with $K$ arms that give a feedback (the reward, and not the sensing information) that is far from being \iid{} from some distribution, due to the collisions.
As such, the algorithm does not appear to be well justified, and one may rather want to use adversarial bandit algorithms like $\mathrm{EXP3}$ \citep{Auer02EXP3}, that do not require a stochastic (\iid) assumption on arms.
However, we found out empirically that \Selfish{} is doing surprisingly well, as already noted by \cite{Bonnefoi17}, who did some experiments in the context of IoT applications.
We show in Section~\ref{sec:upperbounds} that \Selfish{} does have a (very) small probability to fail (badly), for some problem with small $K$,
which precludes the possibility of a logarithmic regret for any problem.
However, in most cases it empirically performs similarly to all the algorithms described before,
and usually outperforms \rhoRand,
even if it neither exploits the sensing information, nor the knowledge of the number of players $M$.
As such, practitioners may still be interested by the algorithm, especially for Cognitive Radio applications in which sensing is hard or not considered.

% -----------------------------------------------------------------

% -----------------------------------------------------------------
% -----------------------------------------------------------------
\section{Empirical performances}
\label{sec:experiments}

% Select one problems, max two cases ($M=K$ and $M < K$), and illustrate what we want to discuss in plots with regret in normal scale, one with semi-$\log x$ scale, and at least one histogram showing the bad luck for \Selfish...

% Selfish is awesome... unless $K=M=2$! Not only, we found other case of failure.

% Chose one problem $\boldsymbol{\mu}$, and vary $M \leq K$ for let say $K=3$ and $K=9$. Not more figures.
% Put most of them in appendix.

We illustrate here the empirical performances of the algorithms presented in Section~\ref{sec:algorithms}, used in combination with the \klUCB{} indices. %used with two index policies, \UCB{} and \klUCB.
Some plots are at pages \pageref{fig:MP__K9_M6_T5000_N500__4_algos__all_RegretCentralized__BayesianProblems} and \pageref{fig:MP__K9_M6_T10000_N1000__4_algos} and most of them in Appendix~\ref{app:plotsFromSec5}.
% but observations and conclusions from the experiments are discussed here.

In experiments that are not reported here, we could observe that using \klUCB{} rather than \UCB{} indices always yield better practical performance.
%, as the theoretical guarantees for single player suggested,
As the prupose of this work is not to optimize on the index policy, but rather propose new ways of using indices in a decentralized setting,
% And for some configurations, using a decentralized approach with a more efficient index family (\klUCB{} instead of \UCB) has better performances than the centralized approach.
we only report results for \klUCB.
In a first set of experiments, \MCTopM, \RandTopM{} and \Selfish{} are benchmarked against the state-of-the-art \RhoRand{} algorithm.
We also include a centralized multiple-play \klUCB{} algorithm,
%as defined by \cite{Anantharam87a},
essentially to check that the \emph{``price of decentralized learning''} is not too large.

\begin{figure}[!ht]
  \centering
  \begin{subfigure}[!ht]{0.85\textwidth}
    \includegraphics[width=1.00\textwidth]{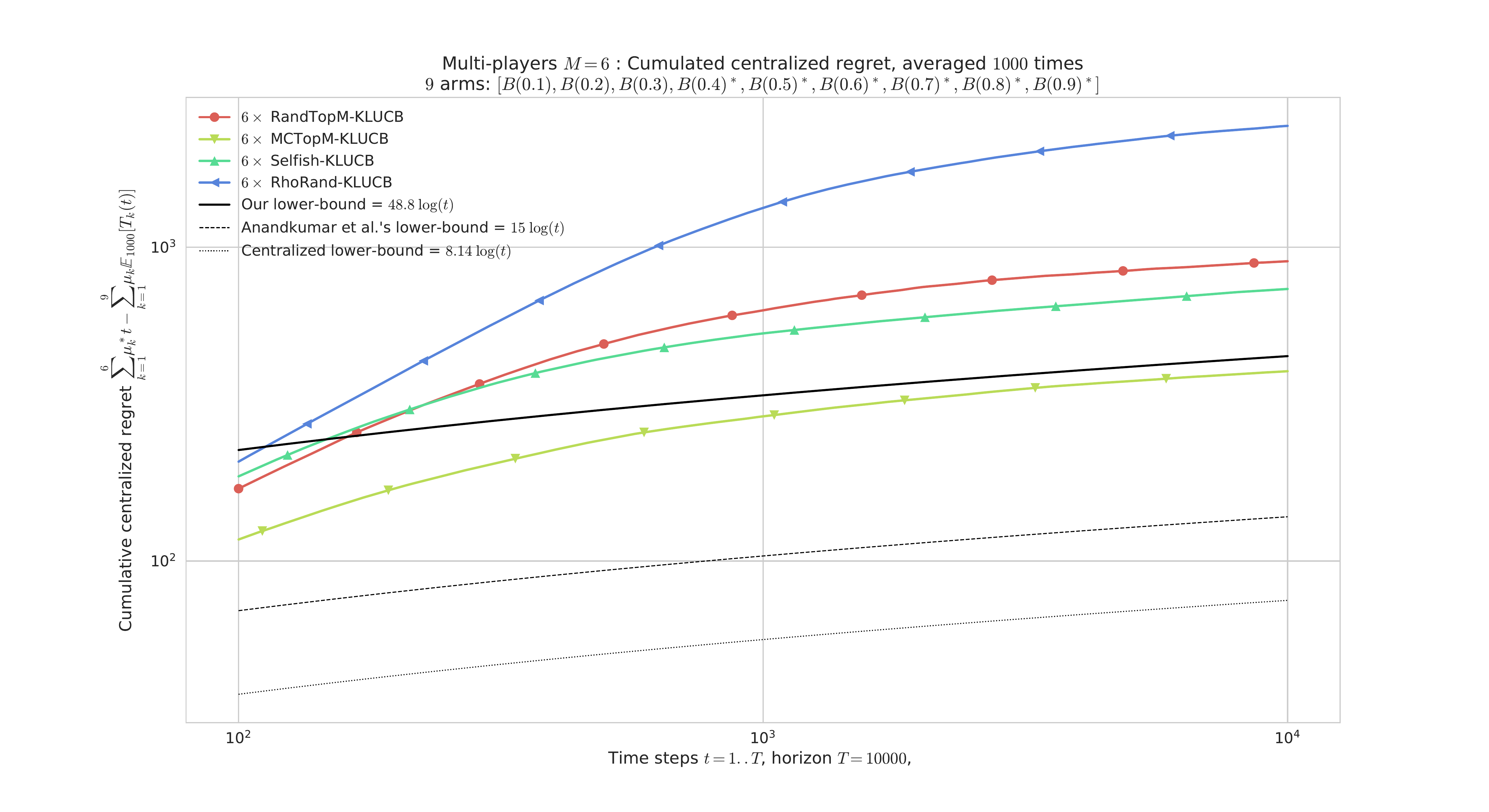}
  \end{subfigure}
  ~
  \begin{subfigure}[!ht]{0.85\textwidth}
    \includegraphics[width=1.00\textwidth]{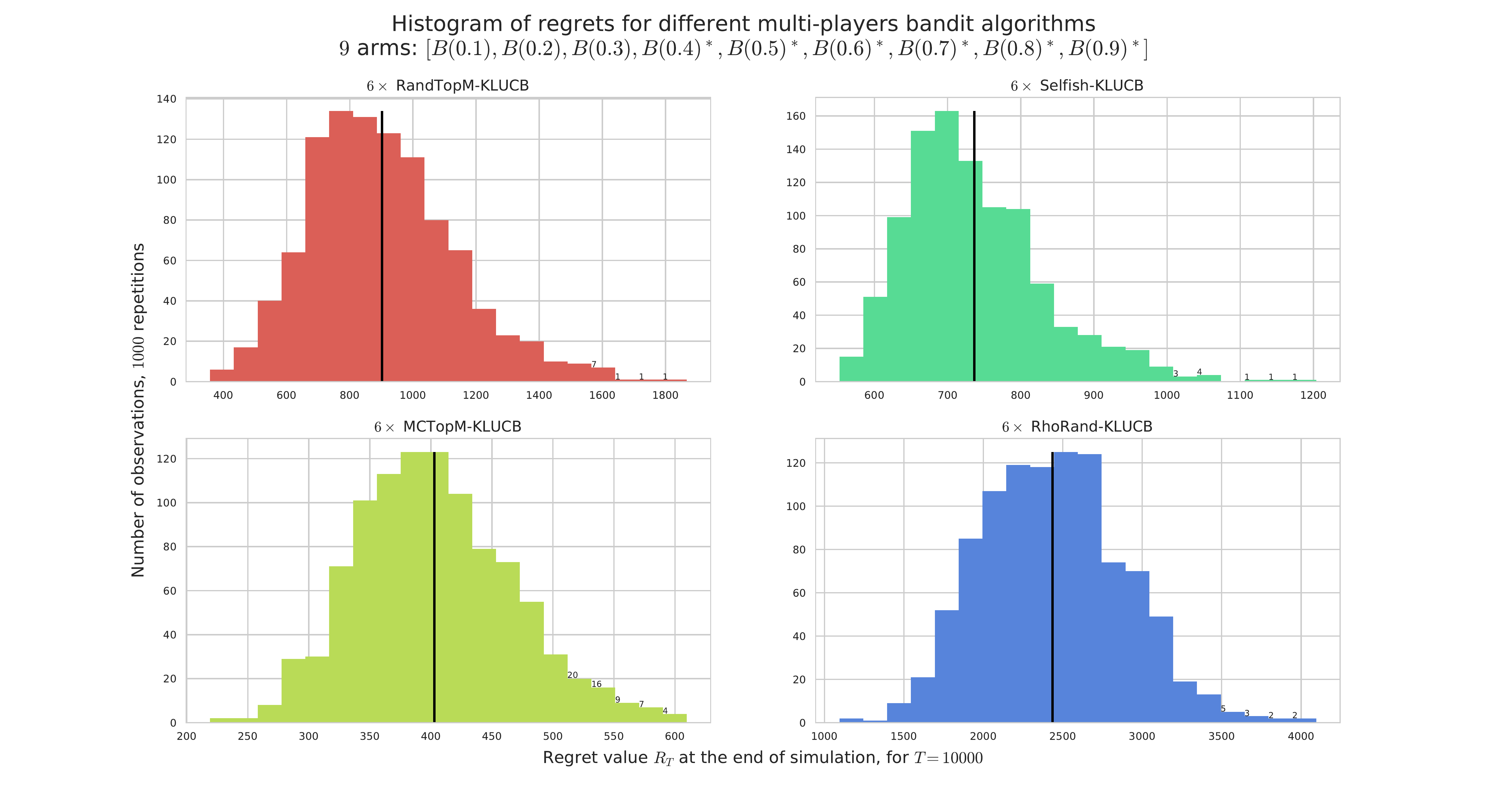}
  \end{subfigure}
  \caption{Regret (in $\log\log$ scale), for $M=6$ players for $K=9$ arms, horizon $T=5000$, for $1000$ repetitions on problem $\boldsymbol{\mu}=[0.1,\dots,0.9]$. \RandTopM{} (\textcolor{gold}{yellow} curve) outperforms \Selfish{} (\textcolor{darkgreen}{green}), both clearly outperform \rhoRand. The regret of \MCTopM{} is logarithmic, empirically with the same slope as the lower bound. The $x$ axis on the regret histograms have different scale for each algorithm.}
  \label{fig:MP__K9_M6_T10000_N1000__4_algos}
\end{figure}

We present results for two bandit instance: one with $K=3$ arms and means %\footnote{But of course the means are unknown to the algorithms, and their order is not important.}
$\boldsymbol{\mu} = [0.1, 0.5, 0.9]$, for which two cases $M=2$ and $M=3$ are presented in Figure~\ref{fig:selfish_fail1}. For the second instance $K=9$ and $\boldsymbol{\mu} = [0.1, 0.2, \dots, 0.9]$,
three cases are presented: $M=6$ in Figure~\ref{fig:MP__K9_M6_T10000_N1000__4_algos},
and for the two limit cases $M=2$ and $M=9=K$ in Figure~\ref{fig:MP__K9_M2-6-9_T10000_N200__4_algos}.
% Note the different algorithms know the number of players $M$.
Performance is measured with the \emph{expected} regret up to horizon $T=10000$, estimated based on $1000$ repetitions on the same bandit instance. %, from $t=1$ to the horizon $T=10000$ (also unknown by the algorithms),
We also include histograms showing the \emph{distribution} of regret at $t=T$,
as this allows to check if the regret is indeed small for \emph{each} run of the simulation.
For the plots showing the regret, our \emph{asymptotic} lower bound from Theorem~\ref{thm:BetterLowerBound} is displayed.
% and it should be surprising to see it being larger than some regret values: it is only asymptotic.

\begin{figure}[!ht]
  \centering
  % \begin{subfigure}[!ht]{0.49\textwidth}
  %   \includegraphics[width=1.10\textwidth]{all_RegretCentralized____env1-1_3251433209347345969.pdf}
  % \end{subfigure}
  % % ~
  % \begin{subfigure}[!ht]{0.49\textwidth}
    \includegraphics[width=0.8\textwidth]{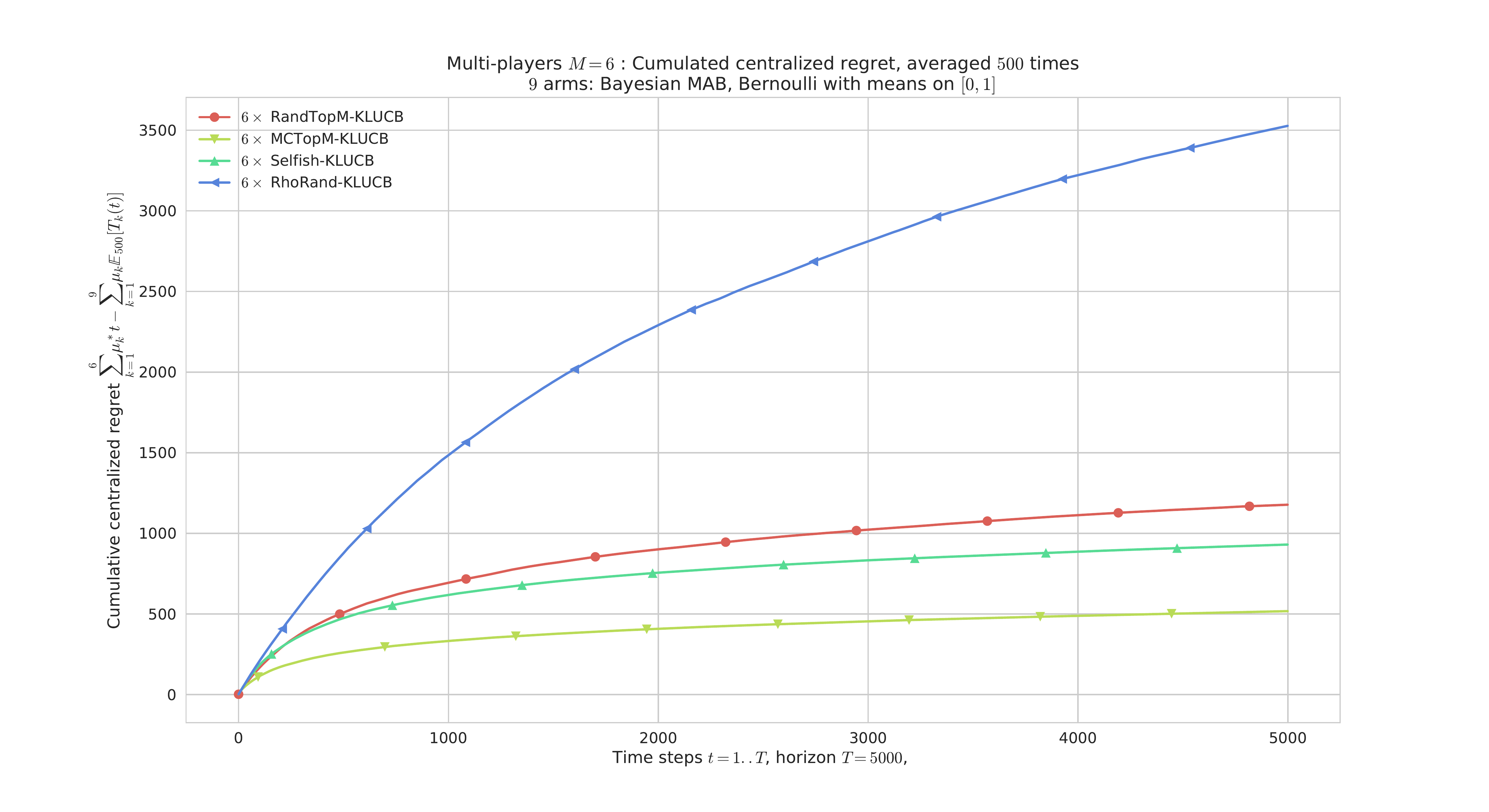}
  % \end{subfigure}
  \caption{Regret, $M=6$ players, $K=9$ arms, horizon $T=5000$, against $500$ problems $\boldsymbol{\mu}$ uniformly sampled in $[0,1]^K$. \rhoRand{} (top \textcolor{blue}{blue} curve) is outperformed by the other algorithms (and the gain increases with $M$). \MCTopM{} (bottom \textcolor{gold}{yellow}) outperforms all the other algorithms is most cases.}
  % \label{fig:MP__K9_M2-6_T5000_N500__4_algos__all_RegretCentralized__BayesianProblems}
  \label{fig:MP__K9_M6_T5000_N500__4_algos__all_RegretCentralized__BayesianProblems}
  % \vspace*{-15pt}  % XXX remove if problem
\end{figure}

Experiments with a different problem for each repetition (uniformly sampled $\boldsymbol{\mu} \sim \cU([0,1]^K)$),
are also considered, in Figure~\ref{fig:MP__K9_M6_T5000_N500__4_algos__all_RegretCentralized__BayesianProblems} and \ref{fig:MP__K9_M2_T5000_N500__4_algos__all_RegretCentralized__BayesianProblems}.
This helps to check that no matter the \emph{complexity} of the considered problem (one measure of complexity being the constant in our lower bound),
\MCTopM{} performs similarly or better than all the other algorithms,
and \Selfish{} outperforms \rhoRand{} in most cases.
%Figure~\ref{fig:MP__K9_M6_T5000_N500__4_algos__all_RegretCentralized__BayesianProblems} is a good example
%of outstanding performances of \MCTopM{} and \Selfish{} in comparison to \rhoRand.
Empirically, our proposals were found to almost always outperform \rhoRand, and except for \Selfish{} that can fail badly on problems with small $K$,
we verified that \MCTopM{} outperforms the state-of-the-art algorithms in many different problems, and is more and more efficient as $M$ and $K$ grows.

%
% Regular plots of centralized regrets
%

% Dynamic settings, when $M$ can change in time,
% were considered in the analysis of both
In the presence of sensing (observation model \modeldeux), we also compared our algorithms to  %multi-player multi-armed bandits algorithms
with \MEGA{} \citep{Avner15} and \MusicalChair{} \citep{Rosenski16}. Yet these two algorithms were found hard to use efficiently in practice and we show in
% that were also proposed
%
Figure~\ref{fig:MP__K9_M3_T123456_N100__8_algos} that they perform poorly in comparison to \rhoRand, \RandTopM{} and \MCTopM.
\MEGA{} needs a careful tuning of \emph{five} parameters ($c$, $d$, $p_0$, $\alpha$ and $\beta$) to attain reasonable performances. No good guideline for tuning them is provided and using cross validation, as suggested,
can be considered out of the scope of online sequential learning.
%In practice, on a fixed instance, the authors do not indicate how to select the parameters, even with a perfect knowledge of the parameters ($\boldsymbol{\mu}$ and $T$).
%
\MusicalChair{} consists of an random exploration phase of length $T_0$ after which the players quickly converge to orthogonal strategies targeting the $M$ best arms. With probability $1-\delta$, its regret is proved to be ``constant'' (of order $\log(1/\delta)$). The theoretical minimal value for $T_0$ depends on $\delta$, on the horizon $T$ and on a lower bound $\epsilon$ on the gap $\Delta = \mu^*_M - \mu^*_{M+1}$, and the practical tuning is hard too. %% which are both unavailable in our setting.

% -----------------------------------------------------------------
% -----------------------------------------------------------------
\section{Theoretical elements}
\label{sec:upperbounds}

Section~\ref{sub:UpperBoundSelections} gives
an asymptotically optimal analysis of the expected number of sub-optimal draws
for \RandTopM, \MCTopM{} and \rhoRand{} combined with \klUCB{} indices,
and Section~\ref{sub:UpperBoundCollisions} proves that the number of  collisions, hence the regret of \MCTopM{} are also logarithmic.
Section~\ref{sub:SelfishFails} shortly discusses a disappointing result regarding \Selfish, with more insights provided in Appendix~\ref{app:SelfishFails}.

% -----------------------------------------------------------------
\subsection{Common Analysis for \RandTopM- and \MCTopM-\klUCB{}}\label{sub:UpperBoundSelections}

Lemma~\ref{lem:SubOptimalSelections} gives a finite-time upper bound on the expected number of draws of a sub-optimal arm $k$ for any player $j$, that
holds for both \RandTopM-\klUCB{} and \MCTopM-\klUCB.
Our improved analysis also applies to \rhoRand.
Explicit expressions for $C_{\boldsymbol{\mu}}$, $D_{\boldsymbol{\mu}}$ can be found in the proof given in Appendix~\ref{proof:SubOptimalSelections}.

\begin{lemma}\label{lem:SubOptimalSelections}
  For any $\boldsymbol{\mu}\in\cP_M$,
  if player $j\in\{1,\dots,M\}$ uses the \RandTopM-, \MCTopM- or \rhoRand-\klUCB{}
  decentralized policy with exploration function $f(t) = \log(t) + 3 \log\log(t)$,
  then for any sub-optimal arm $k \in \Mworst$, there exists two problem depend constants $C_{\boldsymbol{\mu}}$, $D_{\boldsymbol{\mu}}$ such that
  % %
  % \hfill{}
  % (Proved in Appendix~\ref{proof:SubOptimalSelections})
  \begin{equation}\label{eq:SubOptimalSelections}
      \E_{\mu}[T_k^j(T)] \leq \frac{\log(T)}{\kl(\mu_k,\mu_{M}^*)}+ C_{\boldsymbol{\mu}} \sqrt{\log(T)} + D_{\boldsymbol{\mu}}\log\log(T) + 3M + 1.
  \end{equation}
\end{lemma}

It is important to notice that the leading constant in front of $\log(T)$ is the same as in the constant featured in Equation~\eqref{eq:LBDraws} of Theorem~\ref{thm:BetterLowerBound}. This result proves that the lower bound on sub-optimal selections is asymptotically matched for the three considered algorithms. This is a strong improvement in comparison to the previous state-of-the-art results
\citep{Zhao10,Anandkumar11}.

% -----------------------------------------------------------------

% For \rhoRand-\klUCB{} we used results from \cite{Anandkumar11}
% to control the number of collisions on sub-optimal arms.
% %
% For \RandTopM-\klUCB{} we have a better control over the number of collisions,
% and as expected we can bound it by a logarithmic term with a much smaller constant.
% The \RandTopM{} policy was indeed designed to be more ``conservative'' in its dynamics
% in order to reduce the number of collisions caused by a player who changes from his current chosen arm
% in an orthogonal configuration.

% For \rhoRand, \cite{Anandkumar11} essentially proved on the one hand that conditionnaly to a certain ``good'' event
% (all players have a correct ordering of all the $M$ best arms)
% proved to happen with high probability,
% the total number of collisions is upper bounded asymptotically by ${2 M - 1 \choose M}$,
% which is constant but grows very quickly as $M$ grows.
% And on the other hand, they proved that the total number of steps when this event is violated is $\bigO{\log T}$, with large constants in the asymptotic notation.

As announced, Lemma~\ref{lem:elementaryLemma_RandTopM_MCTopM} controls
the number of switches of arm that are due to the current arm leaving $\TopM(t)$,
for both \RandTopM{} and \MCTopM. It essentially proves that Lines~$3$-$5$ in Algorithm~\ref{algo:MCTopM} (when a new arm is sampled from the non-empty subset of $\TopM(t)$)
happen a logarithmic number of times. The proof of this result is given in Appendix~\ref{proof:elementaryLemma_RandTopM_MCTopM}.

\begin{lemma}\label{lem:elementaryLemma_RandTopM_MCTopM}
  For any $\boldsymbol{\mu}\in\cP_M$,
  any player $j \in \{1, \dots, M\}$ using
  \RandTopM- or \MCTopM-\klUCB,
  and any arm $k$,
  it holds that
  \begin{equation*}
    \sum_{t=1}^T
    \Pr\left(A^j(t)=k, k\notin \TopM(t)\right)
    = \left(\sum_{k', \mu_{k'} < \mu_k}\frac{1}{\kl(\mu_k,\mu_{k'})} + \sum_{k', \mu_{k'} > \mu_k}\frac{1}{\kl(\mu_{k'},\mu_{k})}\right) \log(T) + o(\log(T)).
  \end{equation*}
\end{lemma}

% -----------------------------------------------------------------
\subsection{Regret Analysis of \MCTopM-\klUCB}\label{sub:UpperBoundCollisions}

For \MCTopM, we are furthermore able to obtain a logarithmic regret upper bound, by proposing an original approach to control the number of collisions under this algorithm.
First, we can bound the number of collisions by the number of collisions for players not yet ``fixed on their arms'' ($\overline{s^j(t)}$),
that we can then bound by the number of changes of arms
(\cf{} proof in Appendix~\ref{proof:collisionsMCTopM}).
An interesting consequence of the proof of this result is that
it also bounds the number of \emph{switches of arms}, $\sum_{t=1}^T \Pr(A^j(t+1) \neq A^j(t))$,
and this additional guarantee was never clearly stated for previous state-of-the-art works, like \rhoRand.
Even though minimizing switching was not a goal\footnote{Introducing \emph{switching costs}, like it was done in previous works, \eg, \cite{Koren17}, could be an interesting future work.},
this guarantee is interesting for Cognitive Radio applications,
where switching arms means reconfiguring a radio hardware, an operation that costs energy.
%and thus an algorithm guaranting a small number of switches is interesting.

\begin{lemma}\label{lem:collisionsMCTopM}
  For any $\boldsymbol{\mu}\in\cP_M$,
  if all players use the
  \MCTopM-\klUCB{} decentralized policy,
  and $M \leq K$,
  then the total average number of collisions (on all arms)
  is upper-bounded by
  \hfill{}
  (Proved in Appendix~\ref{proof:collisionsMCTopM})
  \begin{equation}
    \E_{\mu}\left[\sum_{k=1}^K \cC_k(T)\right]
    \leq M^2\left(2 M + 1\right) \left(\sum_{a,b=1,\dots,K,\;\mu_a < \mu_b} \frac{1}{\kl(\mu_a,\mu_b)}\right) \log(T) + \smallO{\log T}.
  \end{equation}
\end{lemma}

% https://tex.stackexchange.com/questions/19587/why-does-amsmath-print-a-warning-about-a-foreign-command#comment37128_19592
Note that this bound is in $\bigO{M^3}$,
which significantly improves the $\bigO{M  \binom{2M-1}{M}}$ proved by \cite{Anandkumar11} for \rhoRand. It is worse than the $\bigO{M^2}$ proved by \cite{Rosenski16} for \MusicalChair. %, due to our trick of focusing on collisions for non-sitted players.
However, unlike \MusicalChair{} our algorithm does not need the knowledge of $\mu^*_{M}-\mu^*_{M+1}$.

% -----------------------------------------------------------------
% \subsection{Logarithmic Regret for \MCTopM}\label{sub:Regret}

Now that the sub-optimal arms selections and the collisions
are both proved to be at most logarithmic in Lemmas~\ref{lem:SubOptimalSelections} and \ref{lem:collisionsMCTopM},
it follows from our regret decomposition (Lemma~\ref{lem:DecompositionRegret}) together with Lemma~\ref{lem:1stUpperBound} that the regret of \MCTopM-\klUCB{} is logarithmic. More precisely, one obtains a finite-time problem-depend upper bound on the regret of this algorithm.

\begin{theorem}\label{thm:LogarithmicRegret_MCTopMklUCB}
  If all $M$ players use
  \MCTopM-\klUCB, and $M \leq K$,
  then for any problem $\boldsymbol{\mu} \in \cP_M$,
  there exists a problem dependent constant $G_{M,\boldsymbol{\mu}}$, such that
  the regret satisfies:
  \begin{equation}\label{eq:LogarithmicRegret_MCTopMklUCB}
    R_T(\boldsymbol{\mu}, M, \rho) \leq G_{M,\boldsymbol{\mu}} \log(T) + \smallO{\log T}.
  \end{equation}
\end{theorem}

% -----------------------------------------------------------------
\subsection{Discussion on \Selfish} \label{sub:SelfishFails}

The analysis of \Selfish{} is harder, but we tried our best to obtain some understanding of the behavior of this algorithm, that seems to be doing surprisingly well in many contexts, as in our experiments with $K=9$ arms and in extensive experiments not reported in this paper. However, a disappointing result is that we found simple problems, usually with small number of arms, for which the algorithm may fail. For example with $M=2$ or $M=3$ players competing for $K=3$ arms,
with means $\boldsymbol{\mu} = [0.1, 0.5, 0.9]$, the histograms in Figure~\ref{fig:selfish_fail1} suggests that with a small probability, the regret $R_T$ of \Selfish-\klUCB{} can be very large. We provide a discussion in Appendix~\ref{app:SelfishFails} about when such situations may happen, including a conjectured (constant, but small) lower bound on the probability that \Selfish{} experience collision almost at every round. This result would then prevent \Selfish{} from having a logarithmic regret. However, it is to be noted that the lower bound of Theorem~\ref{thm:BetterLowerBound} does not apply to the censored observation model \modeltrois{} under which \Selfish{} operates, and it is not known yet whether logarithmic regret is at all possible.

% -----------------------------------------------------------------
% -----------------------------------------------------------------
\section{Conclusion and future work}
\label{sec:conclusion}

To summarize, we presented three variants of Multi-Player Multi-Arm Bandits,
with different level of feedback being available to the decentralized players, under which we proposed efficient algorithms.
For the two easiest models --with sensing--, our theoretical contribution improves both the state-of-the-art upper and lower bounds on the regret. In the absence of sensing, we also provide some motivation for the practical use of the interesting \Selfish{} heuristic, a simple index policy based on hybrid indices that are directly taking  into account the collision information.

This work suggests several interesting further research directions. First, we want to investigate the notion of \emph{optimal algorithms} in the decentralized multi-player model with sensing information. So far we provided the first matching upper and lower bound on the expected number of sub-optimal arms selections, which suggests some form of (asymptotic) optimality. However, sub-optimal draws turn out not be the dominant terms in the regret, both in our upper bounds and in practice, thus an interesting future work is to identify some notion of \emph{minimal number of collisions}. Second, it remains an open question to know if a simple decentralized algorithm can be as efficient as \MCTopM{} without knowing $M$ in advance, or in dynamic settings (when $M$ can change in time). We shall start by proposing variants of our algorithm that are inspired by the \rhoRandEst{} variant of \rhoRand{} proposed by \cite{Anandkumar11}.
Finally, we want to strengthen the guarantees obtained in the absence of sensing, that is to know whether logarithmic regret is achievable and to have a better analysis of the \Selfish{} approach.  Indeed, in most cases, it performs comparably to \RandTopM{} even with limited feedback and without knowing the number of players $M$, which makes it a good candidate for applications to Internet of Things networks.

% Finally, to propose a better models for IoT applications, in the future we would be interested in a different regime in which $M$ can be very large in front of $K$, but every player to transmit at every round.

\newpage

\newpage

% -----------------------------------------------------------------
% Acknowledgements should go at the end, before appendices and references

% % FIXME really ?
% % Acknowledgments can be in the final version only

\vfill{}
% % \begin{small}
\textbf{Acknowledgements:}
  This work is supported by the French Agence Nationale de la Recherche (ANR), under grant ANR-16-CE40-0002 (project BADASS), by the CNRS under the PEPS project BIO, by the French Ministry of Higher Education and Research (MENESR) and ENS Paris-Saclay.
  Thanks to Christophe Moy, R{\'e}mi Bonnefoi and Vincent Gouldieff for useful discussions.
% % \end{small}


% -----------------------------------------------------------------

% biblio usually before the supplementary material
\bibliography{BK__ALT_2018}

% \newpage

% -----------------------------------------------------------------
\appendix

\hr{}

% -----------------------------------------------------------------
% -----------------------------------------------------------------
\section{Regret Decompositions}
\label{app:regretdec}

% -----------------------------------------------------------------
\subsection{Proof of Lemma~\ref{lem:DecompositionRegret}}
\label{proof:DecompositionRegret}

    Using the definition of regret $R_T$ from \eqref{eq:regret}, and this collision indicator $\eta^j(t):=\indic(\overline{C^j(t)})$,
    \begin{align*}
      R(T)
      &= \left(\sum_{k=1}^{M}\mu_k^*\right)T - \E_{\mu}\left[\sum_{t=1}^T\sum_{j=1}^M Y_{A^j(t),t} \eta^j(t) \right]
       = \left(\sum_{k=1}^{M}\mu_k^*\right)T - \E_{\mu}\left[\sum_{t=1}^T\sum_{j=1}^M \mu_{A^j(t)} \eta^j(t)\right]
      \shortintertext{The last equality comes from the linearity of expectations, and the fact that $\E_{\mu}[Y_{k,t}] = \mu_k$ (for all $t$, from the \iid{} hypothesis), and the independence from $A^j(t)$, $\eta^j(t)$ and $Y_{k,t}$ (observed \emph{after} playing $A^j(t)$). So $\E_{\mu}[Y_{A^j(t),t} \eta^j(t)] = \sum_{k} \E_{\mu}[\mu_k \indic(A^j(t),t) \eta^j(t)] = \E_{\mu}[\mu_{A^j(t)} \eta^j(t)]$. And so}
      R(T)
      &= \E_{\mu}\left[ \sum_{t=1}^{T}\sum_{j \in \Mbest} \mu_j
        - \sum_{t=1}^{T} \sum_{j=1}^{M} \mu_{A^j(t)} \eta^j(t) \right] \\
        % &= \underbrace{\left( \frac{1}{M} \sum_{j \in \Mbest} \mu_j \right)}_{:= \overline{\mu}^*} (T M)
      &= \left( \frac{1}{M} \sum_{j \in \Mbest} \mu_j \right)
        - \sum_{k=1}^{K} \sum_{j=1}^{M} \mu_k \E_{\mu}\left[ T^j_k(T) \right]
        + \sum_{k=1}^{K} \mu_k \E_{\mu}\left[ \cC_k(T) \right].
      \shortintertext{For the first term, we have $T M = \sum\limits_{k=1}^{K} \sum\limits_{j=1}^{M} \E_{\mu}\left[ T^j_k(T)\right]$, and if we denote $\overline{\mu}^* := \frac{1}{M} \sum\limits_{j \in \Mbest} \mu_j$ the average mean of the $M$-best arms, then,}
      &= \sum_{k=1}^{K} \sum_{j=1}^{M} (\overline{\mu}^* - \mu_k) \E_{\mu}\left[ T^j_k(T) \right]
        + \sum_{k=1}^{K} \mu_k \E_{\mu}\left[ \cC_k(T) \right].
      \shortintertext{Let $\overline{\Delta_k} := \overline{\mu}^* - \mu_k$ be the gap between the mean of the arm $k$ and the $M$-best average mean, and if $M^*$ denotes the index of the worst of the $M$-best arms (\ie, $M^* = \arg\min_{k\in\Mbest}(\mu_k)$), then by splitting $\{1,\dots,K\}$ into three disjoint sets $\Mbest \cupdot \Mworst = (\Mbest\setminus\{M^*\}) \cupdot \{M^*\} \cupdot \Mworst$, we get}
      &= \sum_{k \in \Mbest\setminus\{M\}} \overline{\Delta_k} \E_{\mu}\left[ T_k(T) \right]
        + \overline{\Delta_{M^*}} \E_{\mu}\left[ T_{M^*}(T) \right] \\
        &\;\;\;\;\;\;\;\; + \sum_{k \in \Mworst} \overline{\Delta_k} \E_{\mu}\left[ T_k(T) \right]
        + \sum_{k=1}^{K} \mu_k \E_{\mu}\left[ \cC_k(T) \right].
      \shortintertext{But for $k = M^*$, $T_{M^*}(T) = T M^* - \sum\limits_{k \in \Mbest\setminus\{M\}} \E_{\mu}\left[ T_k(T) \right] - \sum\limits_{k \in \Mworst} \E_{\mu}\left[ T_k(T) \right]$, so by recombining the terms, we obtain,}
      &= \sum_{k \in \Mbest\setminus\{M\}} (\overline{\Delta_k} - \overline{\Delta_{M^*}}) \E_{\mu}\left[ T_k(T) \right]
        + \overline{\Delta_{M^*}} T M^* \\
        &\;\;\;\;\;\;\;\; + \sum_{k \in \Mworst} (\overline{\Delta_k} - \overline{\Delta_{M^*}}) \E_{\mu}\left[ T_k(T) \right]
        + \sum_{k=1}^{K} \mu_k \E_{\mu}\left[ \cC_k(T) \right].
    \end{align*}
    The term $\overline{\Delta_k} - \overline{\Delta_{M^*}}$ simplifies to $\mu_{M^*} - \mu_k$, and so $\overline{\Delta_{M^*}} = \frac{1}{M} \sum_{k=1}^{M} \mu_k - \mu_{M^*}$ by definition of $\overline{\mu}^*$. And for $k=M^*$, $\mu_{M^*} - \mu_k = 0$, so the first sum can be written for $k = 1,\dots,M$ only, so
    %
    % \vspace*{5pt}
    \begin{align*}
      R(T)
      &= \sum_{k \in \Mbest} (\mu_{M^*} - \mu_k) \E_{\mu}\left[ T_k(T) \right]
        + \sum_{k \in \Mbest} (\mu_k - \mu_{M^*}) T \\
        &\;\;\;\;\;\;\;\; + \sum_{k \in \Mworst} (\mu_{M^*} - \mu_k) \E_{\mu}\left[ T_k(T) \right]
        + \sum_{k=1}^{K} \mu_k \E_{\mu}\left[ \cC_k(T) \right]
    \end{align*}
    And so we obtain the decomposition with three terms \termeun, \termedeux{} and \termetrois.
    \begin{align*}
    R(T)
      &= \sum_{k \in \Mbest} (\mu_k - \mu_{M^*}) \left(T - \E_{\mu}\left[ T_k(T) \right]\right)
        + \sum_{k \in \Mworst} (\mu_{M^*} - \mu_k) \E_{\mu}\left[ T_k(T) \right]
        + \sum_{k=1}^{K} \mu_k \E_{\mu}\left[ \cC_k(T) \right].
    \end{align*}
    % Which is exactly the decomposition we wanted to prove.

% -----------------------------------------------------------------
\subsection{Proof of Lemma~\ref{lem:1stLowerBound}}
\label{proof:1stLowerBound}

  Note that term \termetrois{} is clearly lower bounded by $0$
  but it is not obvious for \termedeux{} as there is no reason for $T_k(T)$ to be upper bounded by $T$.
  Let $T_k^{!}(T) := \sum_{t=1}^{T} \indic(\exists! j, A^j(t)=k)$,
  where the notation $\exists!$ stands for ``there exists a unique''.
  Then $T_k(T) = \sum_{t=1}^{T} \sum_{j=1}^{M} \indic(A^j(t) = k)$ can be decomposed as
  \begin{equation*}
    T_k(T) = \sum_{t=1}^{T} \indic(\exists! j, A^j(t) = k) + \sum_{t=1}^{T} \sum_{j=1}^{M} c_{k,t} \indic(A^j(t) = k)
    = T_k^{!}(T) + C_k(T).
  \end{equation*}
  By focusing on the two terms $\termedeux + \termetrois$ from the decomposition of $R_T(\boldsymbol{\mu}, M, \rho)$ from Lemma~\ref{lem:DecompositionRegret}, we have
  \begin{align*}
    \termedeux + \termetrois &=
    \sum_{k \in \Mbest} (\mu_k - \mu_M^*) (T - \E_{\mu}[T_k^!(T)])
    + \sum_{k \in \Mbest} \mu_M^* \E_{\mu}[C_k(T)] \\
    & \;\;\;\;\;\; + \sum_{k=1}^{M} \mu_k \E_{\mu}[C_k(T)]
    - \sum_{k \in \Mbest} \mu_k \E_{\mu}[C_k(T)] \\
    &=
    \sum_{k \in \Mbest} (\mu_k - \mu_M^*) (T - \E_{\mu}[T_k^!(T)])
    + \sum_{k \in \Mbest} \mu_M^* \E_{\mu}[C_k(T)]
    + \sum_{k \in \Mworst} \mu_k \E_{\mu}[C_k(T)] \\
    &=
    \sum_{k \in \Mbest} (\mu_k - \mu_M^*) (T - \E_{\mu}[T_k^!(T)])
    + \sum_{k=1}^{M} \min(\mu_M^*, \mu_k) \E_{\mu}[C_k(T)].
  \end{align*}
  And now both terms are non-negative, as $T_k^!(T) \leq T$, $\min(\mu_M^*, \mu_k)\geq 0$, and $C_k(T) \geq 0$, so $\termedeux + \termetrois \geq 0$
  which proves that $R_T(\boldsymbol{\mu}, M, \rho) \geq \termeun$, as wanted.

% -----------------------------------------------------------------
\subsection{Proof of Lemma~\ref{lem:1stUpperBound}}
\label{proof:1stUpperBound}

Recall that we want to upper bound
$ (b) : = \sum_{k \in \Mbest} (\mu_k - \mu_{M*}) \left(T - \E_{\mu}[T_k(T)]\right)$.
First, we observe that, for all $k\in \Mbest$,
\begin{eqnarray*}
T - \E_{\mu}[T_k(T)] & \leq & T - \E_{\mu}\left[\sum_{t=1}^T \indic(\exists j : A^j(t) = k)\right] = \E_{\mu}\left[\sum_{t=1}^T \indic(\forall j, A_j(t) \neq k)\right] = \E_{\mu}\left[\sum_{t=1}^T \indic(k \notin \widehat{S}_t)\right],\end{eqnarray*}
where we denote by $\widehat{S}_t = \{A^j(t), j \in \{1,\dots,M\}\}$ the set of selected arms at time $t$ (with no repetition). With this notation one can write
\begin{eqnarray*}
 (b) & \leq & (\mu_1 - \mu_{M^*})  \sum_{k \in \Mbest} \left(T - \E_{\mu}[T_k(T)]\right) \leq  (\mu_1 - \mu_{M^*})  \E_{\mu}\left[\sum_{k \in \Mbest} \sum_{t = 1}^T \indic(k \notin \widehat{S}_t)\right] \\
 & = &  (\mu_1 - \mu_{M^*})  \E_{\mu}\left[ \sum_{t = 1}^T \sum_{k \in \Mbest}\indic(k \notin \widehat{S}_t)\right].
\end{eqnarray*}
The quantity $\sum_{k \in \Mbest}\indic(k \notin \widehat{S}_t)$ counts the number of optimal arms that have not been selected at time $t$. For each mis-selection of an optimal arm, there either exists a sub-optimal arm that has been selected, or an arm in $\Mbest$ on which a collision occurs. Hence
\[\sum_{k \in \Mbest}\indic(k \notin \widehat{S}_t) = \sum_{k \in \Mbest}\indic(C_k(t)) + \sum_{k \in \Mworst} \indic(\exists j : A^j(t) = k),\]
which yields
\[\E_{\mu}\left[ \sum_{t = 1}^T \sum_{k \in \Mbest}\indic(k \notin \widehat{S}_t)\right] \leq \sum_{k \in \Mbest}\E_{\mu}\left[\cC_k(T)\right] + \sum_{k \in \Mworst} \E_{\mu}\left[T_k(T)\right]\]
and Lemma~\ref{lem:1stUpperBound} follows.

% -----------------------------------------------------------------
% -----------------------------------------------------------------
\section{Lower Bound: Proof of Theorem~\ref{thm:BetterLowerBound}}
\label{proof:BetterLowerBound}

% -----------------------------------------------------------------
\subsection{Proof of Theorem~\ref{thm:BetterLowerBound}}

The lower bound that we present relies on the following \emph{change-of-distribution} lemma that we prove in the next section, following recent arguments from \cite{Garivier16TrueShape} that have to be adapted to incorporate the collision information.

\begin{lemma}\label{lem:CD} Under observation model \modelun{} and \modeldeux, for every event $A$ that is $\cF_{T}^{j}$-measurable, considering two multi-player bandit models denoted by $\mu$ and $\lambda$ respectively, it holds that
\begin{equation}
  \sum_{k=1}^{K}\E_{\mu}\left[T_k^j(T)\right]\kl(\mu_k,\lambda_k) \geq \kl \left(\Pr_{\mu}(A),\Pr_{\lambda}(A)\right).
\end{equation}
\end{lemma}

Let $k$ be a sub-optimal arm under $\mu$,
fix $\varepsilon \in \left(0, \mu_{M-1}^* - \mu_M^*\right)$,
% We define the bandit instance $\lambda$ such that
and let $\lambda$ be the bandit instance such that
\[\left\{\begin{array}{ccl}
          \lambda_\ell & = & \mu_\ell \ \ \  \text{for all } \ell \neq k, \\
          \lambda_k & = & \mu_{M^*} + \varepsilon.
         \end{array}
\right.\]
Clearly, $\lambda\in\cP_M$ also,
and the set of $M$ best arms under $\mu$ and $\lambda$ differ by one arm: if \Mbest$_{\mu} = \{1^*,\dots,M^*\}$ then $\Mbest_{\lambda} = \{1^*,\dots,(M-1)^*,k\}$.
Thus, one expects the ($\cF^j_T$-mesurable) event
\[A_T = \left(T_k^j(T) > \frac{T}{2M}\right)\]
to have a small probability under $\mu$ (under which $k$ is sub-optimal) and a large probability under $\lambda$ (under which $k$ is one of the optimal arms, and is likely to be drawn a lot).

Applying the inequality in Lemma~\ref{lem:CD}, and noting that the sum in the left-hand side reduces to on term as there is a single arm whose distribution is changed, one obtains
\begin{eqnarray*}
  \E_{\mu}\left[T_k^j(T)\right] \kl (\mu_k, \mu_{M^*}+\varepsilon) &\geq & \kl \left(\Pr_{\mu}(A_T),\Pr_{\lambda}(A_T)\right),\nonumber \\
  \E_{\mu}\left[T_k^j(T)\right] \kl (\mu_k, \mu_{M^*}+\varepsilon) &\geq & \left(1 - \Pr_{\mu}(A_T)\right) \log \left(\frac{1}{\Pr_{\lambda}(\overline{A_T})}\right) - \log(2),
\end{eqnarray*}
using the fact that the binary KL-divergence satisfies $\kl(x,y) = \kl(1-x,1-y)$ as well as the inequality  $\kl(x, y) \geq x \log\left(1/y\right) - \log(2)$, proved by \cite{Garivier16TrueShape}.
Now, using Markov inequality yields
\begin{align*}
  \Pr_{\mu}\left(A_T\right)
    & \leq 2M\frac{\E_\mu\left[T^j_k(T)\right]}{T} =: x_T, \\
  \Pr_{\lambda}\left(\overline{A_T}\right)
    & = \Pr_{\lambda}\left(\frac{T}{M} - T^j_k(T) > \frac{T}{2M}\right) \leq 2M\frac{\E_\lambda\left[ \frac{T}{M} - T^j_k(T)\right]}{T} =: \frac{y_T}{T},
\end{align*}
which defines two sequences $x_T$ and $y_T$, such that
\begin{eqnarray}\label{eq:CqCD}
  \E_{\mu}\left[T_k^j(T)\right] \kl (\mu_k, \mu_{M^*}+\varepsilon) &\geq & \left(1 - x_T\right) \log \left(\frac{T}{y_T}\right) - \log(2).
\end{eqnarray}
The strong uniform efficiency assumption  (see Definition~\ref{def:DecentralizedUniformEfficiency}) further tells us that $x_T \rightarrow 0$ (as $\E_\mu[T_k^j(T)] = o(T^\alpha)$ for all $\alpha$) and $y_T = o(T^\alpha)$ when $T \to \infty$, for all $\alpha \in (0,1)$.
As a consequence, observe that $\log(y_T)/\log(T) \rightarrow 0$ when $T$ tends to infinity and
\[
  \frac{\left(1 - x_T\right) \log \left({T}/{y_T}\right) - \log(2)}{\log(T)} = 1-x_T - (1-x_T) \frac{\log(y_T)}{\log(T)} - \frac{\log(2)}{\log(T)}
\]
tends to one when $T$ tends to infinity.
From Equation~\eqref{eq:CqCD}, this yields
\begin{equation}\label{eq:CqCD2}
  \liminf_{T\to\infty} \frac{\E_{\mu}[T_k^j(T)]}{\log(T)} \geq \frac{1}{\kl(\mu_k,\mu_{M^*}+\varepsilon)},
\end{equation}
for all $\varepsilon \in \left(0, \mu_{M-1}^* - \mu_M^*\right)$.
Letting $\varepsilon$ go to zero gives the conclusion (as $\kl$ is continuous).

% -----------------------------------------------------------------

\subsection{Proof of Lemma~\ref{lem:CD}}

Under observation model \modelun{} and \modeldeux, the strategy $A^j(t)$ decides which arm to play based on the information contained in $I_{t-1}$, where $I_0 = U_0$ and
\[\forall t > 0, \ \ I_t = (U_0,Y_1,C_1,U_1,\dots,Y_t,C_t,U_t)\]
where $Y_{t} := Y_{A^j(t-1),t}$ denotes the sensing information, $C_t := C^j(t)$ denotes the collision information (not always completely exploited under observation model \modeldeux) and $U_t$ denotes some external source of randomness\footnote{For instance, \MCTopM, \RandTopM{} and \rhoRand{} draws from a uniform variable in $\{1,\dots,M\}$ for new ranks or arms.} useful to select $A^j(t)$.
Formally, one can say that $A^j(t)$ is $\sigma(I_{t-1})$ measurable\footnote{$\sigma(I_t)$ denotes the sigma-Algebra generated by the observations $I_t$.} (as $\cF^j(t) \subseteq \sigma(I_{t})$, with an equality under observation model \modelun).

Under two bandit models $\mu$ and $\lambda$, we let $\Pr_{\mu}^{I_t}$ (resp. $\Pr_{\lambda}^{I_t}$) be the distribution of the observations under model $\mu$ (resp. $\lambda$), given a fixed algorithm.
Using the exact same technique as \cite{Garivier16TrueShape} (the contraction of entropy principle), one can establish that for any event $A$ that is $\sigma(I_t)$-measurable\footnote{In the work of \cite{Garivier16TrueShape}, the statement is more general and the probability of an event $A$ is replaced by the expectation of any $\cF_T$-measurable random variable $Z$ bounded in $[0,1]$.},
\[\KL\left(\Pr_{\mu}^{I_t},\Pr_{\lambda}^{I_t}\right) \geq \kl\left(\Pr_{\mu}(A),\Pr_{\lambda}(A)\right).\]

The next step is to relate the complicated KL-divergence $\KL\left(\Pr_{\mu}^{I_t},\Pr_{\lambda}^{I_t}\right)$ to the number of arm selections.
Proceeding similarly as \cite{Garivier16TrueShape}, one can write, using the chain rule for KL-divergence, that
\begin{equation}\label{eq:Iterative}\KL\left(\Pr_{\mu}^{I_{t}},\Pr_{\lambda}^{I_{t}}\right) =  \KL\left(\Pr_{\mu}^{I_{t-1}},\Pr_{\lambda}^{I_{t-1}}\right) + \KL\left(\Pr_{\mu}^{Y_{t},C_t,U_t | I_{t-1}},\Pr_{\lambda}^{Y_{t},C_t,U_t | I_{t-1}}\right).\end{equation}
Now observe that conditionally to $I_{t-1}$, $U_t$, $Y_t$ and $C_t$ are independent, as once the selected arm is known, the value of the sensing $Y_t$ does not influence the other players selecting that arm, and $U_t$ is some exogenous randomness.
Using further that the distribution of $U_t$ is the same under $\mu$ and $\lambda$, one obtains
\begin{equation}\KL\left(\Pr_{\mu}^{Y_{t},C_t | I_{t-1}},\Pr_{\lambda}^{Y_{t},C_t | I_{t-1}}\right) =
 \KL\left(\Pr_{\mu}^{Y_{t}| I_{t-1}},\Pr_{\lambda}^{Y_{t}| I_{t-1}}\right) + \KL\left(\Pr_{\mu}^{C_t | I_{t-1}},\Pr_{\lambda}^{C_t | I_{t-1}}\right).\label{eq:DecKL}
\end{equation}

The first term in \eqref{eq:DecKL} can be rewritten using the same argument as \cite{Garivier16TrueShape}, that relies on the fact that conditionally to $I_{t-1}$, $Y_t$ is a Bernoulli distribution with mean $\mu_{A^j(t)}$ under the instance $\mu$ and $\lambda_{A^j(t)}$ under the instance $\lambda$:
\begin{eqnarray*}
\KL\left(\Pr_{\mu}^{Y_{t}| I_{t-1}},\Pr_{\lambda}^{Y_{t}| I_{t-1}}\right)  & = & \E_\mu\left[\E_\mu\left[ \KL\left(\Pr_{\mu}^{Y_{t}| I_{t-1}},\Pr_{\lambda}^{Y_{t}| I_{t-1}}\right) | I_{t-1} \right]\right] \\
& = & \E_\mu \left[\kl \left(\mu_{A^j(t)},\lambda_{A^j(t)}\right)\right] \\
& = & \E_\mu  \left[\sum_{k=1}^K \indic{(A^j(t)=k)} \kl \left(\mu_{k},\lambda_{k}\right)\right],
\end{eqnarray*}
We now show that second term in \eqref{eq:DecKL} is zero:
\begin{eqnarray*}
\KL\left(\Pr_{\mu}^{C_t | I_{t-1}},\Pr_{\lambda}^{C_t | I_{t-1}}\right)
& = & \E_\mu\left[\E_\mu\left[ \KL\left(\Pr_{\mu}^{C_t| I_{t-1}},\Pr_{\lambda}^{C_t| I_{t-1}}\right) | I_{t-1} \right]\right] \\
& = & \E_\mu\left[\E_\mu\left[\E_\mu\left[ \KL\left(\Pr_{\mu}^{C_t| I_{t-1}},\Pr_{\lambda}^{C_t| I_{t-1}}\right) | \bigcup_{j' \neq j} I_{t-1}^{j'}\right]| I_{t-1} \right]\right], \\
\end{eqnarray*}
where $I_{t-1}^{j'}$ denote the information available to player $j' \neq j$.
Knowing the information available to all other players player $C_t$ is an almost surely constant random variable, whose distribution is the same under $\mu$ and $\lambda$.
Hence the inner expectation is zero and so does $\KL\left(\Pr_{\mu}^{C_t | I_{t-1}},\Pr_{\lambda}^{C_t | I_{t-1}}\right)$.

Putting things together, we showed that
\[\KL\left(\Pr_{\mu}^{I_t},\Pr_{\lambda}^{I_t}\right) = \KL\left(\Pr_{\mu}^{I_{t-1}},\Pr_{\lambda}^{I_{t-1}}\right) + \E_\mu  \left[\sum_{k=1}^K \indic{(A^j(t)=k)} \kl \left(\mu_{k},\lambda_{k}\right)\right].\]
Iterating this equality and using that $\KL\left(\Pr_{\mu}^{I_{0}},\Pr_{\lambda}^{I_{0}}\right)=0$ yields that
\[\sum_{k=1}^{K}\E_{\mu}\left[T_k^j(T)\right]\kl(\mu_k,\lambda_k) \geq \kl \left(\Pr_{\mu}(A),\Pr_{\lambda}(A)\right),\]
for all $A \in \sigma(I_T)$, in particular for all $A \in \cF^j_T$.

% -----------------------------------------------------------------
% -----------------------------------------------------------------
\section{The \RandTopM{} algorithm}
\label{app:RandTopM}

We now state precisely the \RandTopM{} algorithm below in Algorithm~\ref{algo:RandTopM} (page \pageref{algo:RandTopM}).
It is essentially the same algorithm as \MCTopM,
but in a simpler version as the ``Chair'' aspect is removed, that is, there is no notion of state $s^j(t)$ (\cf{} Algorithm~\ref{algo:MCTopM}).
Player $j$ is always considered ``not fixed'',
and a \emph{collision always forces a uniform sampling of the next arm} from $\TopM(t)$
in the case of \RandTopM.

\vspace*{-5pt}  % XXX remove if problem
\begin{small}  % XXX remove if problem
  \begin{figure}[!ht]
      % \begin{framed}  % XXX remove if problem
      \begin{small}  % XXX remove if problem
      \centering
      % Documentation at http://mirror.ctan.org/tex-archive/macros/latex/contrib/algorithm2e/doc/algorithm2e.pdf if needed
      % Or https://en.wikibooks.org/wiki/LaTeX/Algorithms#Typesetting_using_the_algorithm2e_package
      % \removelatexerror% Nullify \@latex@error % Cf. http://tex.stackexchange.com/a/82272/
      \begin{algorithm}[H]
          % XXX Options
          % \LinesNumbered  % XXX Option to number the line
          % \RestyleAlgo{boxed}
          % XXX Input, data and output
          % \KwIn{$K$ and policy $P^j$ for arms set $\{1,\dots,K\}$\;}
          % \KwData{Data}
          % \KwResult{Result}
          % XXX Algorithm
              Let $A^j(1) \sim \cU(\{1,\dots,K\})$ and $C^j(1)=\mathrm{False}$ \\
              \For{$t = 0, \dots, T - 1$}{
                  \eIf{$A^j(t) \notin \TopM(t)$}{
                    \eIf(\tcp*[f]{collision}){$C^j(t)$}{
                      $A^j(t+1) \sim \cU \left(\TopM(t)\right)$
                      \tcp*[f]{randomly switch}
                      }(\tcp*[f]{randomly switch on an arm that had smaller UCB at $t-1$}){
                        $A^j(t+1) \sim \cU \left(\TopM(t) \cap \left\{k : g_k^j(t-1) \leq g^j_{A^j(t)}(t-1)\right\}\right)$
                      }
                    }{
                      $A^j(t+1) = A^j(t)$
                      \tcp*[f]{stays on the same arm}
                    }
                  Play arm $A^j(t+1)$, get new observations (sensing and collision), \\
                  Compute the indices $g^j_k(t+1)$ and set $\TopM(t+1)$ for next step.
              }
              \caption{The \RandTopM{} decentralized learning policy (for a fixed underlying index policy $g^j$).}
          \label{algo:RandTopM}
      \end{algorithm}
      \end{small}  % XXX remove if problem
      % \end{framed}  % XXX remove if problem
  \end{figure}
\end{small}  % XXX remove if problem
\vspace*{-5pt}  % XXX remove if problem

% -----------------------------------------------------------------
% -----------------------------------------------------------------

\section{Proofs Elements Related to Regret Upper Bounds}
\label{proof:RegretUpperBounds}

This Appendix includes the main proofs, missing from the content of the article,
that yield the regret upper bound.
We start by controlling the sub-optimal draws when the \klUCB{} indices are used (instead of \UCB),
with any of our proposed algorithms (\MCTopM, \RandTopM) or \rhoRand. Then we focus on controlling collisions for \MCTopM-\klUCB.

\subsection{Control of the Sub-optimal Draws for \klUCB: Proof of Lemma~\ref{lem:SubOptimalSelections}}
\label{proof:SubOptimalSelections}

%
%     Recall $g_k^j(t) \in \mathbb{R}$ denote the index of arm $k$ for user $j$ at time $t$.
%     As only user $j$ is considered here, the superscript $j$ is dropped.
%
Fix $k\in\Mworst$ and a player $j \in \{1,\dots,M\}$.
The key observation is that for \MCTopM, \RandTopM{} as well as the \rhoRand{} algorithm, it holds that
\begin{equation}\left(A^j(t) = k\right) = \left(A^j(t) = k , \exists m \in \Mbest : g_m^j(t) < g_k^j(t) \right).\label{eq:KeyInclusion}\end{equation}
Indeed, for the three algorithms, an arm selected at time $t+1$ belongs to the set $\TopM(t)$ of arms with $M$ largest indices.
If the sub-optimal arm $k$ is selected at time $t$, it implies that $k \in \TopM(t)$, and, because there are $M$ arms in both \Mbest{} and $\TopM(t)$, one of the arms in \Mbest{} must be excluded from $\TopM(t)$.
In particular, the index of arm $k$ must be larger than the index of this particular arm $m$.

Using \eqref{eq:KeyInclusion}, one can then upper bound the number of selections of arm $k$ by user $j$ up to round $T$ as
\begin{align*}
\E_{\mu}[T_k^j(T)]
&= \E_{\mu}\left[ \sum_{t=1}^T \indic\left( A^j(t) = k \right) \right]
= \sum_{t=1}^T \Pr\left( A^j(t) = k \right).\\
%&= \sum_{t=1}^T \Pr\left( A^j(t) = k,\;\; \exists m \in\Mbest,\; g_m(t) < g_k(t) \right) \\
&= \sum_{t=1}^T \Pr\left( A^j(t) = k,\;\; \exists m \in\{1,\dots,M\}:\; g_{m^*}^j(t) < g_k^j(t) \right).
\shortintertext{Considering the relative position of the upper-confidence bound $g_{m^*}^j(t)$ and the corresponding mean $\mu_m^* = \mu_{m^*}$, one can write the decomposition}
\E_{\mu}[T_k^j(T)] &\leq \sum_{t=1}^T\Pr\left(A^j(t) = k,\;\; \exists m \in\{1,\dots,M\}:\; g_{m^*}(t) \leq g_k(t) , \forall m \in \{1,\dots,M\}: \;  g_{m^*}(t) \geq \mu_m^* \right)\\
&\;\;\;\;\;\;\;\; + \sum_{t=1}^T\Pr\left( \exists m \in\{1,\dots,M\}: \; g_{m^*}(t) < \mu_m^* \right) \\
&\leq \sum_{t=1}^T\Pr\left(A^j(t) = k,\;\; \exists m \in\{1,\dots,M\}:\; \mu_{m}^* \leq g_k(t)\right)+ \sum_{m=1}^M\sum_{t=1}^T\Pr\left(g_{m^*}(t) < \mu_m^* \right) \\
&\leq
\sum_{t=1}^T
  \Pr\left( A^j(t) = k,\; \mu_{M^*} \leq g_k(t) \right)
+
\sum_{m=1}^M
  \sum_{t=1}^T \Pr\left( g_{m^*}(t) < \mu_m^* \right),
\end{align*}
where the last inequality (for the first term) comes from the fact that $\mu_{M^*}$ is the smallest of the $\mu_{m^*}$ for $m \in \{1, \dots,M\}$.

Now each of the two terms in the right hand side can directly be upper bounded using tools developed by \cite{KLUCBJournal} for the analysis of kl-UCB.
The rightmost term can be controlled using Lemma~\ref{lem:Fact1KLUCB} below that relies on a self-normalized deviation inequality, whose proof exactly follows from the proof of Fact 1 in Appendix A of \cite{KLUCBJournal}.
The leftmost term can be controlled using Lemma~\ref{lem:Fact2KLUCB} stated below, that is a direct consequence of the proof of Fact 2 in Appendix A of \cite{KLUCBJournal}.

\begin{lemma}\label{lem:Fact1KLUCB}
    For any arm $k$, if $g_k^{j}(t)$ is the \klUCB{} index with exploration function $f(t)=\log(t)+3\log\log(t)$,
    \begin{equation}
      \sum_{t=1}^T \Pr\left(g_k^{j}(t) < \mu_k\right) \leq 3 + 4e \log\log(T).
    \end{equation}
\end{lemma}

Denote $\kl'(x,y)$ the derivative of the function $x \mapsto \kl(x,y)$ (for any fixed $y\neq 0, 1$).

\begin{lemma}\label{lem:Fact2KLUCB} For any arms $k$ and $k'$ such that $\mu_{k'} > \mu_{k}$, if $g_k^{j}(t)$ is the kl-UCB index with exploration function $f(t)$,
\begin{align*}
  &\sum_{t=1}^T\Pr\left( A^j(t) = k, \mu_{k'} \leq g_k^{j}(t) \right) & \leq  \frac{f(T)}{\kl(\mu_k,\mu_{k'})}
  + \sqrt{2\pi} \sqrt{\frac{\kl'(\mu_k,\mu_{k'})^2}{\kl(\mu_k,\mu_{k'})^3}}\sqrt{f(T)} + 2\left(\frac{\kl'(\mu_k,\mu_{k'})}{\kl(\mu_k,\mu_{k'})}\right)^2 + 1.\end{align*}
\end{lemma}

Putting things together, one obtains the non-asymptotic upper bound
\begin{align}\label{eq:UBprecise}
  \E_{\mu}\left[T_k^j(T)\right]
  & \leq \frac{\log(T) + 3 \log\log(T)}{\kl(\mu_k,\mu_{M^*})} + \sqrt{2\pi} \sqrt{\frac{\kl'(\mu_k,\mu_{M^*})^2}{\kl(\mu_k,\mu_{M*})^3}}\sqrt{\log(T) + 3\log\log(T)}  \nonumber\\
  & \;\;\;\;\;\;+ 2\left(\frac{\kl'(\mu_k,\mu_{M^*})}{\kl(\mu_k,\mu_{M^*})}\right)^2 + 4Me \log\log(T) + 3M+1,
\end{align}
which yields Lemma~\ref{lem:SubOptimalSelections},
with explicit constants $C_{\boldsymbol{\mu}}$ and $D_{\boldsymbol{\mu}}$.

% --------------------------------------------------------------------------------
\subsection{Proof of Lemma~\ref{lem:elementaryLemma_RandTopM_MCTopM}}\label{proof:elementaryLemma_RandTopM_MCTopM}

Using the behavior of the algorithm when the current arm leaves the set $\TopM$ (Line 4), one has
\begin{align*}
  & \sum_{t=1}^T \bP\left(A^j(t) = k, k \notin \hat{M}^j(t)\right)  \\ & \leq \sum_{t=1}^T \bP\left(A^j(t) = k, k \notin \hat{M}^j(t), A^j(t+1) \in \TopM(t) \cap \{k': g_{k'}^j(t-1) \leq g_k^j(t-1)\}\right) \\
  & \leq   \sum_{t=1}^T \sum_{k' \neq k}\bP\left(A^j(t) = k,A^j(t+1) = k', g_{k'}^j(t) \geq g_{k}^j(t), g_{k'}^j(t-1) \leq g_{k}^j(t-1)  \right) \\
  & = \sum_{k' \neq k} \underbrace{\sum_{t=1}^T \bP\left(A^j(t) = k,A^j(t+1) = k', g_{k'}^j(t) \geq g_{k}^j(t), g_{k'}^j(t-1) \leq g_{k}^j(t-1)  \right)}_{:= T_{k'}}
\end{align*}

Now, to control $T_{k'}$, we distinguish two cases. If $\mu_k < \mu_{k'}$, one can write
\[T_{k'} \leq \sum_{t=1}^T \bP\left(g_{k'}^j(t) \leq \mu_{k'}\right) + \sum_{t=1}^T \bP\left(A^j(t) = k, g_{k}^j(t-1) \geq \mu_{k'}\right)\]
The first term in the right hand side is $o(\log(T))$ by Lemma~\ref{lem:Fact1KLUCB}. To control the second term, we apply the same trick that led to the proof of Lemma~\ref{lem:Fact2KLUCB} in \cite{KLUCBJournal}.
Letting $\kl^+(x, y) := \kl(x, y) \indic(x \geq y)$,
and $\widehat{\mu}^j_{k,s}$ be the empirical mean of the $s$ first observations from arm $k$ by player $j$, one has
\begin{align}
  & \sum_{t=1}^T \bP\left(A^j(t) = k, g_{k}^j(t-1) \geq \mu_{k'}\right)
  =  \bE \left[ \sum_{t=1}^T\sum_{s=1}^{t-1} \indic{\left(A^j(t) = k , N^j_k(t-1) = s\right)}\indic{\left( s  \times \kl^+\left(\widehat{\mu}_{k,s}^j , \mu_k\right) \leq f(t)\right)} \right]\nonumber\\
  & \hspace{1cm} \leq \bE \left[ \sum_{s=1}^{T}\indic{\left( s \times \kl^+\left(\widehat{\mu}_{k,s}^j , \mu_k\right) \leq f(T)\right)}\sum_{t=s-1}^T \indic{\left(A^j(t) = k , N^j_k(t-1) = s\right)} \right] \nonumber \\
  & \hspace{1cm}  \leq \sum_{s=1}^T \bP\left( s \times \kl^+\left(\widehat{\mu}_{k,s}^j , \mu_k\right) \leq f(T)\right),\label{eq:FromHere}
\end{align}
where the last inequality uses that for all $s$, \[\sum_{t=s-1}^T \indic{\left(A^j(t) = k , N^j_k(t-1) = s\right)} = \sum_{t=s-1}^T \indic{\left(A^j(t) = k , N^j_k(t) = s + 1\right)} \leq 1.\]
From \eqref{eq:FromHere}, the same upper bound as that of Lemma~\ref{lem:Fact2KLUCB} can be obtained using the tools from \cite{KLUCBJournal}, which proves that for $T\to\infty$,
\[T_{k'} = \frac{\log(T)}{\kl(\mu_k,\mu_{k'})} + o(\log(T)).\]

If $\mu_k > \mu_{k'}$, we rather use that
\[T_{k'} \leq \sum_{t=1}^T \bP\left(g_{k}^j(t) \leq \mu_{k}\right) + \sum_{t=1}^T \bP\left(A^j(t+1) = k', g_{k'}^j(t) \geq \mu_{k}\right)\]
and similarly Lemma~\ref{lem:Fact1KLUCB} and a slight variant of Lemma~\ref{lem:Fact2KLUCB} to deal with the modified time indices yields
\[T_{k'} = \frac{\log(T)}{\kl(\mu_{k'},\mu_{k})} + o(\log(T)).\]
Summing over $k'$ yields the result.

\subsection{Controlling Collisions for \MCTopM: Proof of Lemma~\ref{lem:collisionsMCTopM}}
\label{proof:collisionsMCTopM}

A key feature of both the \RandTopM{} and \MCTopM{} algorithms is Lemma~\ref{lem:elementaryLemma_RandTopM_MCTopM}, that states that the probability of switching from some arm because this arm leaves $\TopM(t)$ is small. Its proof is postponed to the end of this section.

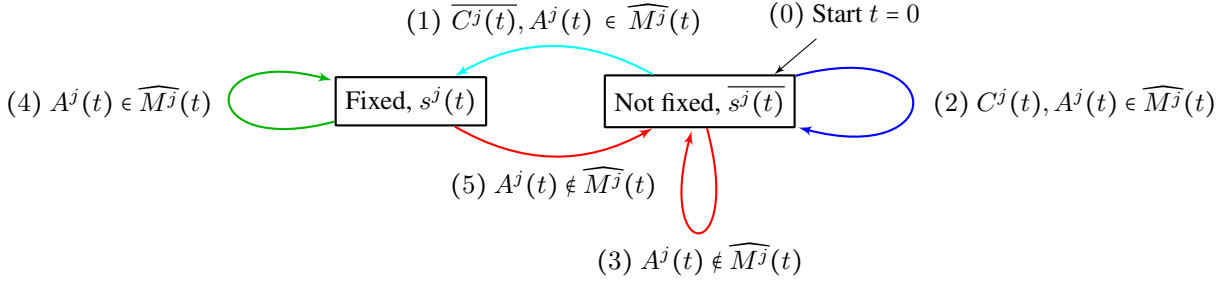
\begin{figure}[!ht]
  \begin{tikzpicture}[>=latex',line join=bevel,scale=3.8]
      \node (start) at (1.5,0.30) {$(0)$ Start $t=0$};
      \node (notfixed) at (1,0) [draw,rectangle,thick] {Not fixed, $\overline{s^j(t)}$};
      \node (fixed) at (0,0) [draw,rectangle,thick] {Fixed, $s^j(t)$};
      \draw [black,->] (start) -> (notfixed.20);
      \draw [color=cyan,thick,->] (notfixed) to[bend right] node[midway,above,text width=5cm,text centered,black] {\small $(1)$ $\overline{C^j(t)}, A^j(t) \in \TopM(t)$} (fixed);
      \path [color=blue,thick,->] (notfixed) edge[loop right] node[right,text width=4cm,text badly centered,black] {\small $(2)$  $C^j(t), A^j(t) \in \TopM(t)$} (1);
      \path [color=red,thick,->] (notfixed) edge[loop below] node[below,text centered,black] {\small $(3)$  $A^j(t) \notin \TopM(t)$} (1);
      \path [color=darkgreen,thick,->] (fixed) edge[loop left] node[left,text width=2.9cm,text badly centered,black] {\small $(4)$ $A^j(t) \in \TopM(t)$} (fixed);
      \draw [color=red,thick,->] (fixed) to[bend right] node[midway,below,text centered,black] {\small $(5)$  $A^j(t) \notin \TopM(t)$} (notfixed);
  \end{tikzpicture}
  \caption{\small Player $j$ using $\mathrm{MCTopM}$, represented as ``state machine'' with $5$ transitions.
  Taking one of the five transitions means playing one round of the Algorithm~\ref{algo:MCTopM}, to decide $A^j(t+1)$ using information of previous steps.}
  \label{fig:StateMachineAlgorithm_MCTopM}
\end{figure}

Figure~\ref{fig:StateMachineAlgorithm_MCTopM} below provides a schematic representation of the execution of the \MCTopM{} algorithm, that has to be exploited in order to properly control the number of collisions.
The sketch of the proof is the following: by focusing only on collisions in the ``not fixed'' state, bounding the number of transitions $(2)$ and $(3)$ is enough.
Then, we show that both the number of transitions $(3)$ and $(5)$ are small: as a consequence of Lemma~\ref{lem:elementaryLemma_RandTopM_MCTopM}, the average number of these transitions is $\bigO{\log T}$.
Finally, we use that the length of a sequence of consecutive transitions $(2)$ is also small (on average smaller than $M$), and except for possibly the first one, starting a new sequence implies a previous transition $(3)$ or $(5)$ to arrive in the state ``not fixed''. This gives a logarithmic number of transitions $(2)$ and $(3)$, and so gives $\E_{\mu}[\sum_k\cC_k(T)] = \bigO{\log T}$,
with explicit constants depending on $\boldsymbol{\mu}$ and $M$.

% Explanation of the figure

As in Algorithm~\ref{algo:MCTopM}, $s^j(t)$ is the event that player $j$ decided to fix herself on an arm at the end of round $t-1$.
Formally, $s^j(0)$ is false, and $s^j(t+1)$ is defined inductively from $s^j(t)$ as
% following the algorithm of \MCTopM:
\begin{equation}
    s^j(t+1) =
    \left( s^j(t) \cup \left( \overline{s^j(t)} \cap \overline{C^j(t)} \right) \right)
    \cap \left( A^j(t) \in \TopM(t) \right).
\end{equation}

For the sake of clarity, we now explain Figure~\ref{fig:StateMachineAlgorithm_MCTopM} in words. At step $t$, if player $j$ is not fixed ($\overline{s^j(t)}$), she can have three behaviors when executing \MCTopM.
She keeps the same arm and goes to the other state $s^j(t)$ with transition $(1)$,
or she stays in state $\overline{s^j(t)}$,
with two cases.
Either she sampled $A^j(t+1)$ uniformly
from $\TopM(t) \cap \{ m : g_m^j(t) \leq g_k^j(t) \}$
with transition $(3)$,
in case of collision and if $A^j(t+1) \in \TopM(t)$,
or she sampled $A^j(t+1)$ uniformly
from $\TopM(t)$ with transition $(2)$,
if $A^j(t+1) \notin \TopM(t)$.
In particular, note that if $\overline{C^j(t)}$, transition $(3)$ is executed and not $(2)$.
%
% The case $(2)$ happens when there is no collision at time $t-1$ but a collision at time $t$,
% and conversely case $(3)$ happens when the player was not fixed on her arm at time $t-1$ (\ie, $\overline{s^j(t-1)}$)
% and experienced a collision at time $t-1$
Transition $(3)$ is a uniform sampling from $\TopM(t)$ (the ``Musical Chair'' step).

For player $j$ and round $t$, we now introduce a few events that are useful in the proof. First, for every $x=1,2,3,4,5$, we denote $I_x^j(t)$ the event that a transition of type $(x)$ occurs for player $j$ after the first $t$ observations (\ie, between round $t$ and round $t+1$, to decide $A^j(t+1)$).
Formally they are defined by
\begin{align*}
  I_1^j(t) &:= \left(\overline{s^j(t)},\overline{C_j(t)}, A^j(t) \in \TopM(t)\right), \\
  % go sitted
  I_2^j(t) &:= \left(\overline{s^j(t)},C_j(t), A^j(t) \in \TopM(t)\right),
  \ \ \ &\text{and} \ \ \
  I_3(t) &:= \left(\overline{s^j(t)},A^j(t) \notin \TopM(t)\right), \\
  % if sitted
  I_3(t) &:= \left(s^j(t),A^j(t) \in \TopM(t)\right),
  \ \ \ &\text{and} \ \ \
  I_5(t) &:= \left(s^j(t),A^j(t) \notin \TopM(t)\right).
\end{align*}
Then, we introduce $\widetilde{C^j}(t)$ as the event that a collision occurs for player $j$ at round $t$ if she is not yet fixed on her arm, that is
\begin{equation}
    \widetilde{C^j}(t) := \left(C^j(t), \overline{s^j(t)}\right).
\end{equation}

A key observation is that $C^j(t)$ implies $\bigcup_{j'=1}^M \widetilde{C^{j'}}(t)$, as a collision necessarily involves at least one player not yet fixed on her arm ($\overline{s^{j'(t)}}$).
Otherwise, if they are all fixed, \ie, for all $j$, $s^j(t)$, then by definition of $s^j(t)$, none of the player changed their arm from $t-1$ to $t$, and none experienced any collision at time $t-1$ so by induction there is no collision at time $t$.
Thus, $\sum_{j=1}^M \Pr(C^j(t))$ can be upper bounded by $M \sum_{j=1}^M \Pr(\widetilde{C^j}(t))$ (union bound),
and it follows that if $\cC(T) := \sum_{k=1}^K \cC_k(T)$ then
\[\E_{\mu}[\cC(T)] \leq M \sum_{j=1}^M \sum_{t=1}^T \Pr(\widetilde{C^j}(t)).\]
We can further  observe that $\widetilde{C^j}(t)$ implies a transition $(2)$ or $(3)$, as a transition $(1)$ cannot happen in case of collision. Thus another union bound gives
\begin{align}
  \sum_{t=1}^T \Pr(\widetilde{C^j}(t))
  &\leq \sum_{t=1}^T \Pr(I_2^j(t))   + \sum_{t=1}^T \Pr(I_3^j(t)).\label{eq:UBTilde}
\end{align}
In the rest of the proof we focus on bounding the number of transitions $(2)$ and $(3)$.

Let $N_x^j(T)$ be the random variable denoting the number of transitions of type $(x)$.
Neglecting the event $\overline{s^j(t)}$ for $x=3$ and $s^j(t)$ for $x=5$, one has
\begin{equation}
    \E_{\mu}[N_x^j(t)]
    = \sum_{t=1}^T \Pr(I_x^j(t))
    \leq \sum_{t=1}^T \Pr \left(A^j(t) \notin \TopM(t)\right)
    \leq \sum_{t=1}^T \sum_{k=1}^K \Pr \left(A^j(t)=k, k \notin \TopM(t)\right),
\end{equation}
which is $\bigO{\log T}$ (with known constants) by Lemma~\ref{lem:elementaryLemma_RandTopM_MCTopM}. In particular, this controls the second term in the right hand side of \eqref{eq:UBTilde}.

To control the first term $\sum_{t=1}^T \Pr(I_2^j(t))$
we introduce three sequences of random variables,
the starting times $(\theta_i)_{i \geq 1}$
and the ending times $(\tau_i)_{i \geq 1}$
(possibly larger than $T$),
of sequences during which $I_2(s)$ is true for all $s=\theta_i,\dots,\tau_i-1$ but not before and after,
that is
$\forall i \in \{1,\dots,n(T)\},
\overline{I_2^j(\theta_i - 1)}
\cap \bigcap_{t=\theta_i}^{\tau_i-1} I_2^j(t)
\cap \overline{I_2^j(\tau_i)}
$
with $n(T)$ the number of such sequences,
\ie,
$n(T) := \inf \{i \geq 1 : \min(\theta_i, \tau_i) \geq T \}$
(or $0$ if $\theta_1$ does not exist).
If $\theta_i = 1$, the first sequence does not have term $\overline{I_2^j(\theta_i - 1)}$.

Now we can decompose the sum on $t=1,\dots,T$ with the use of consecutive sequences,
\begin{align}
  \E_{\mu}[N_2^j(t)]
  &= \E_{\mu}\left[ \sum_{t=1}^T \indic{\left(I_2^j(t)\right)} \right]
  =
  \E_{\mu}\left[ \sum_{i=1}^{n(T)} \left( \sum_{t=\theta_i}^{\tau_i - 1} 1 + \sum_{t=\tau_i}^{\theta_{i+1} - 1} 0 \right) \right]
  =
  \E_{\mu}\left[ \sum_{i=1}^{n(T)} (\tau_i - \theta_i) \right].
  \intertext{Both $n(T)$ and $\tau_i - \theta_i \geq 0$ have finite averages for any $i$ (as $\tau_i - \theta_i \leq T$), and $n(T)$ is a \emph{stopping time} with respect to the past events (that is, $\cF^j_T$), and so we can obtain}
  % so we can use Wald's Lemma \citep{Wald45}
  &\leq \E_{\mu}\left[ n(T) \right] \times \max_{i\in\N} \E_{\mu}\left[ \tau_i - \theta_i \right]
  = (\alpha) \times (\beta).
\end{align}

$(\alpha)$ To control $\E_{\mu}[n(T)]$, we can observe that
the number of sequences $n(T)$ is smaller than $1$ plus the number of times when \emph{a sequence begins} ($1$ plus because maybe the game starts in a sequence).
And beginning a sequence
% \footnote{Similarly, one could focus on the number of times \emph{a sequence finishes}, and $I_2^j(\theta_i) \cap \overline{I_2^j(\theta_i+1)}$ implies a transition of type $(3)$ or $(1)$.
% The key is that after a transition $(1)$, a new sequence of transitions $(2)$ is possible only after a transition $(5)$, which total number is also controlled as $\bigO{\log T}$.}
at time $\theta_i$ implies
$\overline{I_2^j(\theta_i-1)} \cap I_2^j(\theta_i)$,
which implies a transition of type $(3)$ or $(5)$ at time $\theta_i - 1$, as player j is in state ``not fixed'' at time $\theta_i$ (transitions $(1)$ and $(4)$ are impossible).
As stated above, $\E_{\mu}[N_x^j(T)] = \bigO{\log T}$ for both $x=3$ and $x=5$,
and so $\E_{\mu}[n(T)] = \bigO{\log T}$ also.

$(\beta)$ To control $\E_{\mu}\left[ \tau_i - \theta_i \right]$,
a simple argument can be used.
$\bigcup_{t=\theta_i}^{\tau_i-1} I_2^j(t)$
implies $C^j(t)$ for $\tau_i - \theta_i$ consecutive times.
The very structure of \RandTopM{} gives that in this sequence of transitions $(2)$,
the successive collisions (\ie, $C^j(t-1) \cap C^j(t)$)
implies that each new arm $A^j(t+1)$ for $t \in \{\theta_i, \tau_i-1\}$ is selected uniformly from
$\TopM(t+1)$,
a set of size $M$ with at least one available arm.
Indeed, as there is $M-1$ other players, at time $t+1$ \emph{at least} one arm in $\TopM(t+1)$ is not selected by any player $k'\neq k$,
and so player $j$ has \emph{at least} a probability $1/M$ to select
a free arm, which implies $\overline{C^j(t+1)}$, and so implies the end of the sequence.
In other words, the average length of sequences of transitions $(2)$,
$\E_{\mu}\left[ \tau_i - \theta_i \right]$,
is bounded by the expected number of failed trial of a repeated Bernoulli experiment, with probability of success larger than $1/M$ (by the uniform choice of $A^j(t+1)$ in a set of size $M$ with at least one available arm).
We recognize the mean of a geometric random variable, of parameter $\lambda \geq 1/M$, and so $\E_{\mu}\left[ \tau_i - \theta_i \right] = \frac{1}{\lambda} \leq \frac{1}{1/M} = M$.

This finishes the proof as $\E_{\mu}[N_2^j(T)] = \sum_{t=1}^T \Pr(I_2^j(t)) = \bigO{\log T}$ and so
$\sum_{t=1}^T \Pr(\widetilde{C^j}(t) \cap (A^j(t) = k)) = \bigO{\log T}$
and finally
$\E_{\mu}[\cC(T)] = \sum_{k=1}^K \E_{\mu}[\cC^k(T)] = \bigO{\log T}$ also.

We can be more precise about the constants, all the previous arguments can be used successively:
\begin{align}
  \E_{\mu}[\cC(T)]
  &\leq M \sum_{j=1}^M \left(\sum_{t=1}^T \Pr(I_2^j(t)) + \sum_{t=1}^T \Pr(I_3^j(t))\right)
  = M \left(\sum_{j=1}^M \E_{\mu}[N_2^j(T)] + \E_{\mu}[N_3^j(T)]\right) \\
  &\leq M^2 \left(\E_{\mu}[n(T)] \E_{\mu}[\theta_i - \tau_i] \right) + M^2 \E_{\mu}[N_3^1(T)] \notag  \\
  &\leq M^2 (1+\E_{\mu}[N_3^1(T)] + \E_{\mu}[N_5^1(T)]) M + M^2 \E_{\mu}[N_3^1(T)] \notag \\
  &\leq 2 M^3 \E_{\mu}[N_3^1(T)] + \smallO{\log T} + M^2 \left(\sum_{a,b=1,\dots,K,\;\mu_a < \mu_b} \frac{1}{\kl(\mu_a,\mu_b)}\right) \log(T) + \smallO{\log T} \notag  \\
  &\leq \left(2 M^3 + M^2\right) \left(\sum_{a,b=1,\dots,K,\;\mu_a < \mu_b} \frac{1}{\kl(\mu_a,\mu_b)}\right) \log(T) + \smallO{\log T}.
\end{align}

And so we obtain the desired inequality, with explicit constants, that depend only on $\boldsymbol{\mu}$ and $M$.
\begin{equation}
  \sum_{k=1}^K \E_{\mu}[\cC^k(T)] = \E_{\mu}[\cC(T)]
  \leq M^2\left(2 M + 1\right) \left(\sum_{a,b=1,\dots,K,\;\mu_a < \mu_b} \frac{1}{\kl(\mu_a,\mu_b)}\right) \log(T) + \smallO{\log T}.
\end{equation}

\paragraph{Number of switches}\label{app:NumberSwitches}
Note that we controlled the total number of transitions $(2)$, $(3)$ and $(5)$,
which are the only transitions when a player can switch from arm $k$ to arm $k'\neq k$.
Thus, the total number of arm switches is also proved to be logarithmic, if all players uses
the \MCTopM-\klUCB{} algorithm.

\paragraph{Strong uniform efficiency}\label{app:JustifyingDefinition5}
As soon as $R_T = \bigO{\log T}$ for all problem, \MCTopM{} is clearly proved
to be uniformly efficient, as $\log T$ is $\smallO{T^{\alpha}}$ for any $\alpha\in(0,1)$.
And as justified after Definition~\ref{def:DecentralizedUniformEfficiency} (page~\pageref{def:DecentralizedUniformEfficiency}), uniform efficiency and invariance under permutations of the users implies strong uniform efficiency, and so \MCTopM{} satisfies Definition~\ref{def:DecentralizedUniformEfficiency}.
This is a sanity check: the lower-bound of
Theorem~\ref{thm:BetterLowerBound} indeed applies to our algorithm \MCTopM,
and finally this highlights that it is order-optimal for the regret, in the sense that it matches the lower-bound up-to a multiplicative constant,
and optimal for the term \termeun.

% -----------------------------------------------------------------
% -----------------------------------------------------------------
\section{Additional Discussions on \Selfish}
\label{app:SelfishFails}

% OBSERVATION
% Empirically it works fine in average of regrets on lots of simulations, but on worst case regret (or in a histogram) we can see it does not have a logarithmic average regret
%
% BAD
% Explain on a small example ($K=2, M=3$ or $K=M=2$) that we found both manually, empirically and with a formal game tree exploration a small probability of failure, for both \Selfish-\UCB{} and \Selfish-\klUCB...
%
% a small probability of failure... $K=2, M=3$ or $K=M=2$. (idée: comparer \Selfish{} et \rhoRand{} en terme d'histogramme de la distribution de $R_T$)
%
% DONE estimer la proba ? ou la calculer avec le calcul formel ? ca c'est fait c'est bon
%
% a curiosity: when $K=M=2$ and $\mu_1 = \mu_2$, using \Selfish{} KL-UCB does orthogonalize !

\begin{figure}[!ht]
  \centering
  % \begin{subfigure}[!ht]{0.49\textwidth}
    \includegraphics[width=0.90\textwidth]{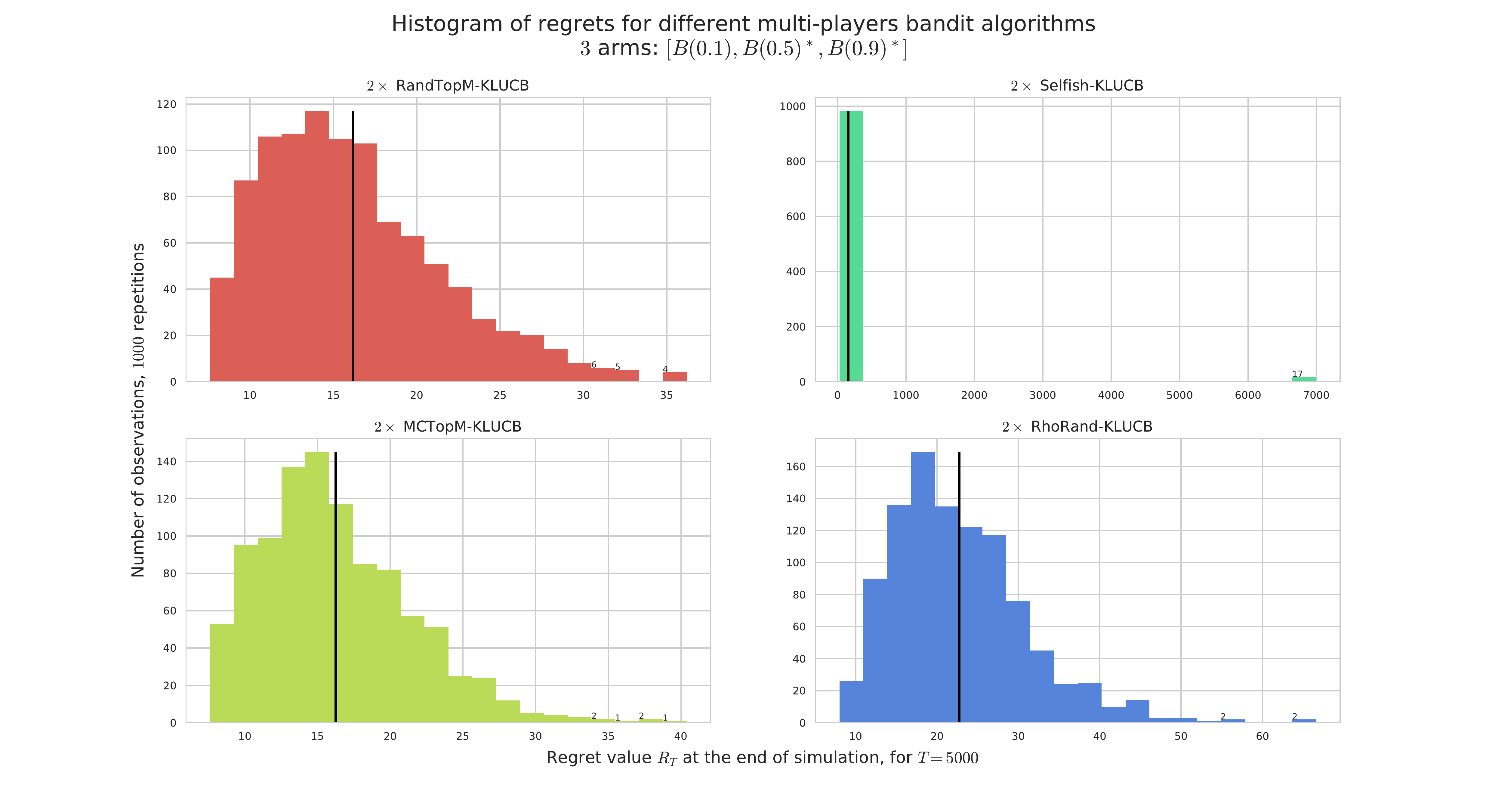}
  % \end{subfigure}
  % % ~
  % \begin{subfigure}[!ht]{0.49\textwidth}
  %   \includegraphics[width=1.10\textwidth]{all_HistogramsRegret____env1-1_1035303196230283176.pdf}
  % \end{subfigure}
  \caption{Regret for $M=2$ players, $K=3$ arms, horizon $T=5000$, $1000$ repetitions and $\boldsymbol{\mu} = [0.1, 0.5, 0.9]$. Axis $x$ is for regret (different scale for each part), and the \textcolor{darkgreen}{green} curve for \Selfish{} shows a small probability of having a linear regret ($17$ cases of $R_T \geq T$, out of $1000$). The regret for the three other algorithms is very small for this problem, always smaller than $100$ here.}
  \label{fig:selfish_fail1}
  % \vspace*{-15pt}  % XXX remove if problem
\end{figure}

\begin{figure}[!ht]
  \centering
  \begin{subfigure}[!ht]{0.49\textwidth}
    \includegraphics[width=1.10\textwidth]{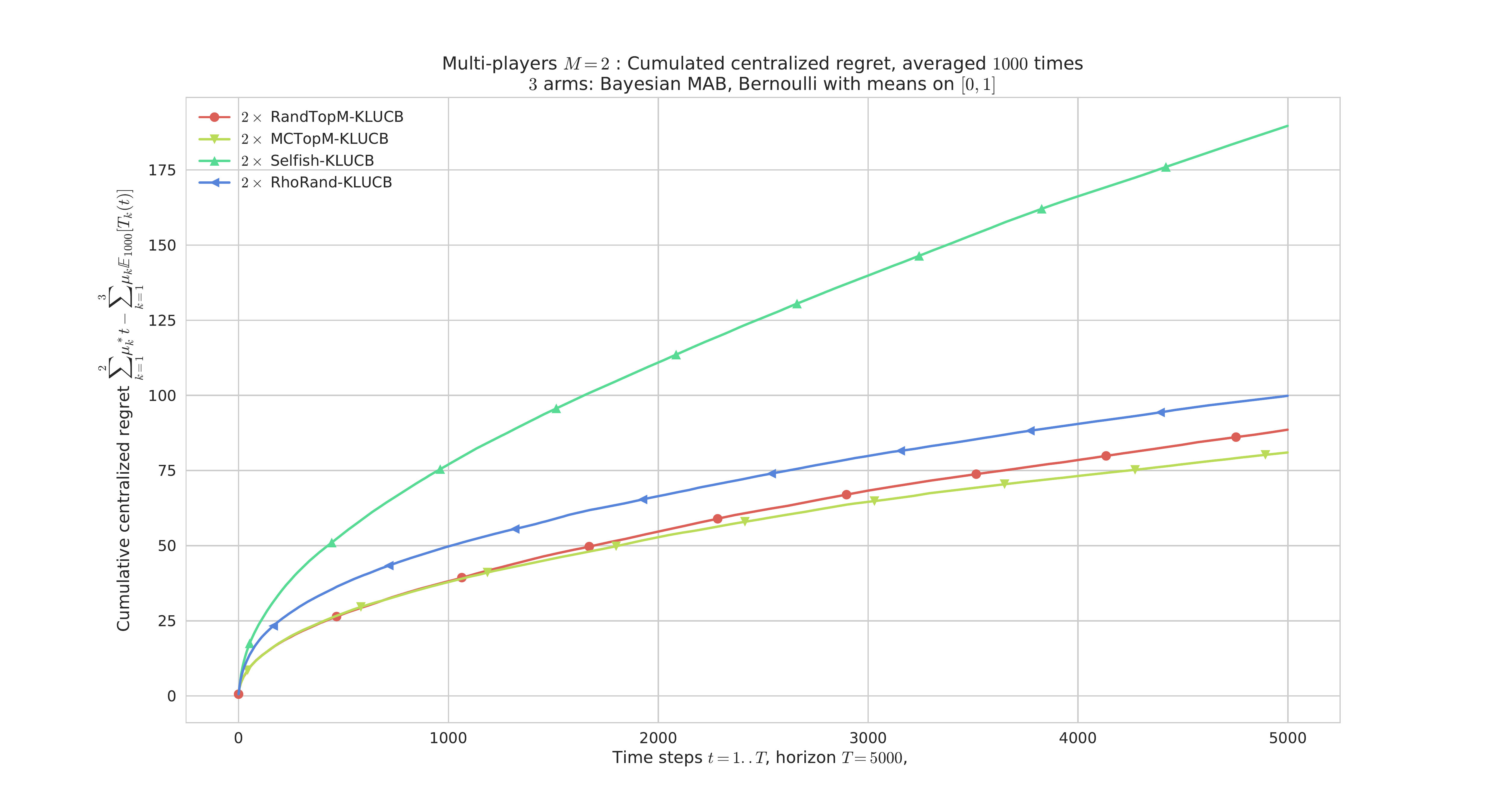}
  \end{subfigure}
  % ~
  \begin{subfigure}[!ht]{0.49\textwidth}
    \includegraphics[width=1.10\textwidth]{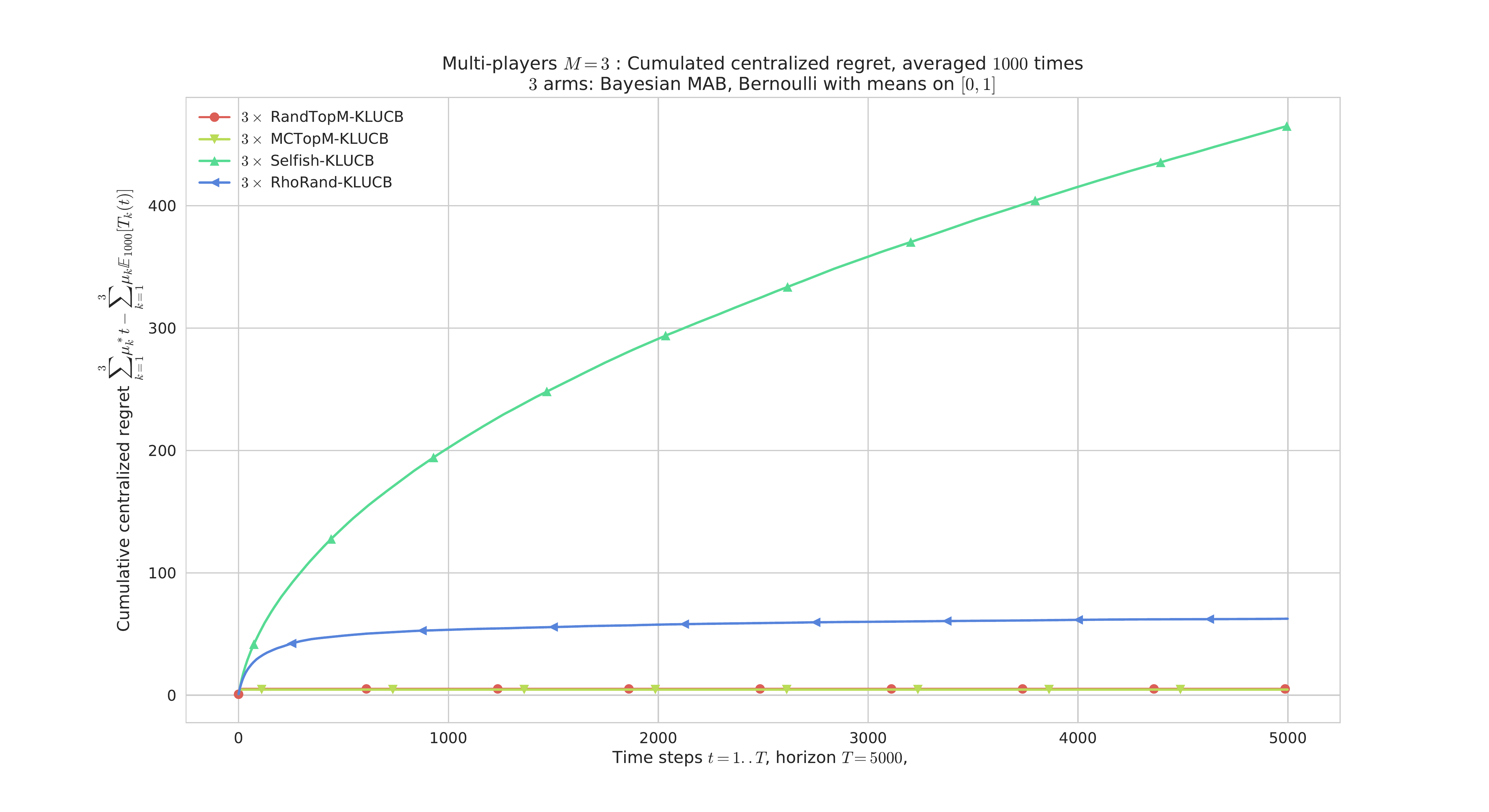}
  \end{subfigure}
  \caption{Regret, $M=2$ and $M=3$ players, $K=3$ arms, horizon $T=5000$, against $1000$ problems $\boldsymbol{\mu}$ uniformly sampled in $[0,1]^K$. \Selfish{} (top curve in \textcolor{darkgreen}{green}) clearly fails in such setting with small $K$.}
  \label{fig:selfish_fail2}
  % \vspace*{-15pt}  % XXX remove if problem
\end{figure}

As said before, analyzing \Selfish{} is harder,
but for instance one can prove that it yields constant
collisions and regret for the trivial case of $\mu_1=\mu_2=1$ and $M=2$.
Empirically, when \Selfish{} is compared to the other algorithms,
% with \UCB{} or \klUCB{} indices,
% and the average regret is considered (for let say $1000$ repetitions),
it is hard to find a case when \Selfish{} performs badly, as its (empirical average) regret always appeared logarithmic.
But an issue of only visualizing the empirical average regret for a certain number of repetitions is
that if a certain ``bad'' run happens only with small probability, it is possible that it never happened
in a simulation, or that it happened a few times but not enough to make the average regret look non-logarithmic\footnote{For instance, $0.999 \log(T) + 0.001 T$ looks more like $\log(T)$ than $T$ so an event yielding linear regret with ``small'' probability $10^{-3}$ cannot be observed from a plot showing the average regret.}.
This is why the distribution of regret, at the end of the simulations, $R_T$, is also displayed (in Appendix~\ref{app:moreplots}).
In a simple problem, with $M=2$ or $M=3$ players competing for $K=3$ arms,
for instance with means $\boldsymbol{\mu} = [0.1, 0.5, 0.9]$,
the histogram in Figure~\ref{fig:selfish_fail1} shows that with a small probability, the regret $R_T$ of \Selfish-\klUCB{} is not small (and appears linear).
Additionally, Figure~\ref{fig:selfish_fail2} shows that \Selfish-\klUCB{} also has bad performance against random uniform problems $\boldsymbol{\mu}\in[0,1]^K$,
for $M=2$ or $3$ and $K=3$, in a lot of cases.
In comparison, the others algorithms seem to have a logarithmic regret (for $M=2$ and all algorithms)
or even a constant regret (for $M=3$ and \RandTopM{} and \MCTopM).

The intuition behind these configurations when \Selfish{} performs poorly is the following,
if all players use the same algorithm and the same indices.
If two players $i$ and $j$ have exactly the same vectors $[\widetilde{S_k}^i(t)] = [\widetilde{S_k}^j(t)]$
and $[T_k^i(t)] = [T_k^j(t)]$ at some time step $t \geq 1$,
with different values for each $k$
and so that the index vectors $\mathbf{g}^j(t) = \mathbf{g}^{i}(t)$ have different values for each arm $k$,
then both players will take the same decisions at time $t$, and collide.
Colliding does not change $[\widetilde{S_k}^j(t+1)]$ but increase one value in $[T_k^j(t+1)]$ by $1$.
Then at the next step the same conditions on $\widetilde{S}^j$ and $N^j$ are preserved,
and if the same condition on $\mathbf{g}^j(t+1)$ are also preserved,
the two players will continue to collide.
We did not succeed in proving mathematically
that the preservation of the first hypothesis on $\widetilde{S}^j$ and $N^j$
implies the preservation of the hypothesis on index $\mathbf{g}^j$,
but numerically it turns out to be always the case:
and so two players colliding in such a setting will continue to do so infinitely:
we denote such configurations as \emph{absorbing}.

\begin{figure}[!ht]
  \centering

\begin{tikzpicture}[>=latex,line join=bevel,scale=0.5]
  \node (155) at (451.19bp,41.0bp) [draw=red,rectangle,thick] {[[1/2,0/1], [1/2,0/1]]};
  \node (10) at (635.19bp,215.0bp) [draw,rectangle] {[[0/0,1/1], [0/1,0/0]]};
  \node (3) at (267.19bp,215.0bp) [draw,rectangle] {[[0/0,0/1], [1/1,0/0]]};
  \node (6) at (451.19bp,215.0bp) [draw,rectangle] {[[1/1,0/0], [0/0,0/1]]};
  \node (14) at (83.193bp,128.0bp) [draw=red,rectangle] {[[0/1,1/1], [0/1,1/1]]};
  \node (1) at (83.193bp,215.0bp) [draw,rectangle] {[[0/1,0/0], [0/0,1/1]]};
  \node (0) at (359.19bp,302.0bp) [draw=darkgreen,rectangle] {[[0/0,0/0], [0/0,0/0]]};
  \node (22) at (267.19bp,128.0bp) [draw=red,rectangle] {[[1/1,0/1], [1/1,0/1]]};
  \node (234) at (635.19bp,41.0bp) [draw=red,rectangle,thick] {[[0/1,1/2], [0/1,1/2]]};
  \node (65) at (83.193bp,41.0bp) [draw=red,rectangle,thick] {[[0/1,1/2], [0/1,1/2]]};
  \node (110) at (267.19bp,41.0bp) [draw=red,rectangle,thick] {[[1/2,0/1], [1/2,0/1]]};
  \node (43) at (635.19bp,128.0bp) [draw=red,rectangle] {[[0/1,1/1], [0/1,1/1]]};
  \node (30) at (451.19bp,128.0bp) [draw=red,rectangle] {[[1/1,0/1], [1/1,0/1]]};
  \draw [red,->] (43) ..controls (635.19bp,98.163bp) and (635.19bp,82.548bp)  .. (234);
  \draw (645.69bp,84.5bp) node {$1$};
  \draw [black,->] (0) ..controls (225.73bp,289.45bp) and (154.26bp,280.04bp)  .. (128.19bp,266.0bp) .. controls (117.33bp,260.15bp) and (107.77bp,250.61bp)  .. (1);
  \draw (199.19bp,258.5bp) node {$\mu_{2}(1-\mu_{1})/4$};
  \draw [black,->] (0) ..controls (300.86bp,281.92bp) and (290.11bp,275.07bp)  .. (282.19bp,266.0bp) .. controls (276.58bp,259.57bp) and (273.06bp,251.1bp)  .. (3);
  \draw (353.19bp,258.5bp) node {$\mu_{1}(1-\mu_{2})/4$};
  \draw [red,->] (22) ..controls (267.19bp,98.163bp) and (267.19bp,82.548bp)  .. (110);
  \draw (277.69bp,84.5bp) node {$1$};
  \draw [black,->] (0) ..controls (485.53bp,296.17bp) and (538.83bp,287.27bp)  .. (582.19bp,266.0bp) .. controls (594.44bp,259.99bp) and (605.77bp,250.18bp)  .. (10);
  \draw (674.19bp,258.5bp) node {$\mu_{2}(1-\mu_{1})/4$};
  \draw [black,->] (1) ..controls (83.193bp,185.16bp) and (83.193bp,169.55bp)  .. (14);
  \draw (149.19bp,171.5bp) node {$\mu_{2}(1-\mu_{1})$};
  \draw [black,->] (10) ..controls (635.19bp,185.16bp) and (635.19bp,169.55bp)  .. (43);
  \draw (701.19bp,171.5bp) node {$\mu_{2}(1-\mu_{1})$};
  \draw [black,->] (6) ..controls (451.19bp,185.16bp) and (451.19bp,169.55bp)  .. (30);
  \draw (517.19bp,171.5bp) node {$\mu_{1}(1-\mu_{2})$};
  \draw [red,->] (14) ..controls (83.193bp,98.163bp) and (83.193bp,82.548bp)  .. (65);
  \draw (93.693bp,84.5bp) node {$1$};
  \draw [red,->] (30) ..controls (451.19bp,98.163bp) and (451.19bp,82.548bp)  .. (155);
  \draw (461.69bp,84.5bp) node {$1$};
  \draw [black,->] (0) ..controls (406.04bp,280.6bp) and (416.21bp,273.97bp)  .. (424.19bp,266.0bp) .. controls (430.81bp,259.39bp) and (436.24bp,250.81bp)  .. (6);
  \draw (507.19bp,258.5bp) node {$\mu_{1}(1-\mu_{2})/4$};
  \draw [black,->] (3) ..controls (267.19bp,185.16bp) and (267.19bp,169.55bp)  .. (22);
  \draw (333.19bp,171.5bp) node {$\mu_{1}(1-\mu_{2})$};
\end{tikzpicture}

  \caption{\small For $K=2$ arms and $M=2$ players using \Selfish-\UCB, for depth$=3$: $2$ \textbf{\textcolor{red}{absorbing configurations}}. Each rectangle represents a configuration, as the matrix $[[\widetilde{S_k}^j(t) / T_k^j(t) ]_j]_k$. Absorbing configurations from depth $2$ are case of equality of the two vectors and the \Selfish{} indices $\widetilde{g_k}^j(t)$. Transitions are labeled with their probabilities.}
  \label{fig:oneGameTree_SelfishKLUCB}
\end{figure}
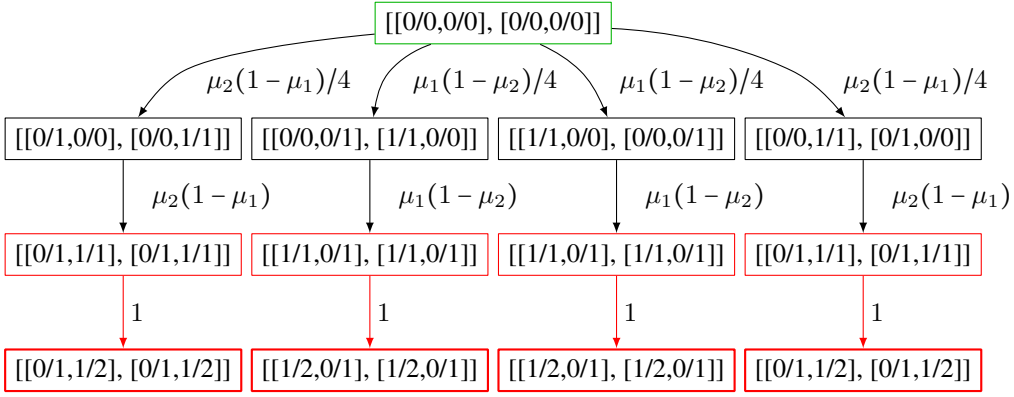

We wrote a script
\footnote{\url{banditslilian.gforge.inria.fr/docs/complete_tree_exploration_for_MP_bandits.html}}
that explores formally all the possible runs,
up-to a certain small time horizon,
by exploring the complete \emph{game tree},
of possible (random) rewards from the $K$ arms
and (random) actions from the $M$ players, up-to a small depth of let say $T=8$.
Such game tree becomes quickly very large, but this was enough to confirm that with
a certain small probability, function of $\mu_1,\dots,\mu_K$,
two players can arrive in just a few steps in a ``bad'' absorbing configuration.
For instance, for only $K=2$ arms, the following game tree in Figure~\ref{fig:oneGameTree_SelfishKLUCB} illustrates the first $3$ steps that can lead to $2$ absorbing configurations.
Using symbolic computations, the probability of reaching any of them was found to be
$\geq \mu_1^2(1-\mu_2)^2/2 + \mu_2^2(1-\mu_1)^2/2$ for \Selfish-\UCB.
That is far from being negligible, as it evaluates to $0.328$ for $\boldsymbol{\mu} = [0.1, 0.5, 0.9]$,
and a numerical simulation on $1000$ runs found $325$ cases of bad performance.
The same game tree exploration can be made for \Selfish-\klUCB,
but so far we were not able to justify why it experiences fewer cases of bad performances even though our software found the same (lower bound on) failure probability.

\begin{figure}[!ht]
  \centering
  \begin{subfigure}[!ht]{1.00\textwidth}
    \includegraphics[width=1.00\textwidth]{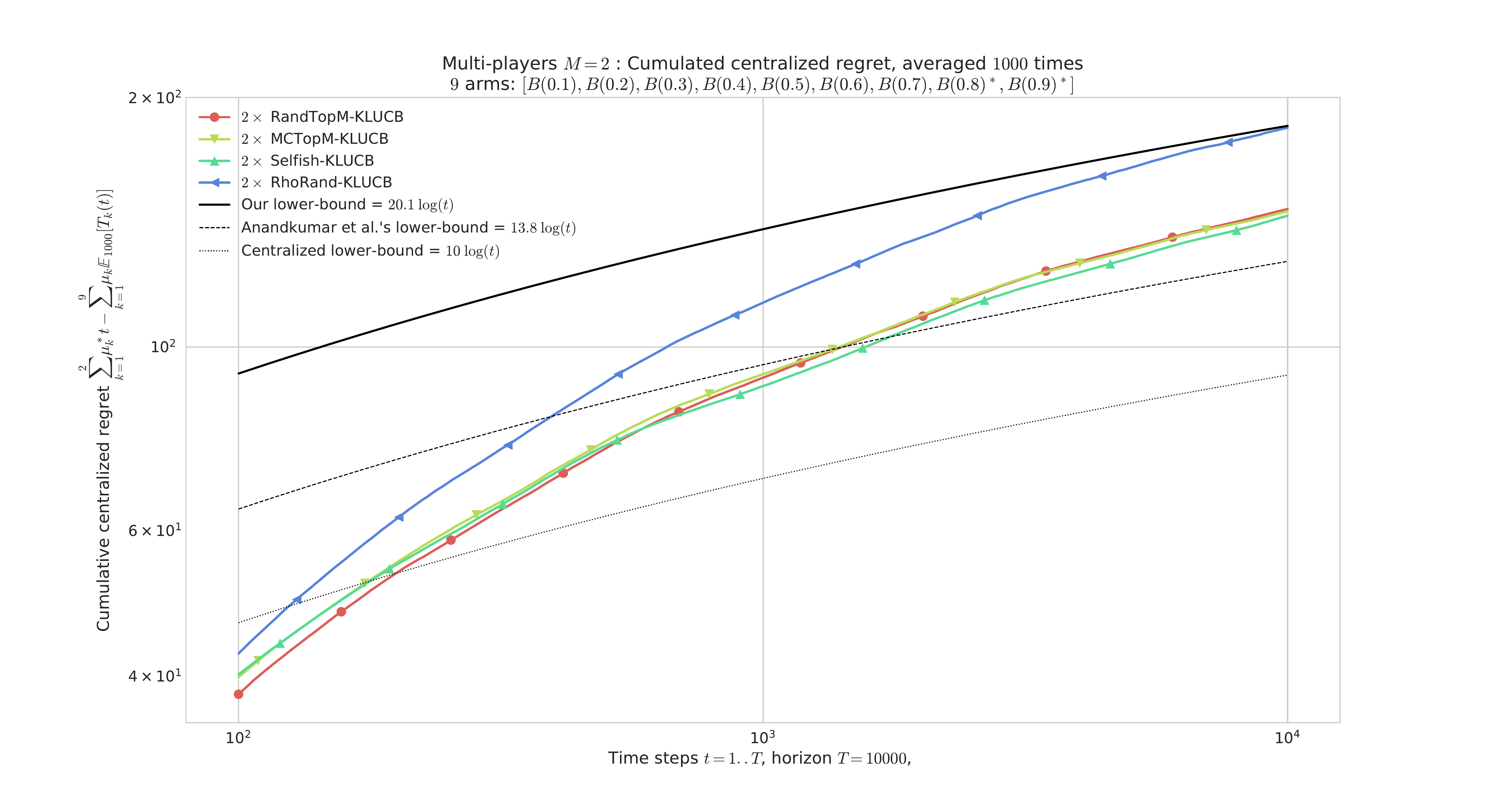}
  \end{subfigure}
  ~
  % \begin{subfigure}[!ht]{1.00\textwidth}
  %   \includegraphics[width=1.00\textwidth]{all_RegretCentralized_loglog____env1-1_8200873569864822246.pdf}
  % \end{subfigure}
  % ~
  \begin{subfigure}[!ht]{1.00\textwidth}
    \includegraphics[width=1.00\textwidth]{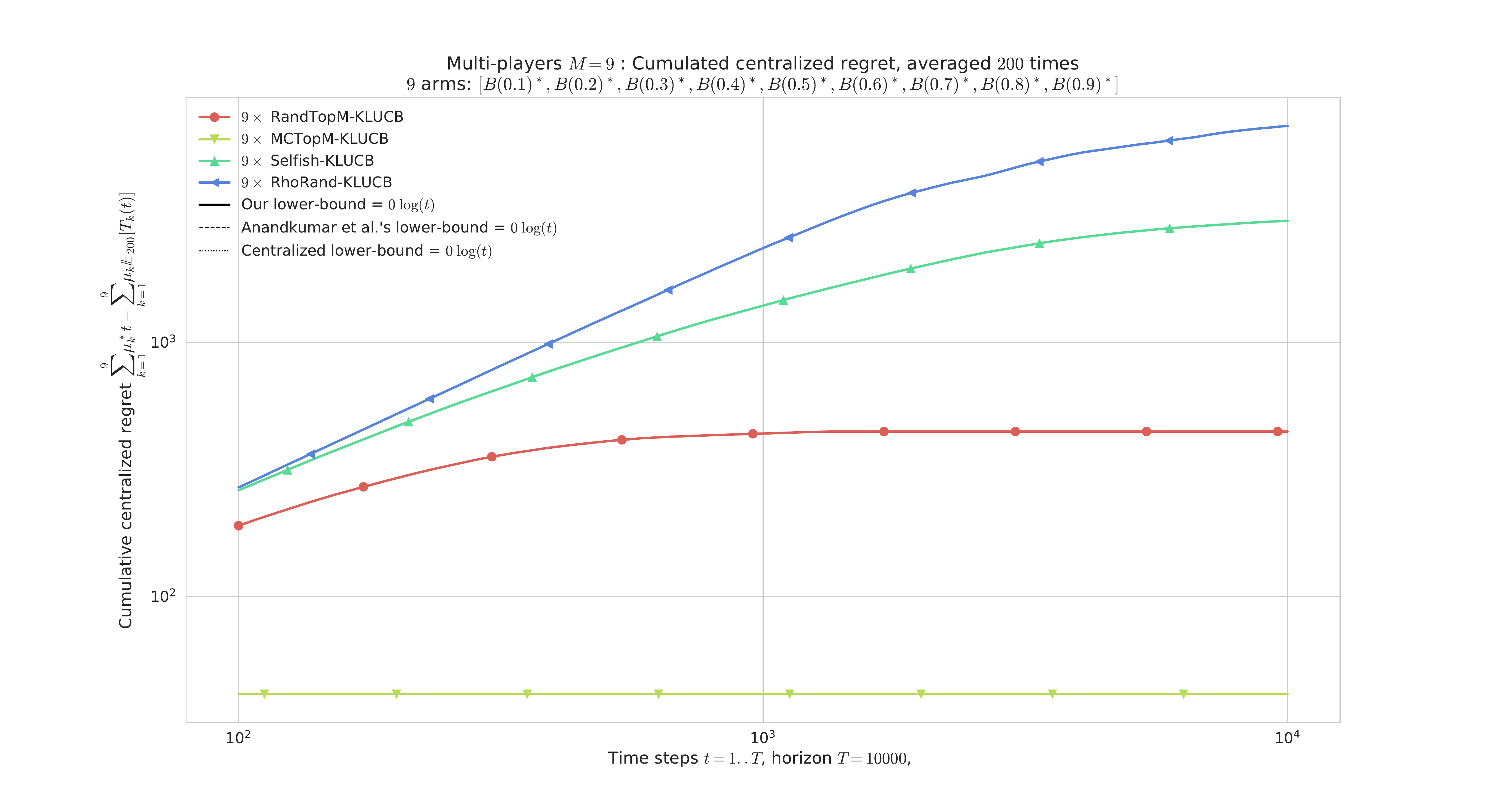}
  \end{subfigure}
  \caption{Regret (in $\log\log$ scale), for $M=2$ and $9$ players for $K=9$ arms, horizon $T=5000$, for problem $\boldsymbol{\mu}=[0.1,\dots,0.9]$. In different settings, \RandTopM{} (\textcolor{gold}{yellow} curve) and \Selfish{} (\textcolor{darkgreen}{green}) can outperform each other, and always outperform \rhoRand. \MCTopM{} is always among the best algorithms, and for $M$ not too small, its regret seems logarithmic with a constant matching the lower bound.}
  \label{fig:MP__K9_M2-6-9_T10000_N200__4_algos}
\end{figure}

\begin{figure}[!ht]
  \centering
  \begin{subfigure}[!ht]{0.75\textwidth}
    \includegraphics[width=1.00\textwidth]{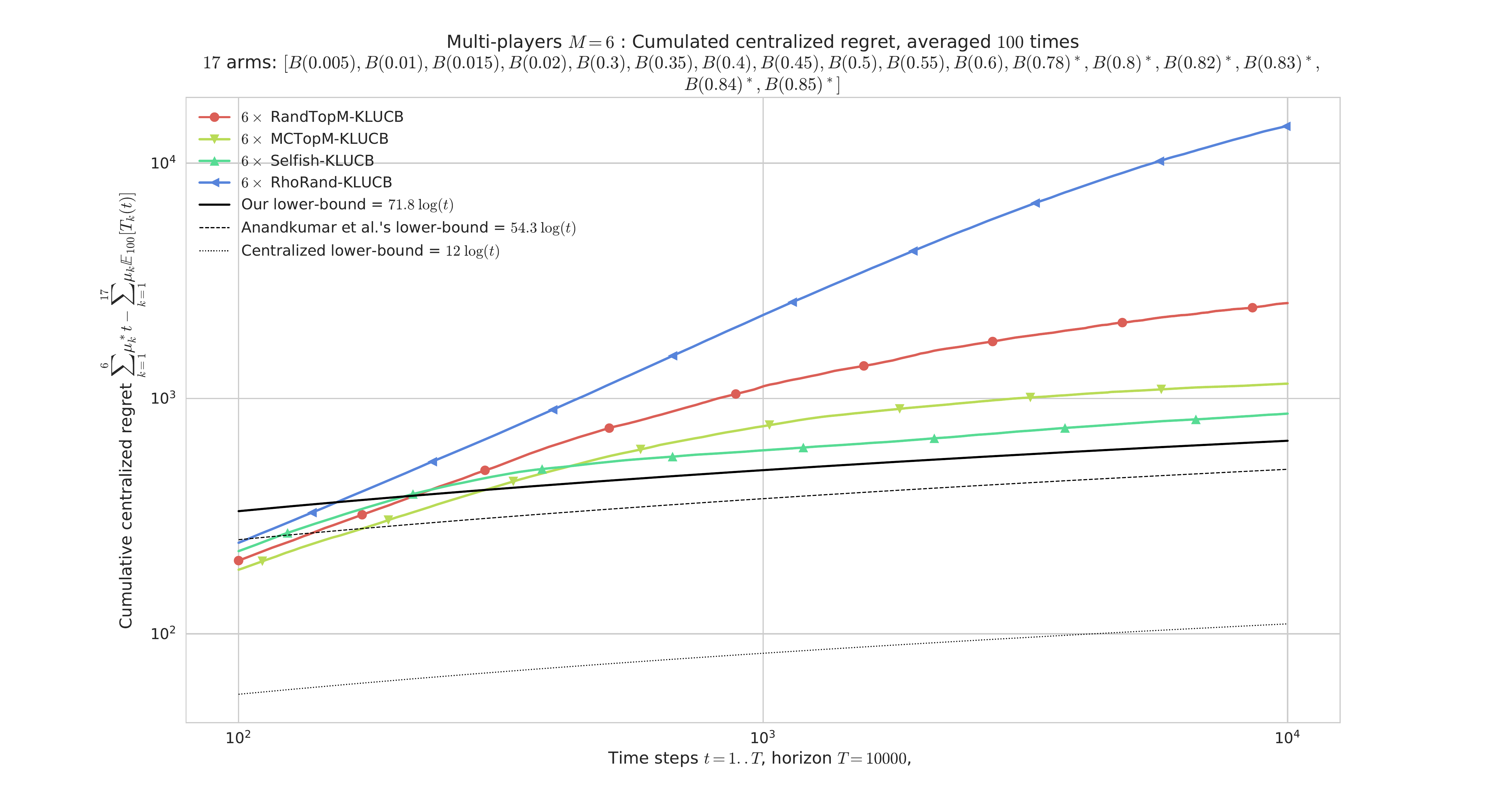}
  \end{subfigure}
  % ~
  \begin{subfigure}[!ht]{0.75\textwidth}
    \includegraphics[width=1.00\textwidth]{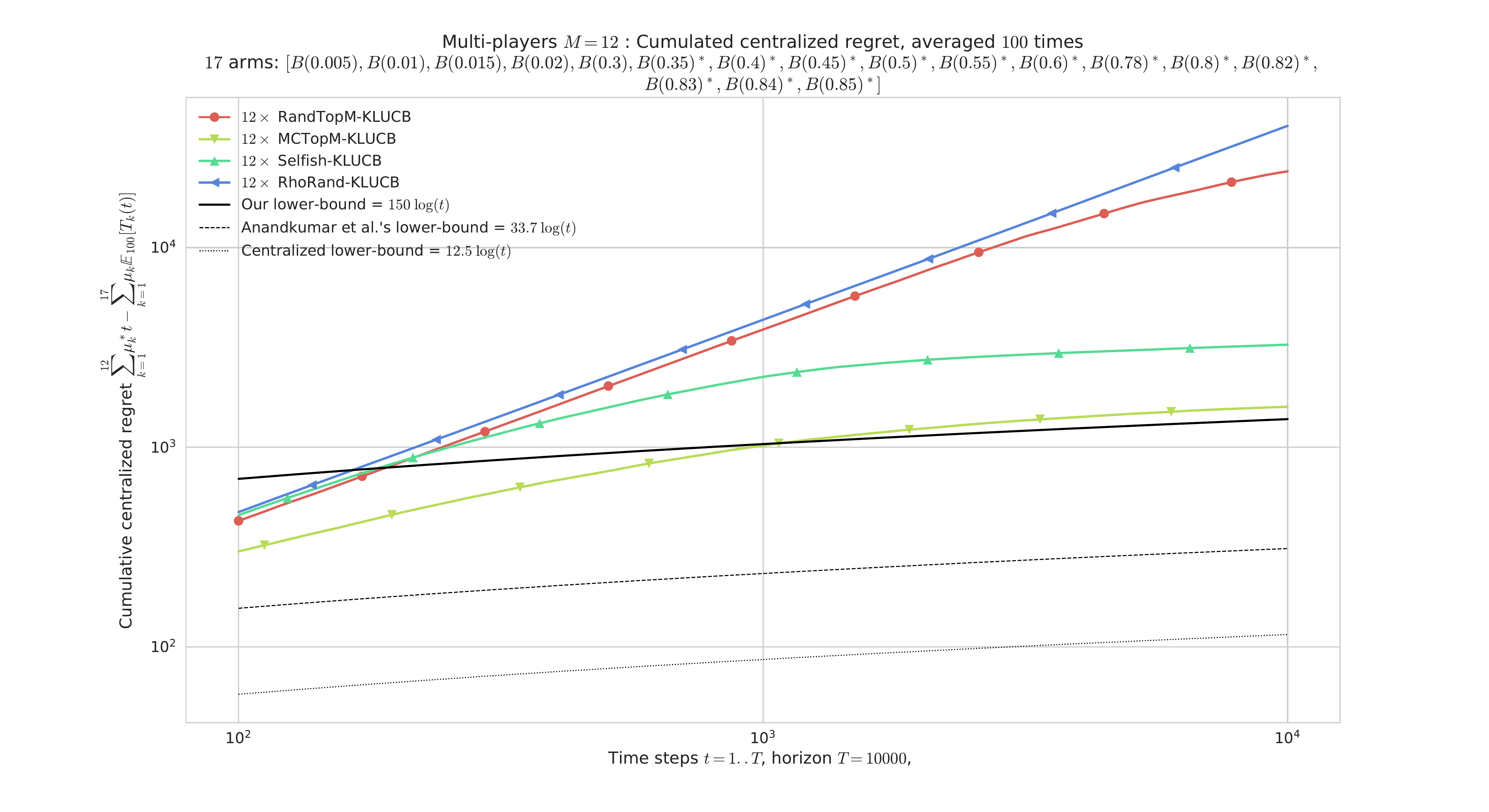}
  \end{subfigure}
  % ~
  \begin{subfigure}[!ht]{0.75\textwidth}
    \includegraphics[width=1.00\textwidth]{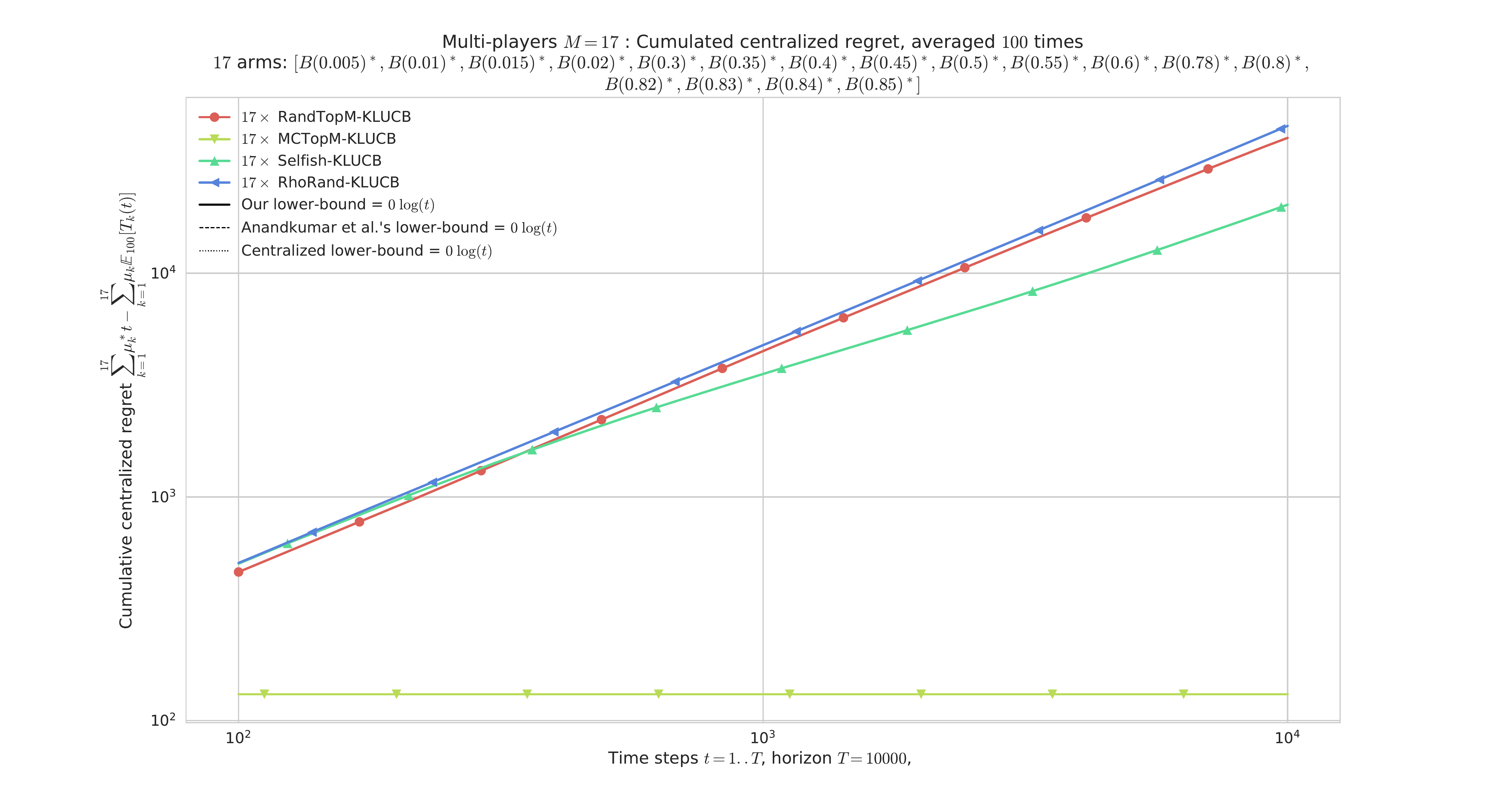}
  \end{subfigure}
  \caption{Regret (in $\log\log$ scale), for $M=6, 12, 17$ players for a ``difficult'' problem with $K=17$, and $T=5000$. The same observation as in Figure~\ref{fig:MP__K9_M2-6-9_T10000_N200__4_algos} can be made. \Selfish{} outperforms \MCTopM{} for $M=2$ here. Additionally, \MCTopM{} is the only algorithm to not fail dramatically when $M=K$ here.}
  \label{fig:MP__K17_M6-12-17_T10000_N100__4_algos}
\end{figure}

From the structure of such game tree, we conjecture that the probability of reaching absorbing configurations (before a certain time $t$) is always lower-bounded
by a polynomial function of $\mu_1,\dots,\mu_K$ and $1-\mu_1,\dots,1-\mu_K$,
of degree at most $t$ in each variable.
% with coefficients smaller than $1$.
% and without simple factors (\ie, $\mu_k^d$).
%
As such, the lower bound on probability of failures should decrease when $K$ and $M$ increase, and this is coherent with the experiments for $K=9$ or $K=17$ (see Figures~\ref{fig:MP__K9_M2-6-9_T10000_N200__4_algos} and \ref{fig:MP__K17_M6-12-17_T10000_N100__4_algos}),
where \Selfish{} is shown to be uniformly more efficient than \rhoRand.
Of course, one cannot run an infinite number of simulations, and the smaller the probability of failure, the less likely it is to observe a failure
in a finite number of runs.

% IDEA
\paragraph{Ideas to fix \Selfish{} ?}
It could be possible to change the \Selfish{} algorithm to add a way to escape such absorbing trajectories.
For instance one could imagine that after seen seeing, \eg, $x=10$ collisions in a row,
a certain random action could be taken by the players.
These tricks can work empirically in some cases,
but they are harder to analyze formally,
and it is hard to tune the parameters (here $x$, but possibly more),
and we do not find such tricks to be promising from a theoretical point-of-view.

% -----------------------------------------------------------------
% -----------------------------------------------------------------
\section{Additional Figures}
\label{app:moreplots}

The plots missing from Section~\ref{sec:experiments} are included here,
as well as some additional numerical results.

% -----------------------------------------------------------------
\subsection{Illustration of the lower bound}
\label{app:illustrationLowerBound}

\begin{figure}[!ht]
  \centering
  \includegraphics[width=0.70\textwidth]{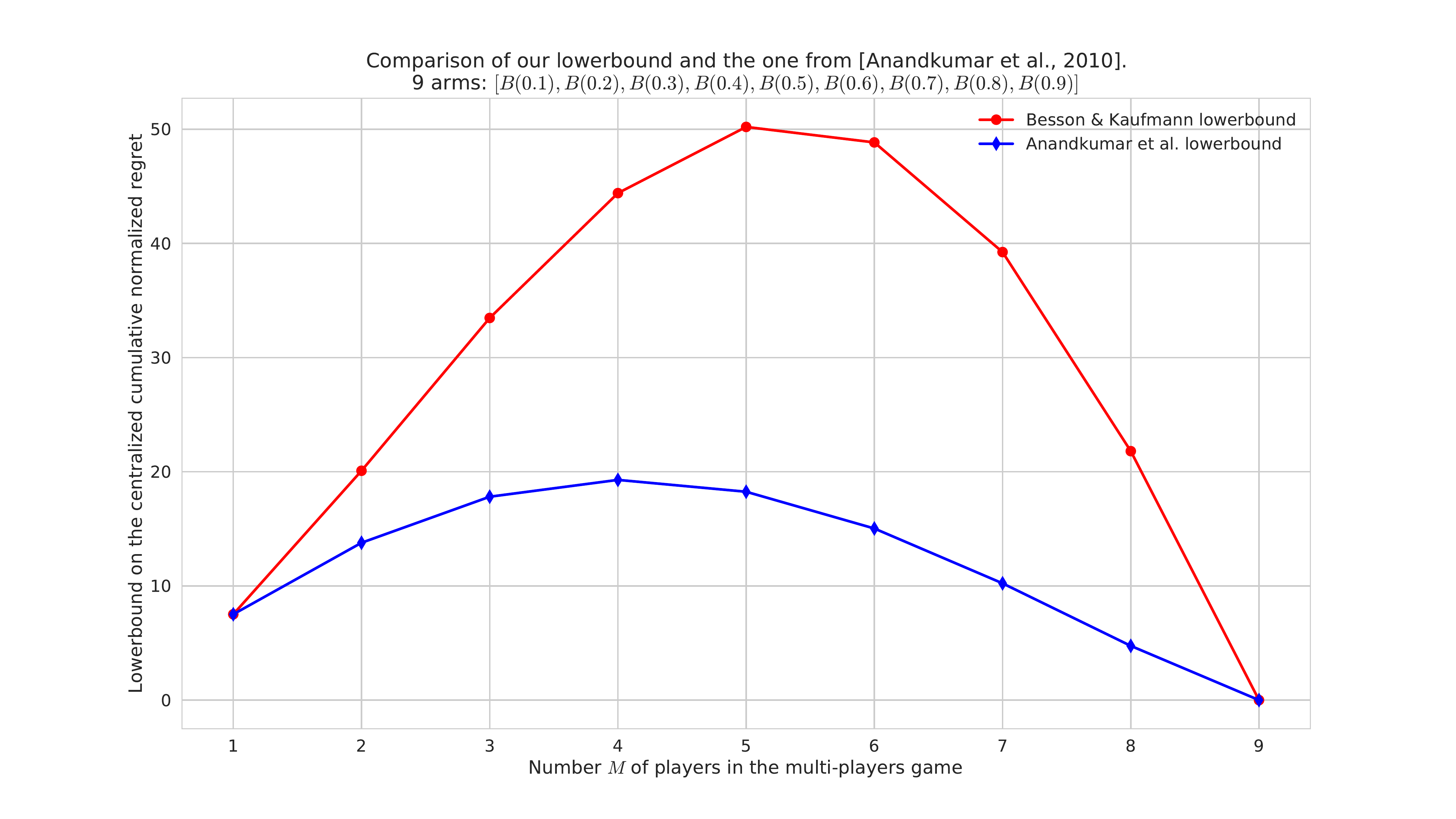}
  \caption{Comparison of our lower bound against the one  from \cite{Zhao10}, on a simple problem with $9$ Bernoulli arms, of means $\boldsymbol{\mu} = [0.1, 0.2, \dots, 0.9]$, as a function of the number of players $M$.}
  \label{fig:CompLowerBounds}
\end{figure}

We proved in Theorem~\ref{thm:BetterLowerBound} that the normalized regret, \ie, $R_T$ divided by
$\log T$, is asymptotically lower bounded by a constant $\mathrm{LB}(\boldsymbol{\mu}, M)$
depending on the problem $\boldsymbol{\mu}$ and the number of players $M$, for any $\rho$.
\begin{equation}
  \mathop{\lim\inf}\limits_{T \to +\infty} \frac{R_T(\boldsymbol{\mu}, M, \rho)}{\log T} \geq \mathrm{LB}(\boldsymbol{\mu}, M).
\end{equation}
For an example problem with $K = 9$ arms, we display below on the $x$ axis is
the number of player, from $1$ player to $9$ players, and on the
$y$ axis is the value of this constant $\mathrm{LB}(\boldsymbol{\mu}, M)$, from the initial
theorem and from our theorem.
We chose a simple problem, with Bernoulli
distributed arms, with $\boldsymbol{\mu} = [0.1, 0.2, \dots, 0.9]$.
% (as used in some articles).
%
Figure~\ref{fig:CompLowerBounds} clearly shows that our improved lower bound is indeed larger than the initial one by \cite{Zhao10},
and both become uninformative when $M=K$ (\ie, null).

%
% System regret and three terms
%
\begin{figure}[!ht]
  \centering
  \begin{subfigure}[!ht]{1.00\textwidth}
    \includegraphics[width=1.07\textwidth]{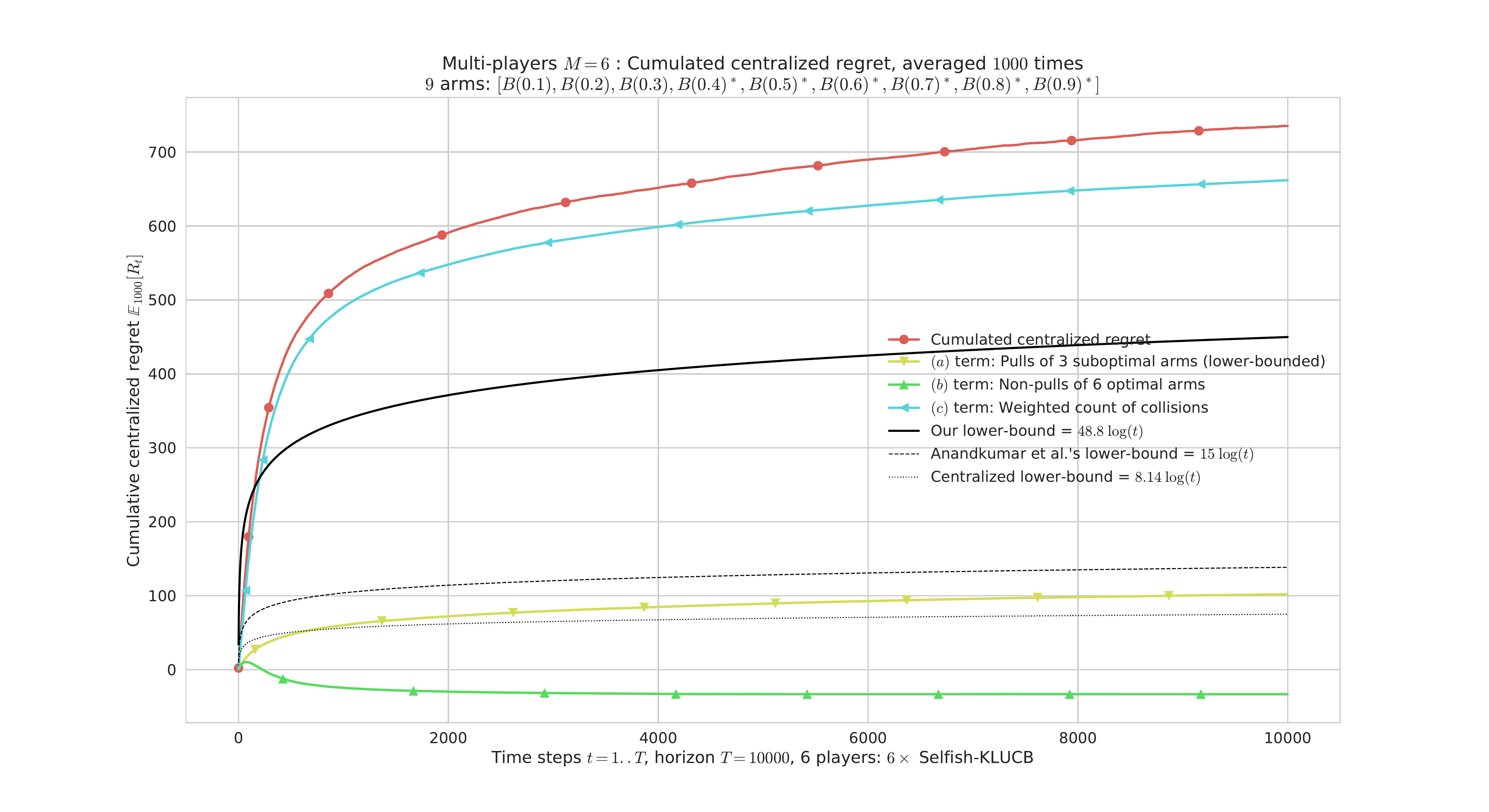}
  \end{subfigure}
  ~
  \begin{subfigure}[!ht]{1.00\textwidth}
    \includegraphics[width=1.07\textwidth]{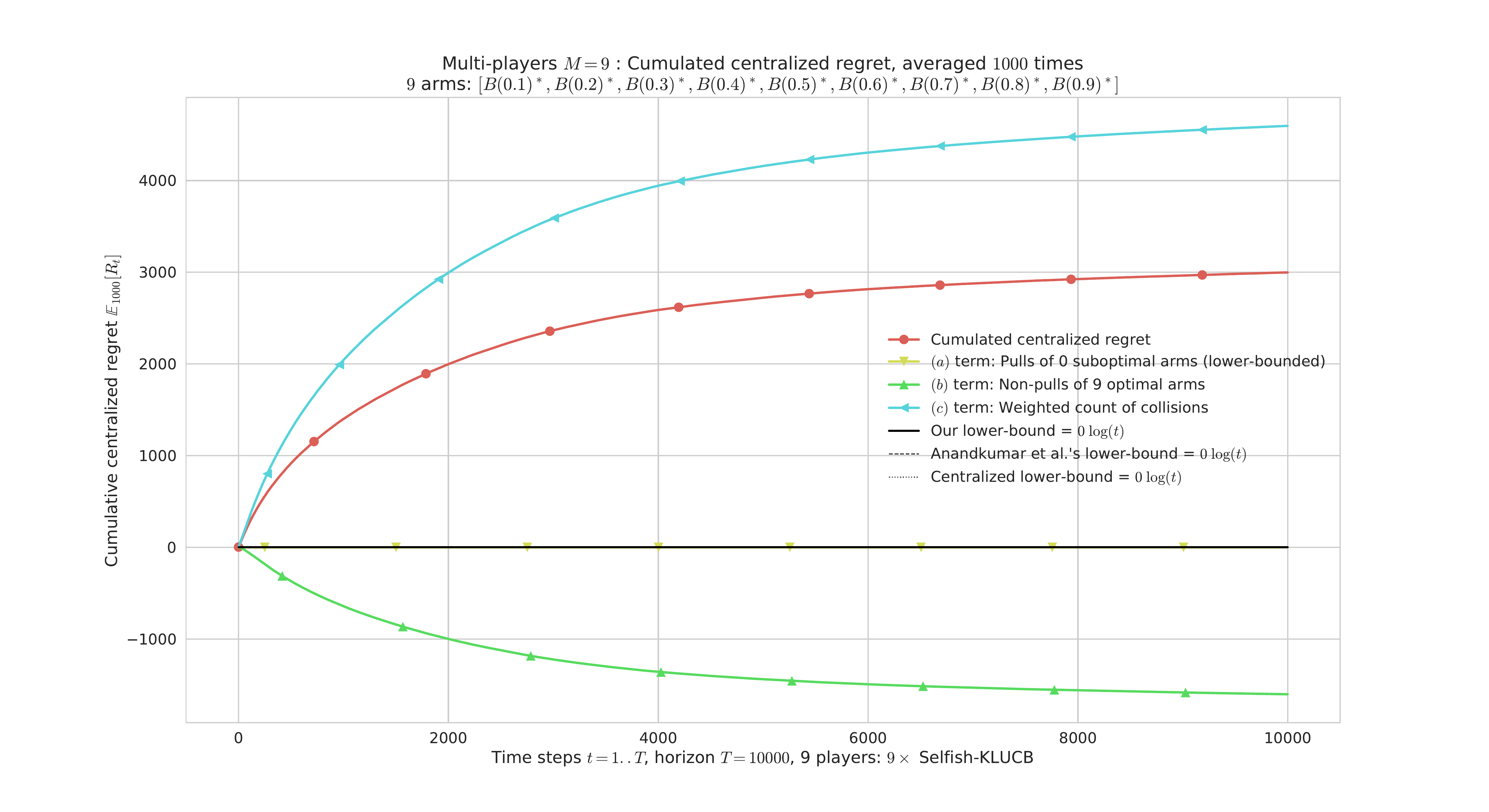}
  \end{subfigure}
  \caption{Regret with its three terms \termeun, \termedeux, \termetrois, and lower bounds \eqref{eq:ourLowerBound} and \eqref{eq:Zhao10LowerBound} in \textbf{black}, for \Selfish-\klUCB: $M=6$ and $M=9$ players, $K=9$ arms, horizon $T=10000$ (for $1000$ runs).}
  \label{fig:MP__M9_K9_T10000_N1000__9_algos__main_RegretCentralized____env6}
  % \vspace*{-15pt}  % XXX remove if problem
\end{figure}

Figures~\ref{fig:MP__M9_K9_T10000_N1000__9_algos__main_RegretCentralized____env6}
show the regret $R_T(\boldsymbol{\mu}, M, \rho)$
on the same example problem $\boldsymbol{\mu}$, with $K = 9$ arms and respectively $M = 6$, or $9$ players, for \Selfish-\klUCB.
It is just a simple way to check that the two lower bounds on the regret indeed appear as valid lower bounds empirically,
and are moreover lower bounds on the count of selections (\termeun, displayed in \textcolor{blue}{cyan}).
The lower bounds (in black) are $C(\boldsymbol{\mu}, M) \log t$, the dashed line
for \citeauthor{Zhao10}'s lower bound, and the continuous line is our lower bound.
These plot show the regret (in red),
% the three lower bounds
% (centralized, \citeauthor{Anandkumar11}'s and our lower bounds, in black),
and the three terms \termeun, \termedeux, \termetrois{} in the decomposition of the regret.
As explained in Lemma~\ref{lem:DecompositionRegret}, term \termedeux{} is not always non-negative.
For $M=9$ and \Selfish, \termetrois{} is actually larger than the regret,
and term \termeun{} is zero, as well as the lower bounds.

\subsection{Figures from Section~\ref{sec:experiments}}
\label{app:plotsFromSec5}

This last Appendix includes the figures used in Section~\ref{sec:experiments},
with additional comments.

\begin{figure}[!ht]
  \centering
  \begin{subfigure}[!ht]{0.85\textwidth}
    \includegraphics[width=1.00\textwidth]{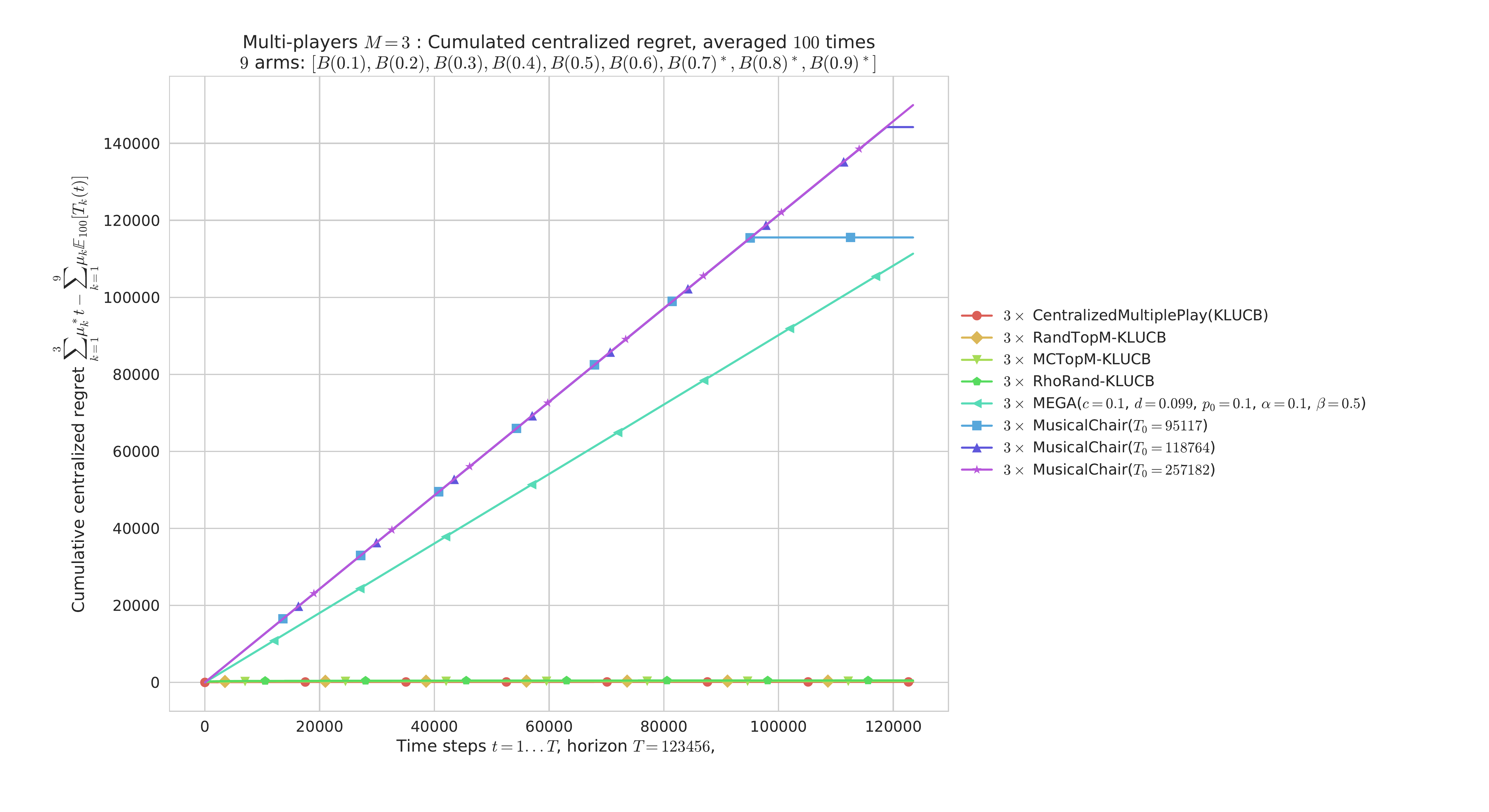}
  \end{subfigure}
  ~
  \begin{subfigure}[!ht]{0.85\textwidth}
    \includegraphics[width=1.00\textwidth]{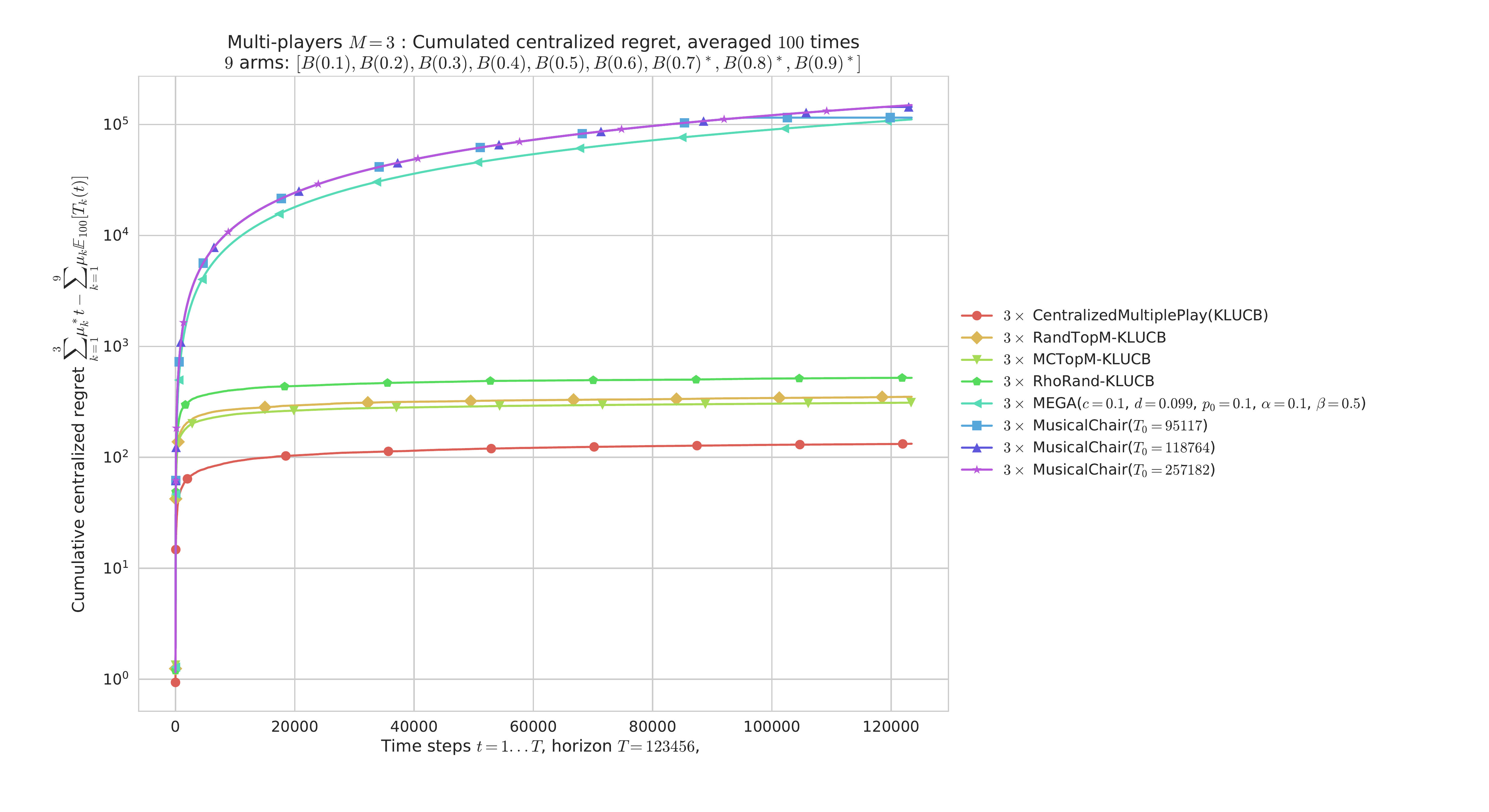}
  \end{subfigure}
  \caption{Regret (normal scale above, $\log$-y below) for $M=3$ players for $K=9$ arms, horizon $T=123456$, for $100$ repetitions on problem $\mu=[0.1,\dots,0.9]$. With a perfect knowledge on the gap ($\Delta=0.1$ here) and by using the parameters suggested from their respective articles, \MEGA{} and \MusicalChair{} perform badly in this simple setting, even with the knowledge of the horizon $T$ for \MusicalChair. The first two \MusicalChair{} instances use the optimal $T_0$ value from \cite{Rosenski16}, with $\varepsilon$ taken slightly smaller than the gap $\Delta$ ($\varepsilon=0.99 \Delta$), and respectively with $\delta=0.5$ and $\delta=0.1$, for which the regret can bounded with probability $0.5$ and $0.9$ respectively. The third instance uses the optimal $T_0$ corresponding to $\delta=1/T$, that is guaranteed to have an expected regret of order $\log(T)$.}
  \label{fig:MP__K9_M3_T123456_N100__8_algos}
\end{figure}

%
% Regular plots of centralized regrets
%

%
% Regular plots of centralized regrets
%

%
% Regular plots of centralized regrets
%
\begin{figure}[!ht]
  \centering
  \begin{subfigure}[!ht]{1.00\textwidth}
    \includegraphics[width=1.07\textwidth]{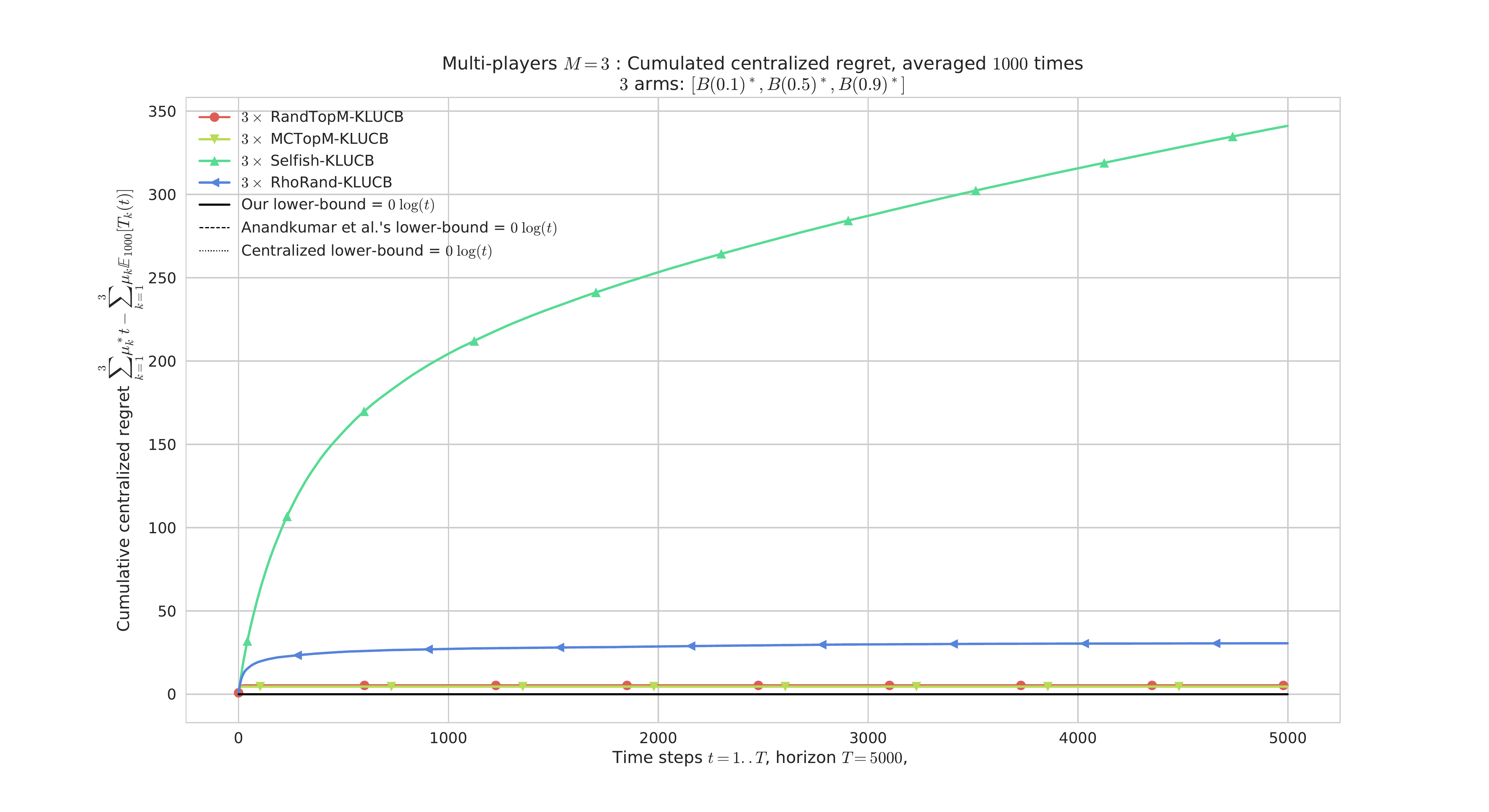}
  \end{subfigure}
  ~
  \begin{subfigure}[!ht]{1.00\textwidth}
    \includegraphics[width=1.07\textwidth]{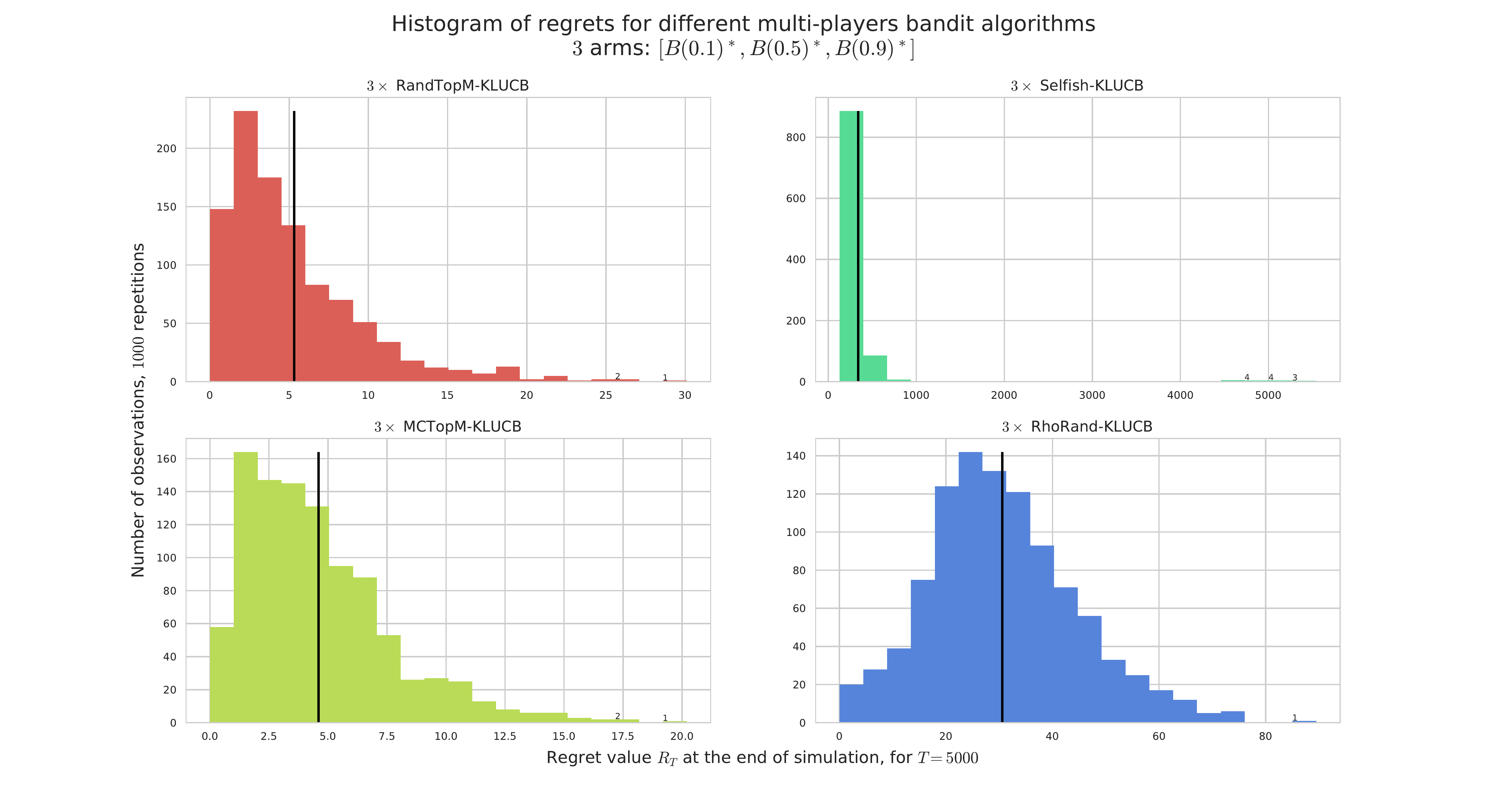}
  \end{subfigure}
  \caption{Regret for $M=3$ players, $K=3$ arms, horizon $T=5000$, $1000$ repetitions and $\boldsymbol{\mu} = [0.1, 0.5, 0.9]$. Axis $x$ is for regret (different scale for each), and the top \textcolor{darkgreen}{green} curve for \Selfish{} shows a small probability of having a linear regret ($11$ cases of $R_T \geq T$, out of $1000$). The regret for the three other algorithms is very small for this problem, and even appears constant.}
  \label{fig:selfish_fail3}
  % \vspace*{-15pt}  % XXX remove if problem
\end{figure}

%
% Regular plots of centralized regrets
%

\begin{figure}[!ht]
  \centering
  \begin{subfigure}[!ht]{1.00\textwidth}
    \includegraphics[width=1.07\textwidth]{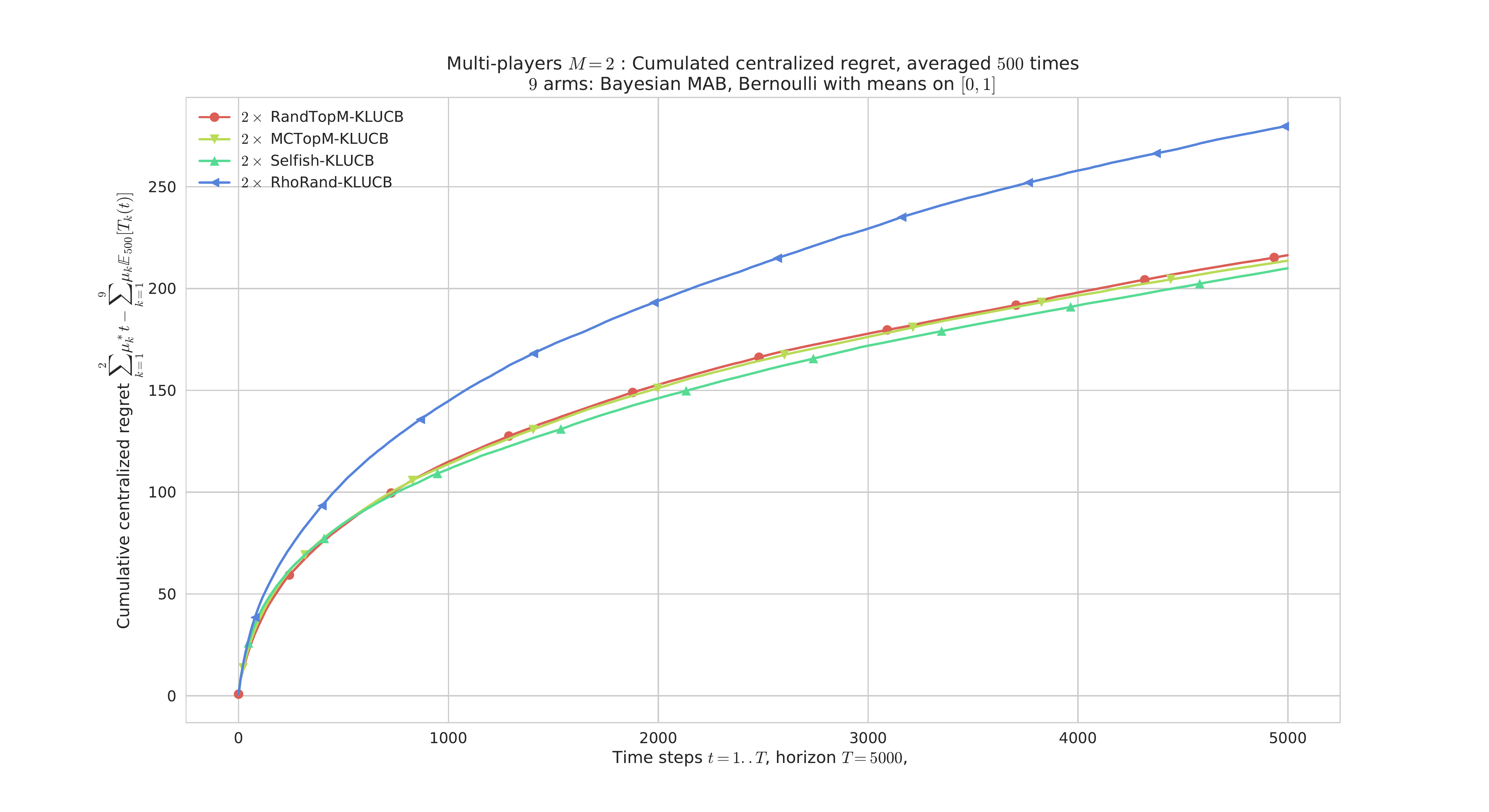}
  \end{subfigure}
  ~
  \begin{subfigure}[!ht]{1.00\textwidth}
    \includegraphics[width=1.07\textwidth]{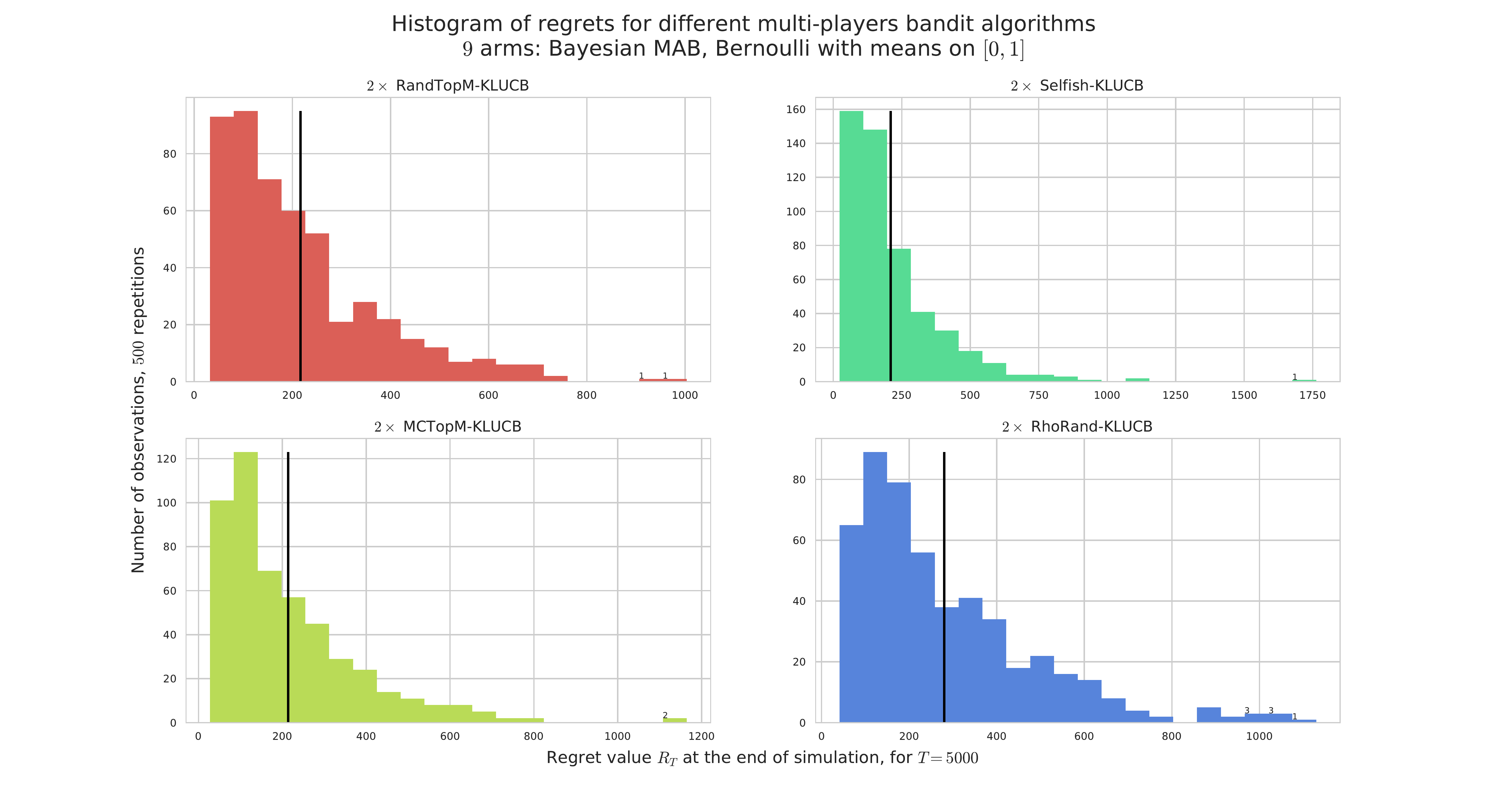}
  \end{subfigure}
  \caption{Regret, $M=2$ players, $K=9$ arms, horizon $T=5000$, against $500$ problems $\boldsymbol{\mu}$ uniformly sampled in $[0,1]^K$. \rhoRand{} (top \textcolor{blue}{blue}) is outperformed by the other algorithms (and the gain increases when $M$ increases), which all perform similarly in such configurations. Note that the (small) tail of the histograms come from complicated problems $\boldsymbol{\mu}$ and not failure cases.}
  \label{fig:MP__K9_M2_T5000_N500__4_algos__all_RegretCentralized__BayesianProblems}
  % \vspace*{-15pt}  % XXX remove if problem
\end{figure}

\begin{figure}[!ht]
  \centering
  \begin{subfigure}[!ht]{1.00\textwidth}
    \includegraphics[width=1.07\textwidth]{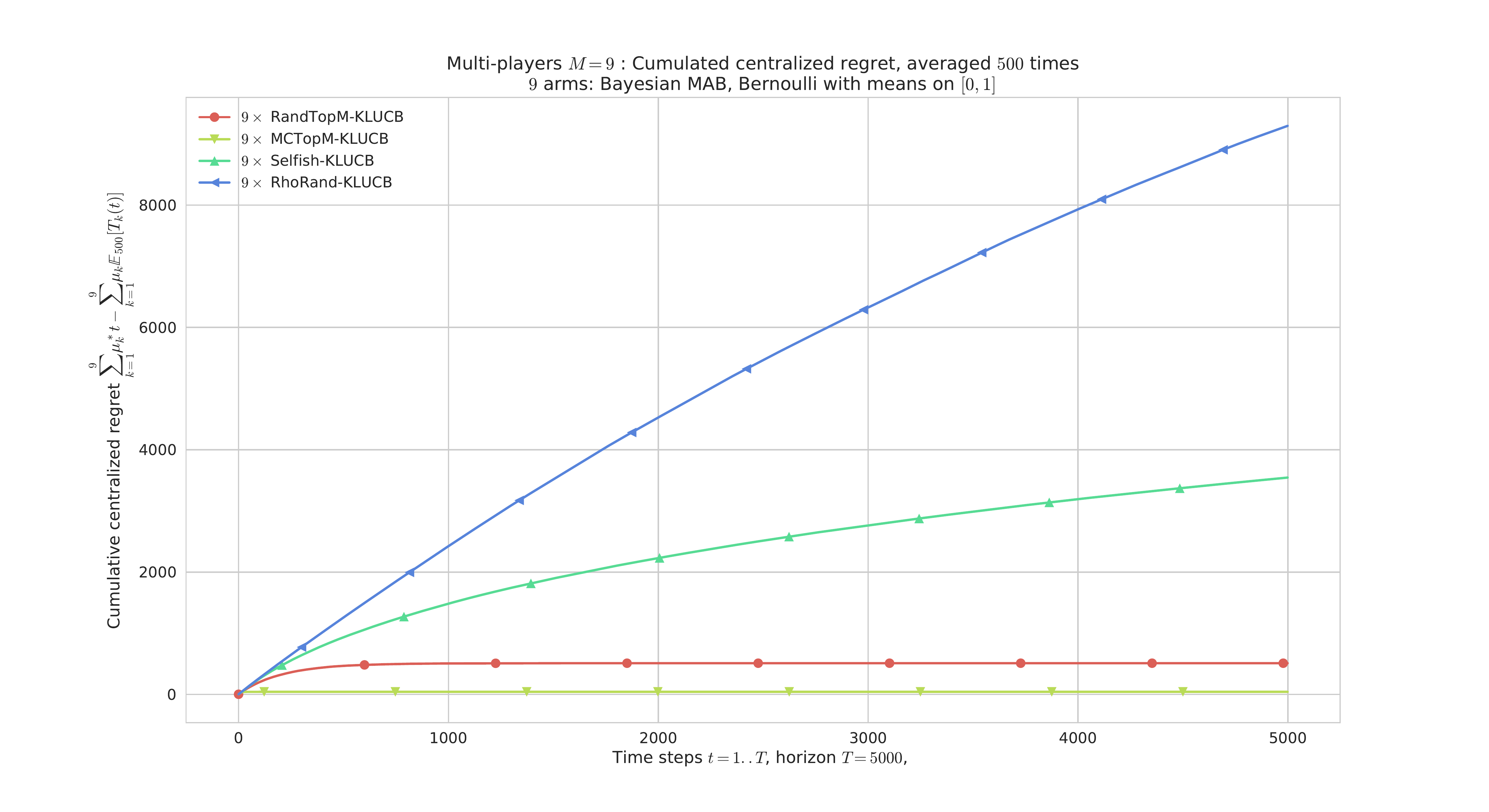}
  \end{subfigure}
  ~
  \begin{subfigure}[!ht]{1.00\textwidth}
    \includegraphics[width=1.07\textwidth]{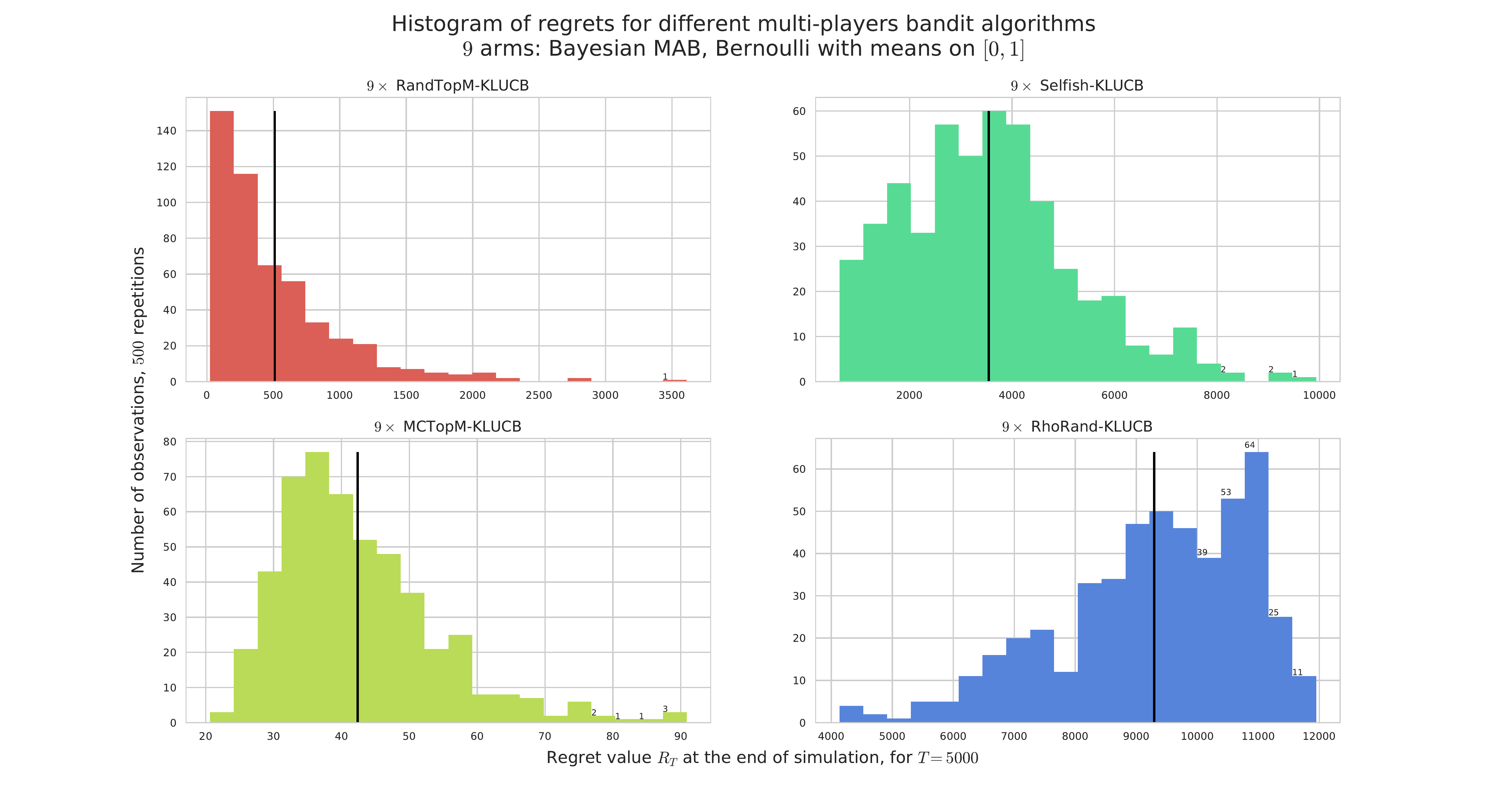}
  \end{subfigure}
  \caption{Regret, $M=9$ players for $K=9$ arms, horizon $T=5000$, against $500$ problems $\boldsymbol{\mu}$ uniformly sampled in $[0,1]^K$. This extreme case $M=K$ shows the drastic difference of behavior between \RandTopM{} and \MCTopM, having constant regret, and \rhoRand{} and \Selfish, having large regret.}
  \label{fig:MP__K9_M9_T5000_N500__4_algos__all_HistogramsRegret}
  % \vspace*{-15pt}  % XXX remove if problem
\end{figure}

% -----------------------------------------------------------------
% \nocite{*}  % XXX remove at the end, to only include quoted references!

\vskip 0.2in
% \hr{}
\emph{Note}:
  the simulation code used for the experiments is using Python 3.
  % https://www.Python.org/
  It is open-sourced at \verb|https://GitHub.com/SMPyBandits/SMPyBandits|
  %  but is available upon request,
  and fully documented at \newline
  \verb|https://SMPyBandits.GitHub.io|.
  % \citep{SMPyBandits}.

\end{document}